\documentclass[sn-mathphys-num]{sn-jnl}

\usepackage{graphicx}%
\usepackage{multirow}%
\usepackage{amsmath,amssymb,amsfonts}%
\usepackage{amsthm}%
\usepackage{mathrsfs}%
\usepackage[title]{appendix}%
\usepackage{xcolor}%
\usepackage{textcomp}%
\usepackage{manyfoot}%
\usepackage{booktabs}%
\usepackage{algorithm}%
\usepackage{algorithmicx}%
\usepackage{algpseudocode}%
\usepackage{listings}%
\usepackage[table]{xcolor}%
\usepackage{fancyhdr}   
\usepackage{hyperref}   

\theoremstyle{thmstyleone}%
\theoremstyle{thmstyletwo}%

\theoremstyle{thmstylethree}%

\renewcommand{\headrulewidth}{3pt}
\renewcommand{\footrulewidth}{0pt}

\begin{document}

\title[Article Title]{A unified framework for detecting point and collective anomalies in operating system logs via collaborative transformers}


\author[1]{\fnm{Mohammad} \sur{Nasirzadeh}}\email{nasirzadeh.moh@it.uut.ac.ir}

\author*[1]{\fnm{Jafar} \sur{Tahmoresnezhad}}\email{j.tahmores@it.uut.ac.ir}

\author[1]{\fnm{Parviz} \sur{Rashidi-Khazaee}}\email{p.rashidi@uut.ac.ir}

\affil*[1]{\orgdiv{Faculty of Information Technology and Computer Engineering}, \orgname{ Urmia University of Technology}, \orgaddress{\street{Band}, \city{Urmia}, \postcode{57166-17165}, \state{West Azerbaijan}, \country{Iran}}}


\abstract{Log anomaly detection is crucial for preserving the security of operating systems. Depending on the source of log data collection, various information is recorded in logs that can be considered log modalities. In light of this intuition, unimodal methods often struggle by ignoring the different modalities of log data. Meanwhile, multimodal methods fail to handle the interactions between these modalities. Applying multimodal sentiment analysis to log anomaly detection, we propose CoLog, a framework that collaboratively encodes logs utilizing various modalities. CoLog utilizes collaborative transformers and multi-head impressed attention to learn interactions among several modalities, ensuring comprehensive anomaly detection. To handle the heterogeneity caused by these interactions, CoLog incorporates a modality adaptation layer, which adapts the representations from different log modalities. This methodology enables CoLog to learn nuanced patterns and dependencies within the data, enhancing its anomaly detection capabilities. Extensive experiments demonstrate CoLog's superiority over existing state-of-the-art methods. Furthermore, in detecting both point and collective anomalies, CoLog achieves a mean precision of 99.63\%, a mean recall of 99.59\%, and a mean F1 score of 99.61\% across seven benchmark datasets for log-based anomaly detection. The comprehensive detection capabilities of CoLog make it highly suitable for cybersecurity, system monitoring, and operational efficiency. CoLog represents a significant advancement in log anomaly detection, providing a sophisticated and effective solution to point and collective anomaly detection through a unified framework and a solution to the complex challenges automatic log data analysis poses. We also provide the implementation of CoLog at \href{https://github.com/NasirzadehMoh/CoLog}{https://github.com/NasirzadehMoh/CoLog}.}





\maketitle

\pagestyle{fancy}

\fancyhf{}

\lhead{Accepted in \href{https://www.nature.com/srep/}{scientific reports}}
\rhead{DOI: \href{https://doi.org/10.1038/s41598-025-27693-4}{10.1038/s41598-025-27693-4}}

\renewcommand{\headrulewidth}{0.4pt}
\renewcommand{\footrulewidth}{0pt}

\section{Introduction}\label{sec1: introduction}

Anything that happens in a system during its interactions is recorded in system logs, including time-stamped events (e.g., transactions, errors, and intrusions). Anomalies, also known as outliers, are patterns or events in log data that deviate from the system's usual behavior. Log-based anomaly detection identifies and locates log outliers \citep{ref7}. Nevertheless, manually analyzing logs or applying traditional methods \citep{ref8, ref9, ref10, ref11, ref12, ref13, ref14, ref15, ref16, ref17, ref18, ref19, ref20, ref21, ref22, ref23, ref24, ref25} becomes more challenging due to log data's large-scale, complicated, and dynamic nature, as these methods primarily rely on manual processing and rule-setting \citep{ref5, ref26, ref27, ref28}. Furthermore, log data originating from various systems may employ distinct terminology. Hence, deep learning-based automated log anomaly detection is crucial to address threats or challenges promptly, define reaction thresholds, and make data-driven decisions \citep{ref29}.

In recent years, deep learning has shown considerable promise in automating many tasks \citep{ref30, ref31, ref32, ref33, ref34, ref35, saleki2024agry, tahmoresnezhad2017visual, khorshidpour2018domain, khodadadi2023hymovulnerabilitydetectionsmart}. One illustration of these tasks is the process of automated log anomaly detection. Due to deep learning techniques' outstanding results in automated log anomaly detection, utilizing these techniques for anomaly detection in log files can provide real-time alerts for crucial scenarios, including unidentified threats, identifying underlying causes, and mitigating cyberattacks, fraudulent activities, or system malfunctions \citep{ref4, ref36, ref37, ref38, ref39}. Consequently, deep neural networks have become the dominant modeling method in automated log-based anomaly detection in recent years. This research direction starts with DeepLog's revolutionary results in detecting anomalies from log files \citep{ref36}. Through the years, different deep learning-based approaches have evolved to become increasingly influential in this field \citep{ref6}. Numerous research directions exist within the automated log anomaly detection domain, such as multi-layer perceptrons (MLPs) \citep{ref40, ref41, ref42, ref43, ref44} that are used as part of other deep learning architectures, convolutional neural networks (CNNs) \citep{ref45, ref46, ref47, ref48, ref49, ref50}, recurrent neural networks (RNNs) \citep{ref36, ref38, ref51, ref52, ref53, ref54, ref55, ref56, ref57, ref58, ref59, ref60, ref61, ref62, ref63, ref64, ref65, ref66}, autoencoders \citep{ref41, ref67, ref68, ref69, ref70, ref71}, generative adversarial networks (GANs) \citep{ref41, ref51, ref72, ref73, ref74, ref75, ref76, ref77}, transformers \citep{ref26, ref78, ref79, ref80, ref81, ref82, ref83, ref84}, attention mechanisms \citep{ref37, ref40, ref47, ref85, ref86, ref87, ref88} that are usually employed to enhance performance in other deep neural architectures, graph neural networks (GNNs) \cite{ref89}, evolving granular neural networks (EGNNs) \cite{ref90}, and large language models (LLMs) \citep{ref4, ref91, ref92}.

\begin{figure}[h]
	\centering
	\includegraphics[width=1\textwidth]{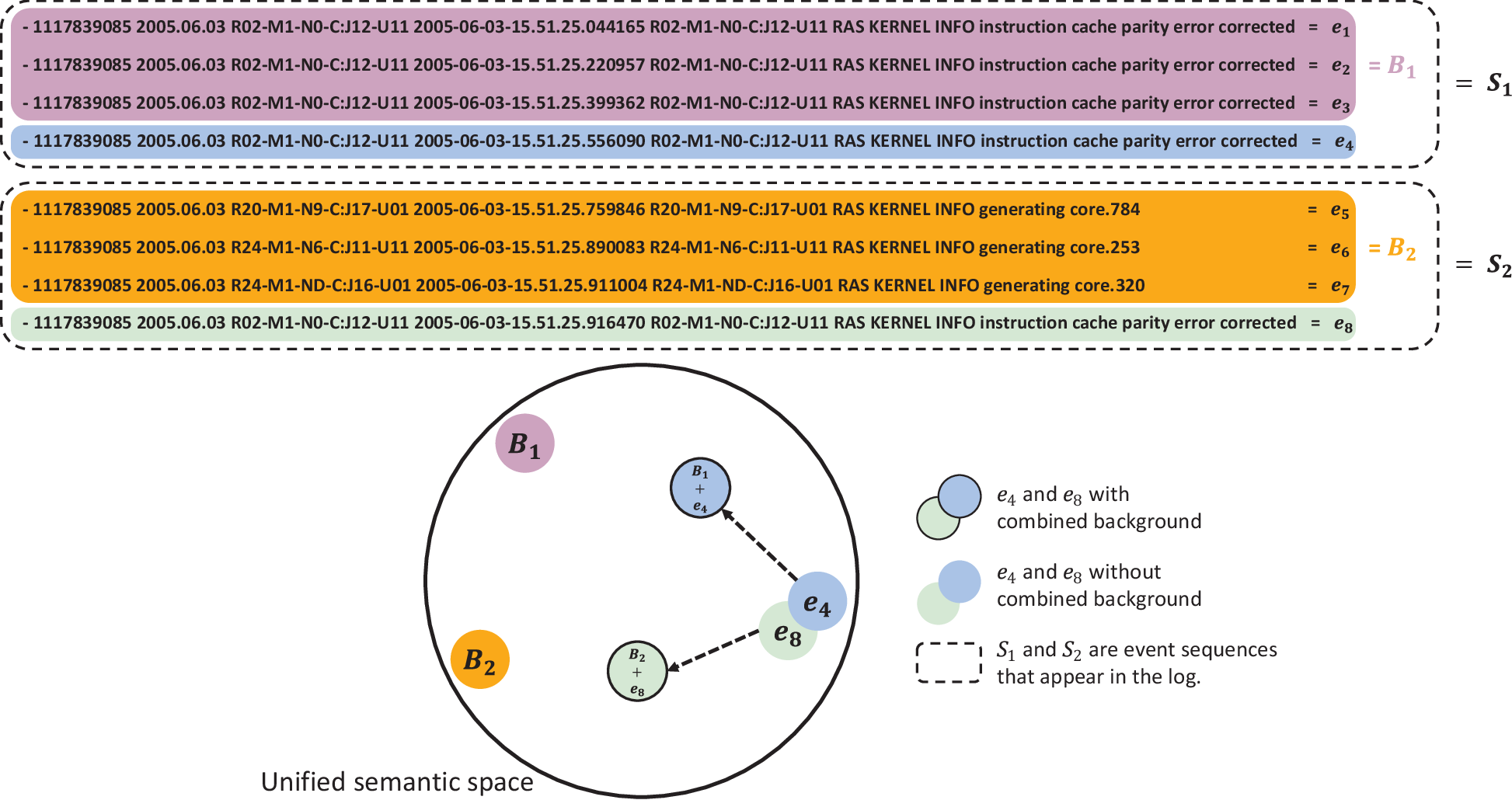}
	\caption{Illustration of how the sequence modality ($B_{1}$, $B_{2}$) adds more information to semantic modality ($e_{4}$, $e_{8}$) and how the same samples with different backgrounds are separated in a unified semantic space. For the $e_{4}$ and $e_{8}$ event vectors, their backgrounds, specifically $B_{1}$ and $B_{2}$, might be utilized to depict $e_{4}$ and $e_{8}$ in a more meaningful manner (it should be noted that $e_{4}$ is equivalent to $e_{8}$). In this sense, we designate the $B_{1}$ and $B_{2}$ as sequence modalities and the $e_{4}$ and $e_{8}$ as semantic modalities.}\label{fig1}
\end{figure}

As demonstrated in Figure~\ref{fig1}, depending on the type of log data collecting source, a wide range of information can be stored in logs. Practically, this data is expressed through different aspects, which we refer these aspects to as modalities. Each log file follows two primary modalities: (1) the sequence of records, which we refer to as sequence modality, and (2) the content of each record, which we refer to as semantic modality.

On one side, the sequence of events feature can contribute information to each log event from a semantic perspective. This means that various types of information in a log file can be used to create a more accurate model. Figure~\ref{fig1} illustrates how sequence and semantic modalities are relevant and how respective sequence vectors can be utilized to distinguish identical events' vectors. On the other side, these modalities allow the model to make decisions from various perspectives, thereby enhancing the precision of outcomes when performing the assigned tasks.

On the one hand, most existing literature approaches concentrate on unimodal anomaly detection from logs \citep{ref4, ref26, ref37, ref38, ref40, ref42, ref45, ref46, ref48, ref49, ref51, ref52, ref53, ref55, ref56, ref57, ref59, ref62, ref63, ref67, ref69, ref70, ref71, ref72, ref73, ref74, ref75, ref76, ref77, ref78, ref79, ref80, ref81, ref86, ref88, ref89, ref91, ref92}. These studies only utilize a specific facet of the information recorded in the log. However, disregarding other modalities of a log leads to the forfeiture of valuable data. In fact, focusing the analysis of log files based on a singular aspect of their information can lead to the model's inability to identify more complex anomalies. On the other hand, the multimodal approaches proposed in the literature utilize early \citep{ref50, ref58, ref60, ref61, ref82, ref84, ref85, ref87, ref90}, intermediate \citep{ref41, ref44, ref47, ref54, ref64, ref65, ref66}, and late \citep{ref43} fusion mechanisms or separate models \citep{ref36, ref68, ref83} for different modalities that they extract, leading to various challenges, which are presented in Table~\ref{tab1}. Concatenating unprocessed data from several modalities can lead to very high-dimensional input, which can be challenging to handle due to high-dimensionality challenges. Noise in the raw data may affect the model because noise in one modality may influence the information as a whole, indicating noise sensitivity challenges. Concatenating raw data leads to complexity, particularly with diverse data types and formats, leading to heightened data complexity challenges. Features derived from several modalities may not consistently possess direct compatibility or effortless combinability, highlighting compatibility challenges. Deploying distinct models for each modality can augment the system's overall complexity, resulting in heightened network complexity challenges. The contributions of several modalities can be challenging, mainly when one modality provides more information than the others, causing balancing contribution challenges. Merging the outcomes at the decision level or utilizing distinct models for each modality might ignore significant interactions between modalities, resulting in a lack of shared information challenges. Performing processes separately for each modality can result in redundant computations, mainly if overlapping features exist, causing redundancy challenges. The potential for enhancing performance may be restricted because the modalities have not been merged until the final decision step, leading to limited improvement challenges.

\begin{table}[h]
	\caption{Challenges arising from the multimodal approaches covered in the log anomaly detection literature.}\label{tab1}
	\begin{tabular*}{\textwidth}{@{\extracolsep\fill}lcccccc}
		\toprule & \multicolumn{4}{@{}c@{}}{Methods}                                                                     \\
		\cmidrule{2-5}%
		                               & Early                         & Intermediate                                                & Late        & Separate \\
		Challenges                     & fusion                        & fusion                                                      & fusion      & Models   \\
		\midrule[0.01cm]
		High dimensionality            & all except                    & \textit{Mdfulog\footnotemark[2], Swisslog\footnotemark[3],} & $\times$    & $\times$ \\
		                               & \textit{MLog}\footnotemark[1] & \textit{WDLog\footnotemark[4], and FSMFLog\footnotemark[5]} &             &          \\
		\midrule[0.01cm]
		Noise sensitivity              & all                           & $\times$                                                    & $\times$    & $\times$ \\
		\midrule[0.01cm]
		Heightened data complexity     & all                           & $\times$                                                    & $\times$    & $\times$ \\
		\midrule[0.01cm]
		Compatibility of features      & all                           & \textit{Mdfulog, Swisslog,}                                 & $\times$    & $\times$ \\
		                               &                               & \textit{WDLog, and FSMFLog}                                 &             &          \\
		\midrule[0.01cm]
		Heightened network complexity  & $\times$                      & all                                                         & all         & all      \\
		\midrule[0.01cm]
		Balancing contributions        & all                           & \textit{Mdfulog, Swisslog,}                                 & all         & $\times$ \\
		                               &                               & \textit{WDLog, and FSMFLog}                                 &             &          \\
		\midrule[0.01cm]
		Lack of shared information     & $\times$                      & $\times$                                                    & all         & all      \\
		\midrule[0.01cm]
		Redundancy                     & $\times$                      & $\times$                                                    & all         & all      \\
		\midrule[0.01cm]
		Limited improvement            & $\times$                      & $\times$                                                    & all         & all      \\
		\botrule
	\end{tabular*}
	\footnotetext[1]{\citeauthor{ref60} proposed \textit{MLog}}
	\footnotetext[2]{\citeauthor{ref44} proposed \textit{Mdfulog}}
	\footnotetext[3]{\citeauthor{ref54} proposed \textit{Swisslog}}
	\footnotetext[4]{\citeauthor{ref64} proposed \textit{WDLog}}
	\footnotetext[5]{\citeauthor{ref66} proposed \textit{FSMFLog}}
\end{table}

Considering the above motivations, we propose a novel automated log-based anomaly detection approach - called CoLog - based on multimodal sentiment analysis (MSA) \citep{ref93, ref94, ref95, ref96, ref97, ref98, ref99, ref100, ref101, ref102} to overcome the above mentioned challenges. CoLog interprets anomalies in the log as negative sentiments. In the same way, normal samples are classified as positive sentiments. Since the two most important modalities of a log file are the semantic features of an event and the features obtained from the sequence of events. The background and context of an event can be formed as sequence modality. CoLog constructs semantic and sequence modalities to learn more about anomalies through the interaction between these modalities. During this process, we employ a form of the guided-attention (GA) mechanism \citep{ref103} to encode each modality in an collaborative manner with other modalities. Finally, to accomplish the objective of prediction through integrating modalities, we apply transformer blocks and then transform the outcome feature vectors of each modality into a higher-dimensional space known as the latent space to perform meaningful fusion.

In addition, logs are often the place where the point and collective anomalies are recorded. Detecting point and collective anomalies in log records is crucial, as each anomaly type exposes distinct facets of system behavior. Point anomalies, representing isolated instances of abnormal behavior, often signal immediate issues like a sudden malfunction or attack. Collective anomalies are patterns of anomalous activity over time that may signify deeper, more systemic issues, such as steady performance degradation or persistent security threats. Detecting different kinds of abnormalities provides a thorough awareness of short-term and long-term issues, facilitating more effective troubleshooting and preventative care. However, all existing works solely concentrate on identifying only one type of log anomaly. Besides that, to the best of our knowledge, two techniques exist outside the log-based anomaly detection area that aim to identify point and collective anomalies utilizing a unified framework. \citeauthor{ref104} processed a time series using temporal convolutional network (TCN) and calculated the anomaly score to assess whether the point or collective data was anomalous based on the fixed-length sequences. However, due to its reliance on a predetermined sequence length, it may fail to identify certain instances of collective anomalies. Moreover, \citeauthor{ref105} used an improved local outlier factor (ILOF) to detect variable-length collective anomalies where uses identified point anomalies as boundary points to partition the residual sequence into many subsequences of varied lengths. Each subsequence's anomaly score is computed using the ILOF method. This method relies on calculating anomaly scores, which can lead to false alarms, particularly in long sequences. Furthermore, neither of these approaches explicitly learns the relationships between the various modalities. Therefore, leveraging the transformer \citep{ref106} architecture, we propose CoLog to handle the problem of detecting point and collective log anomalies by utilizing a unified framework. To achieve this objective, we propose utilizing collaborative transformer (CT) to process log modalities in an end-to-end manner, enabling them to interact and learn about their relationships. Since CoLog is a supervised approach, it does not face the challenge of accumulating errors.

The summarization of our contribution is as follows.

\begin{enumerate}[1.]
	\item We propose CoLog, the first framework to formulate log anomaly detection as a multimodal sentiment analysis problem, enabling the detection of both point and collective anomalies in a unified manner.
	
	\item We design a collaborative transformer architecture with impressed attention and a modality adaptation layer to capture nuanced interactions between semantic and sequence modalities.
	
	\item We introduce a balancing mechanism to adaptively weight contributions of modalities, which helps to ensure that the modalities with inherent differences are represented in the same representation space.
	
	\item We assess CoLog using seven benchmark datasets, achieving state-of-the-art performance, and provide the CoLog implementation to facilitate reproducibility in the GitHub repository.
	
	\item Our research findings and specialized services are available on the \href{https://www.alarmif.com}{Alarmif} website, facilitating a direct link between research and practical application.
\end{enumerate}

The paper is organized as follows. Section~\ref{sec2: relatedwork} discusses previous research in the domain. Define the task, the importance of transformers for this work, the threat model, and the assumptions outlined in Section~\ref{sec3: preliminaries}. Section~\ref{sec4: method} will discuss CT's design and training considerations for anomaly detection from logs based on MSA. The evaluation is presented in Section~\ref{sec5: experiments}, where we conclude our work and then offer suggestions for future research in Section~\ref{sec6: conclusion}.

\section{Related Work}\label{sec2: relatedwork}

Numerous research endeavors have examined log-based anomaly detection \citep{ref5, ref6, ref7, ref27, ref28, ref29, ref107}. These approaches have evolved from regular expressions and rule-based methods to applying deep learning-based techniques. Historically, traditional techniques employed regular expressions \citep{ref25}, rule-based methods \citep{ref11, ref12, ref14, ref15, ref21, ref22, ref23, ref24}, or machine learning (ML) \citep{ref10, ref18, ref108, ref109, ref110} to find and extract abnormal events. Regular expressions-based approaches  often include formulating domain-specific regular expressions. These regular expressions are used to identify recently discovered vulnerabilities by comparing patterns. A primary drawback of these techniques is that they can overlook the latest anomalous events. Also, when each log entry appears normal, but the entire sequence is abnormal, no abnormal activity can be detected using regular expressions-based techniques. Rule-based methods often use predefined rules or signatures to detect abnormal patterns in log data. These principles are typically based on normal and abnormal system behavior observations. Data that conforms to these guidelines is classified as abnormal. The rule-based approaches can only recognize the predefined abnormalities. The ML-based log analysis process for anomaly detection comprises three primary steps: log parsing, feature extraction, and anomaly identification. Methods built on ML can not track the temporal information because they do not have a mechanism to remember past events. Besides, these approaches heavily impact the model output because they need human feature extraction from raw logs. In fact, all of the approaches mentioned above need extensive domain expertise. CoLog is a flexible approach that does not rely on domain expertise or predefined rules. CoLog exhibits a notable aptitude for extracting temporal information.

Outlier detection should be in real-time. Since deep neural networks are successful in real-time text analysis, numerous deep learning-based methods describe log data as a natural language sequence and effectively outlier finding \citep{ref107}. This started a trend of proposing advanced deep learning methods to identify log anomalies. These methods address the problems of approaches mentioned earlier.

Subsequent subsections outline the literature on deep learning for automated log-based anomaly detection. We then review the literature on deep MSA and compare our log anomaly detection approach to deep learning methods.

\subsection{Deep Learning for Anomaly Detection from Logs}\label{subsec2: deeplearningapproachesforanomalydetection}

The inception of deep learning-based methodologies can be traced back to the advent of DeepLog \citep{ref36}, which demonstrated commendable efficiency in detecting log data anomalies. Due to their rapid evolution over the past few years, these techniques now serve as the standard for modeling in this domain. Hence, the present subsection is most centered on models based on deep learning. In the following, we discuss unimodal and multimodal approaches.

\subsubsection{Unimodal Approaches}\label{subsubsec2: unimodalapproaches}

This subsection is dedicated to unimodal models for log anomaly detection despite the existence of various developed deep learning techniques for this purpose. These techniques' primary concept is derived from natural language processing (NLP). Furthermore, they frequently prioritize the sequential nature of log data. Deep learning different techniques, including MLPs \citep{ref40, ref42}, CNNs \citep{ref45, ref46, ref48, ref49}, RNNs \citep{ref38, ref51, ref52, ref53, ref55, ref56, ref57, ref59, ref62, ref63}, autoencoders \citep{ref67, ref69, ref70, ref71}, GANs \citep{ref51, ref72, ref73, ref74, ref75, ref76, ref77}, transformers \citep{ref26, ref78, ref79, ref80, ref81}, attention mechanisms \citep{ref37, ref40, ref86, ref88}, GNNs \citep{ref89}, and LLMs \citep{ref4, ref91, ref92}, are frequently employed in these approaches. Except \citep{ref67}, the unimodal approaches mentioned earlier can be divided into three main categories: sequence-based, semantic-based, and LLM-based. \citeauthor{ref67} proposed a log event anomaly detection method for large-scale networks that embeds diverse log data into hidden states using latent variables without requiring any preprocessing or feature extraction. The key idea is to first translate raw log messages into log time series for each log type, then map the log time series into latent variables per day and per log type using a conditional variational autoencoder. Finally, a clustering method is applied to the latent variables to detect deviations from the detected clusters, interpreted as anomalies.

\bmhead{Sequence-based unimodal approaches}

Sequence-based modeling \citep{ref26, ref37, ref40, ref42, ref45, ref46, ref48, ref49, ref53, ref55, ref56, ref59, ref62, ref63, ref69, ref70, ref71, ref72, ref73, ref74, ref76, ref77, ref78, ref80, ref86, ref88, ref89} is widely recognized as the dominant network architecture in log-based anomaly detection. Typically, these methods interpret the log data as a sequence of natural language. For example, \citeauthor{ref45} extracted log keys from raw logs and assigned them unique numerical values. Shorter vectors are padded with zeros, while longer vectors are truncated to ensure that each session of log keys has the same length. The logs in vectorized form are inputted into an embedding layer, followed by three 1D convolutional layers with varying filter sizes. After that, max-pooling, dropout, and a fully connected softmax layer are applied to classify anomalies. \citeauthor{ref62} demonstrated that log data is unstable due to the evolution of logging statements and noise in log data processing. They proposed LogRobust to tackle this issue by extracting semantic feature vectors from log event sequences. Then, an attention-based bidirectional long short-term Memory (BiLSTM) model is proposed to automatically identify abnormal instances and learn about the value of various log events. LogSpy \citep{ref40} is a method for detecting anomalies in distributed systems. It uses a technique called frequent template tree (FT-Tree) to extract templates and the skip-gram model to construct feature vectors. It employs a CNN architecture and an attention mechanism to analyze the relationships between log templates and improve the efficiency of real-time log anomaly detection. LogBERT \citep{ref78} uses the transformer encoder to model log sequences for self-supervised log anomaly detection. The training process involves two distinct tasks: (1) masked log key prediction to accurately identify randomly masked log keys in normal sequences and (2) volume of hypersphere minimization to bring normal log sequences close in the embedding space. LogBERT encodes normal log sequence patterns and detects anomalies after training. LogGAN \citep{ref73} uses a log-level GAN with permutation event modeling. A log parser converts unstructured system logs into structured events. Permutation event modeling minimizes long short-term memory (LSTM)'s sequential dependency concerns, allowing it to handle out-of-order log arrivals. The adversarial learning system uses a generator to generate synthetic logs and a discriminator to verify them. DeepCASE \citep{ref86} can acquire knowledge regarding the correlation among log events in sequences. Subsequently, an interpreter arranges similar events into similar clusters to detect abnormalities. The system administrator receives reports of abnormality samples, and the administrator's decisions are learned by DeepCASE and applied to future occurrences of identical sequences. \citeauthor{ref71} proposed AutoLog as a semi-supervised deep autoencoder. Its functionality is not contingent upon the concept of log line sequencing. The approach employed involves the extraction of numeric score vectors to handle heterogeneous logs. During the process, AutoLog does not incorporate any application-specific knowledge and refrains from making any presumptions regarding the format and order of the underlying lines within the logs. \citeauthor{ref77} proposed a log-based anomaly detection approach known as Adanomaly. This method employs the bidirectional GAN model for extracting features and a technique for classification based on ensemble learning to detect anomalies. Adanomaly calculates the reconstruction loss and discriminative loss as features with the help of bidirectional GAN. LogEncoder \citep{ref42} is a framework that uses labeled and unlabeled data to detect anomalies in system logs. The system comprises three primary elements: Log2Emb, Emb2Rep, and Anomaly Detection. The Log2Emb procedure transforms discrete log events into semantic vectors. The Emb2Rep procedure employs an attention-based LSTM model to differentiate between normal and abnormal sequences. The framework is adaptable for both offline and online detection. \citeauthor{ref59} proposed LayerLog, an innovative system for detecting anomalies in log sequences. LayerLog utilizes a hierarchical structure called "Word-Log-Log Sequence" to do this. The system comprises three layers: Word, Log, and LogSeq. It extracts semantic information from raw logs without any loss. Empirical assessments demonstrate the efficacy of LayerLog in detecting abnormalities. LogFormer \citep{ref37} is a pipeline consisting of two stages: pre-training and tuning. It is designed to detect log anomalies and enhance generalization across different domains. It extracts semantic information of log sequences and trains a transformer-based model on a specific domain. By employing adapter-based tuning, the training costs are minimized by keeping most parameters fixed and only updating a small number of them. FastLogAD \citep{ref76} is an unsupervised technique for detecting anomalies in log data. It utilizes a mask-guided anomaly generator (MGAG) and a discriminative abnormality separation (DAS) model. The MGAG algorithm produces simulated anomalies based on normal log data, while the DAS model distinguishes between normal log sequences and simulated ones, allowing for effective real-time detection. MLAD \citep{ref80} is a log anomaly detection approach that utilizes the transformer and gaussian mixture model (GMM) to detect anomalies across many systems. MLAD employs SBERT \citep{ref111} to transform log sequences into semantic vectors. It utilizes a sparse self-attention mechanism to capture word-level dependencies and a GMM to emphasize the uncertainty associated with infrequent words in the "identical shortcut" problem. \citeauthor{zhang2025log} proposed E-Log, which provides fine-grained elastic anomaly detection and diagnosis for databases. \citeauthor{qiu2025loganomex} introduced LogAnomEX, an unsupervised log anomaly detection method based on the Electra model and gated bilinear neural networks. \citeauthor{qiu2025fedaware} proposed FedAware, a distributed IoT intrusion detection method that leverages fractal shrinking autoencoders. These methods highlight both supervised and unsupervised perspectives, reinforcing the importance of comparative baselines.

\bmhead{Semantic-based unimodal approaches}

Semantic-based unimodal log anomaly detection is limitedly used \citep{ref38, ref51, ref52, ref57, ref75, ref79, ref81}. Several semantic-based approaches utilize log lines to extract semantic information and build models using unsupervised or semi-supervised approaches \citep{ref75, ref79, ref81}. Other techniques incorporate sentiment analysis (SA) to detect log anomalies \citep{ref38, ref51, ref52, ref57}. These methods utilize the polarity of log messages as a basis for learning and predicting normal and abnormal log events. SA-based log anomaly detection methods classify normal occurrences as positive sentiments and anomalies as negative. For example, \citeauthor{ref52} proposed an approach for detecting and classifying anomalies in log messages by employing three different deep learning architectures: Auto-LSTM, Auto-BiLSTM, and Auto-gated recurrent unit (GRU). The methodology consists of two primary phases: feature extraction with autoencoders separately for positive and negative classes and concatenating outputs to provide a single-label dataset for training an LSTM, BiLSTM, and GRU model. The classification model is trained using the softmax activation function, categorical cross-entropy loss, and the Adam optimizer \citep{ref112}. \citeauthor{ref51} also proposed a log message anomaly detection method using a combination of GAN, autoencoder, and GRU networks. It generates synthetic negative log messages using a proposed SeqGAN to oversample negative log messages and balance the dataset. The autoencoder extracts features from positive and oversampled negative log messages, with separate networks for each class. A GRU network is used for anomaly detection and classification, with improved accuracy compared to models without over-sampling. Logsy \citep{ref79} is a classification-based method for detecting anomalies in unstructured log data. It uses a self-attentive encoder and a hyperspherical loss function to learn compact log vector representations. It uses auxiliary log datasets from other systems to improve normal log representation and prevent overfitting. The learned representations help distinguish normal and abnormal logs using the anomaly score as a distance from the center of a hypersphere. \citeauthor{ref38} proposed pylogsemtiment, a SA-based method for identifying anomalies in production OS log files. This approach uses positive and negative sentiment analysis, with negative sentiment detection similar to detecting anomalies. They used deep learning to predict unseen data and implemented a GRU network to determine sentiments. They considered class imbalance due to the lower frequency of negative messages in real-world logs. They proposed using the Tomek link technique to balance the two sentiment types. SentiLog is an other approach proposed by \citeauthor{ref57} to analyze parallel file system logs and detect anomalies. SentiLog uses a collection of parallel file systems' logging-related source code to train a generic sentimental natural language model. In this manner, SentiLog learns the implicit semantic information based on LSTM developers have embedded within the parallel file system. Like pylogsentiment, SentiLog shows that SA is a promising approach for modeling unstructured and difficult-to-decode system logs. A2Log \citep{ref81} is a two-step, unsupervised anomaly identification method for log data that includes anomaly scoring and decision-making. Initially, a self-attention neural network is employed to assign a score to each log message. Furthermore, it sets the decision boundary by augmenting the existing normal training data without using any anomalous cases. LogELECTRA \citep{ref75} is a self-supervised anomaly detection model designed for unstructured system logs. It examines individual log messages without the need for a log parser. The system acquires an understanding of the context of normal log messages by training a discriminator model to identify replaced tokens in changed sequences. During the inference process, LogELECTRA computes an anomaly score for every log message, identifying anomalies as individual anomalies in real-time without needing time-series analysis.

\bmhead{LLM-based unimodal approaches}

After demonstrating LLMs' unique and emerging abilities in performing various tasks \citep{ref113}, recent studies have proposed using LLMs to detect anomalies from log data. For example, LogFiT \citep{ref4} is an innovative approach for detecting anomalies in system logs. However, it does not rely on predetermined templates or labeled data. Instead, it utilizes a language model based on bidirectional encoder representations from transformers (BERT) to analyze the logs. The system acquires knowledge of language patterns by making predictions on masked sentences using normal log data. LogFiT evaluates the accuracy of predicting the top-k tokens during the inference process to determine a threshold for detecting abnormalities. LogFiT is specifically engineered for easy integration into pre-existing NLP frameworks. LogPrompt \citep{ref92} leverages LLMs to automate log analysis in online scenarios. The system tackles the challenges associated with the restricted ability to understand log anomaly detection results and the ability to handle unseen logs. LogPrompt employs three advanced prompt strategies: self-prompt, chain-of-thought prompt, and in-context prompt. LogPrompt allows for the efficient analysis of log files and identifying abnormal patterns without requiring a large amount of training data. Human assessments demonstrate that LogPrompt's interpretations are valuable and easy to understand, assisting professionals in understanding log data. \citeauthor{ref91} proposed RAGLog to detect anomalies in log data. RAGLog leverages the retrieval augmented generative (RAG) model with a vector database and an LLM. It avoids the necessity of interpreting logs by directly consuming unprocessed log data. The LLM conducts semantic analysis to compare normal log entries with the query entry, enabling RAGLog to function as a zero-shot classifier that does not necessitate abnormal log samples for training.

\subsubsection{Multimodal Approaches}\label{subsubsec2: multimodalapproaches}

This section specifically examines models that utilize multimodal analysis to find anomalies in log data. These approaches are relatively infrequent in comparison to unimodal methods in log analysis. Deep learning employs various techniques, such as MLPs \citep{ref41, ref43, ref44}, CNNs \citep{ref47, ref50}, RNNs \citep{ref36, ref54, ref58, ref60, ref61, ref64, ref65, ref66}, autoencoders \citep{ref41, ref68}, GANs \citep{ref41}, transformers \citep{ref82, ref83, ref84}, attention mechanisms \citep{ref47, ref85, ref87}, and EGNNs \citep{ref90}, which are commonly used in these approaches. The three main categories of multimodal techniques mentioned above are early \citep{ref50, ref58, ref60, ref61, ref82, ref84, ref85, ref87, ref90}, intermediate \citep{ref41, ref44, ref47, ref54, ref64, ref65, ref66}, and late \citep{ref43} fusion-based methods. The exceptions to this rule are \citep{ref36, ref68, ref83}. DeepLog \citep{ref36} is an early example of using deep learning for log anomaly detection. It works by analyzing the sequential relationships between log entries, treating them as a sequence of natural language. Next, learn about the normal log patterns, consisting of a log key and a parameter value vector representing each log entry. DeepLog utilizes two LSTM models to learn about log key sequences and parameter value vectors. In addition, LSTMs are used to forecast future log events. These predictions are then compared to the ground truth to identify any anomalies. \citeauthor{ref68} proposed VeLog, a method for detecting anomalies in distributed systems using variational autoencoders (VAEs). The process consists of two distinct stages: offline training and online detection. During the offline phase, VeLog gathers log data, analyzes it, and produces the log order matrices and the log count vector matrices representing log executions. The VAE model acquires knowledge about normal sequence patterns based on these matrices. VeLog analyzes and extracts features from new logs during the online phase, generating comparable matrices. Anomalies are identified by comparing the resulting matrices with previously learned patterns. This approach is highly effective when applied to large-scale distributed systems. \citeauthor{ref83} proposed UMFLog, a novel unsupervised model for detecting anomalies in log data. UMFLog employs two sub-models to capture the logs' semantic and statistical features, effectively. The first model uses BERT to extract semantic features from log content. In contrast, the second model employs a VAE to learn statistical features, specifically focusing on word frequency in logs. This dual-feature approach improves the model's capacity to identify anomalies without needing labeled data, making it well-suited for practical applications.

\bmhead{Early fusion-based multimodal approaches}

Typically, these approaches include extracting meaningful feature vectors from various aspects of log data and constructing a supervector. Finally, they utilize this supervector as the source of knowledge to train their model. For example, \citeauthor{ref61} add an attention layer following the BiLSTM to assess the value of prior events in predicting subsequent events. They merge log data's sequential and quantitative features with semantic features to achieve effective results. After the sequence, quantitative, and semantic feature vectors have been extracted, they are combined and fed into the model to understand the relationships between the log events. This framework can extract the underlying dependencies from event logs. LogST \citep{ref50} is a log anomaly detection method that combines semantic and topic features extracted from logs. The process involves utilizing BERT and singular value decomposition (SVD) to extract semantic features. Furthermore, latent dirichlet allocation (LDA) extracts topic features. The combined features are subsequently fed into the improved TCN model, which utilizes weighted residual connections for anomaly detection. \citeauthor{ref84} proposed Log-MatchNet, a method for detecting anomalies in log data designed to handle unstructured and imbalanced log data. It employs log parsing to transform log content information into vectors, utilizing the BERT model for universal feature extraction. The approach involves a matching network to learn similarity scores between input and prototype vectors. The prototype vectors represent a small number of labeled abnormal logs. These vectors are utilized to make generalizations about unknown log samples in a few-shot scenario. \citeauthor{ref58} proposed LogMS, a log anomaly detection technique that leverages many sources of information and employs probability label estimation. The system utilizes an multi-source information fusion-based-LSTM network to process semantic, sequential, and quantitative information and a probability label estimation-based gate recurrent unit network to handle pseudolabeled data. The second stage turns on when the abnormalities are not identified in the first stage to optimize efficiency. LogMS efficiently identifies anomalies, even when the log data changes over time. \citeauthor{ref85} proposed MultiLog, a multivariate log-based anomaly detection approach for distributed databases. MultiLog initially processes sequential, quantitative, and semantic information from the logs of each database node. This is done using an LSTM model paired with a self-attention in the standalone estimation module. The system utilizes an autoencoder with a meta-classifier to normalize node probabilities and detect abnormalities throughout the entire cluster.

\bmhead{Intermediate fusion-based multimodal approaches}

These approaches extract features and process each modality independently until a certain point, at which point the feature maps are concatenated before categorization or decision-making. For example, \citeauthor{ref54} found these challenges unmet by prior methods: (1) actively developing and maintaining software systems change log formats, (2) the trivial monitoring tools may miss the underlying reasons for performance issues. Thus, SwissLog, a robust deep learning-based model for log analysis, is proposed. SwissLog utilizes BERT to embed the semantic information onto the embedding vector and puts the temporal information into another embedding vector. SwissLog feeds to an attention-based BiLSTM model, an intermediate fusion of semantic and time embedding vectors to learn the distinctions between normal and abnormal log sequences. \citeauthor{ref47} proposed LogAttn, a method using an autoencoder for unsupervised log anomaly detection. LogAttn begins by parsing log data into structured representations. It generates event count and semantic vector sequences. TCN and deep neural network encoders learn temporal and statistical correlations in log data separately. The decoder reconstructs the log sequence using a latent representation formed by an attention mechanism that weighs feature importance. Comparing the reconstruction error against a predetermined threshold using a model trained entirely on normal log data provides anomaly detection. \citeauthor{ref66} proposed FSMFLog, an approach for detecting log anomalies by employing multi-feature fusion. FSMFLog overcomes the constraints of previous log parsing techniques by utilizing a prefix tree structure to extract semantic data in word lists rather than relying on typical log templates that frequently fail to capture critical semantics. It commences with preprocessing by removing variables from logs and subsequently grouping log sentences utilizing heuristic strategies. FSMFLog utilizes a bidirectional GRU model that is improved using an attention mechanism. This model utilizes semantic, time, and type features. \citeauthor{ref65} proposed a log anomaly detection method integrating a self-attention mechanism with a bidirectional GRU model. It utilizes two bidirectional GRU models: one is responsible for processing the log template sequence, while the other one is dedicated to handling the log template frequency vector. The results from both models are concatenated and fed into a self-attention layer, which fuses the features to predict the probability distribution of the subsequent log template. This method detects anomalies when the expected template is missing from the candidate set. \citeauthor{ref64} proposed a robust log-based anomaly detection framework named WDLog, which integrates wide and deep learning to address challenges in detecting anomalies in evolving log data. The method begins with an optimized log template extraction process that enhances the traditional Drain algorithm by incorporating semantic embedding and clustering, effectively reducing sensitivity to change in log wording. Following this, WDLog extracts three types of features from the generated log templates: temporal features, invariant features, and statistical features. Anomaly detection is then performed using a combination of a GRU model with attention mechanisms for temporal features and a gradient boosting decision tree model for invariant and statistical features.

\bmhead{Late fusion-based multimodal approaches}

Under these methods, the fusion occurs after each modality has been processed and classified, independently. The results from each modality are then combined to make the final decision. For example, \citeauthor{ref43} proposed an approach for detecting web scanning behaviors in online web logs, which are crucial for identifying potential cyber-attacks. It utilizes three lightweight classifiers that analyze distinct features extracted from web traffic logs such as HTTP textual content, status codes, and request frequency. This approach categorizes logs on the source and destination IPs and evaluates the characteristics of these logs to differentiate between scanning and normal behaviors. Each classifier processes specific features. Therefore, textual features are analyzed using an MLP, while HTTP status codes and request frequencies are assessed through support vector machines (SVMs). The classifiers generate probabilities indicating scanning behavior, which are combined using a decision strategy to determine the final classification outcome.

\subsection{Deep Learning for Multimodal Sentiment Analysis}\label{subsec2: deeplearningapproachesformultimodalsentimentanalysis}

Traditional sentiment analysis procedures often focus on extracting user sentiment from a single modality. MSA incorporates visual and audio modalities into text-based SA. Audio and visual characteristics are employed as they provide a superior ability to better explain or depict sentiment compared to a lengthy list of words. Many studies \citep{ref93, ref94, ref95, ref96, ref97, ref98, ref99, ref100, ref101, ref102} have been undertaken in this field, most depending on deep learning concepts. The increasing prevalence of deep learning can be attributed to its ability to boost complex processing, thereby improving the automation process. From a multimodal learning standpoint, comprehensive surveys \citep{al2025overview, liang2024foundations} have outlined foundational principles, challenges, and open questions in multimodal machine learning. These works emphasize systematic integration, balancing, and interaction of modalities - all central issues addressed in CoLog. Moreover, \citeauthor{qiu2024multimodal} proposed a chained interactive attention mechanism for multimodal sentiment analysis, further validating that multimodal fusion strategies developed in the sentiment domain can inspire advances in anomaly detection. \citeauthor{ref100} proposed a transformer encoder architecture that fuses any information about modalities. The model at hand utilizes joint-encoding to concurrently encode each modality, with modular co-attention controlling how a modality attends to itself. \citeauthor{ref98} proposed a framework called UniMSE. The UniMSE framework is a novel method for unified multimodal sentiment analysis and emotion recognition in conversations. It reformulates tasks into a generative format, integrates acoustic, visual, and textual features, and employs inter-modal contrastive learning to minimize intra-class variance and maximize inter-class variance. \citeauthor{ref97} proposed a multimodal sentiment analysis model using transformer architecture and soft mapping. The transformer layer utilizes the attention layer to map modalities, while the soft mapping layer uses stacking modules for multimodal information fusion. This model addresses data interaction issues in multimodal sentiment analysis by considering the relationships between multiple modalities. The TETFN \citep{ref96} is a method for multimodal sentiment analysis that enhances video sentiment recognition by integrating textual, visual, and audio modalities. It uses preprocessing, feature extraction, LSTM networks, and TCNs to encode contextual information and generate unimodal labels. The core of TETFN is a text-enhanced transformer module that leverages a text-oriented multi-head attention (MHA) mechanism to incorporate textual information into the visual and audio modalities, facilitating effective pairwise cross-modal mappings. \citeauthor{ref95} proposed a novel multimodal sentiment analysis method for images containing textual information. Initially, it employed a recognition system to extract text from images using the Google Cloud Vision API, which is noted for its high accuracy in optical character recognition (OCR). The extracted text is preprocessed and analyzed with a RoBERTa-based model for textual sentiment analysis. Concurrently, visual sentiment analysis uses a transfer learning approach to capture visual features. The method integrates the outputs from both analyses through a weighted fusion strategy, enhancing the overall sentiment classification accuracy. \citeauthor{ref93} proposed a robust multimodal sentiment analysis model that uses a modality translation-based approach to handle uncertain missing modalities. It involves translating visual and audio data to text modality, encoding text, and fused into missing joint features (MJFs). The transformer encoder module then encodes these MJFs to learning long-term dependencies between modalities. Therefore, the sentiment classification is based on the transformer encoder module's outputs.

\subsection{Beyond Log-Specific Approaches}\label{subsec2: beyondlog}
Beyond log-specific approaches, multimodal anomaly detection has been explored in other domains such as video understanding. For instance, \citeauthor{SU2025110898} a proposed semantic-driven dual consistency learning for video anomaly detection that leverages semantic-driven representations to align appearance and motion features, enhancing anomaly localization under weak supervision. Similarly, \citeauthor{SU2025103256} proposed federated weakly-supervised video anomaly detection with a mixture of local-to-global experts that employs mixtures of experts within a federated learning framework to address distributed and heterogeneous data scenarios.

While these methods target video modalities, they share conceptual parallels with CoLog in terms of semantic-guided fusion and cross-modal consistency learning. Inspired by such principles, CoLog's collaborative transformer integrates semantic and sequential log modalities through impressed attention and the Modality Adaptation Layer (MAL), enabling unified detection of both point and collective anomalies.

\subsection{Our Method: A Cut Above the Rest}\label{subsec2: ourmethod}

\begin{figure}[h]
	\centering
	\includegraphics[width=1\textwidth]{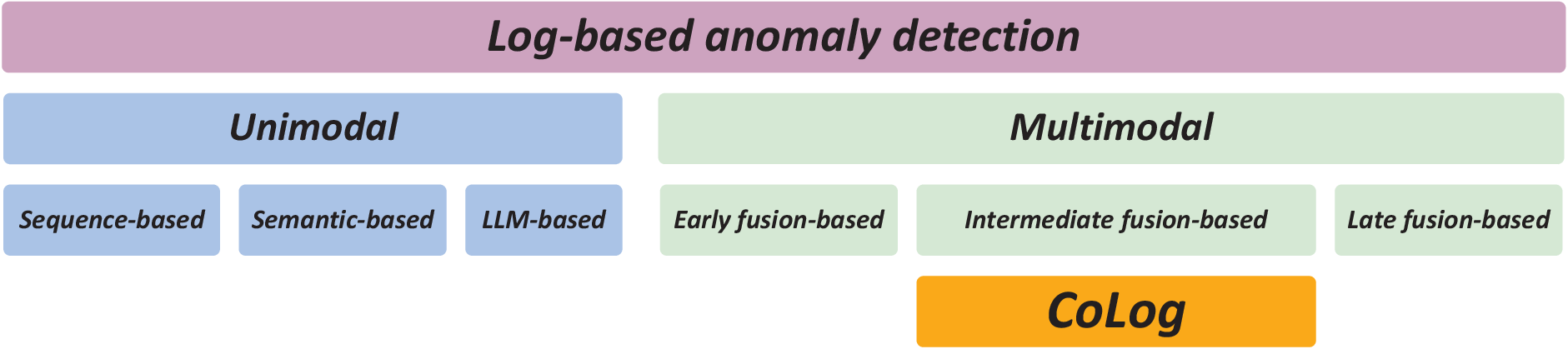}
	\caption{A comprehensive overview of the existing literature dealing with the approaches for identifying anomalies in log data.}\label{fig2}
\end{figure}

According to the literature cited in Section~\ref{subsec2: deeplearningapproachesforanomalydetection}, diverse deep learning techniques are progressively being employed to detect anomalies in log records, which we classified into two general categories and six subcategories. Figure~\ref{fig2} demonstrates a concise overview of log-based anomaly detection. CoLog falls under the umbrella of multimodal-based approaches.

However, our method differs as we aim to use the redundancy of modalities within system logs based on MSA and CT. CT can be utilized to supplement diverse modalities. When data are deficient from a particular modality, data from another modality is employed. Therefore, these transformers enable the extraction of various information from modalities and interactively encode modalities.

The strong performance and generalization of sentiment-based log analysis models, as claimed by \citep{ref38, ref51, ref52, ref57}, makes well-trained models more robust to the evolution of log statements. We offer the CoLog framework that developed based on the SA of logs. But, unlike \citep{ref38, ref51, ref52, ref57}, our model employs a more complex architecture for MSA-based anomaly detection. By evaluating the sentiment of log events in dependence on their background, CoLog can analyze both point and collective anomalies. Moreover, CoLog can distinguish similar events by utilizing auxiliary information from the background of their occurrence, as depicted in Figure~\ref{fig1}. In the following sections, we will demonstrate that the context of event logs can also be utilized.

In contrast to reviewed methods, CoLog relies on modality interaction within the logs. It does not concat different modalities into a super vector in any way. The challenge at hand is tackled through impressed attention based on CT. This form of attention is beneficial to discovering the correlation between two modalities. Since deep learning models are commonly denoted as black-box models due to their complex nature, which surpasses human comprehension, comprehending the reasoning behind each decision is a task that exceeds human cognitive capacity. The utilization of the impressed attention mechanism has the potential to augment the interpretability of our log anomaly detection model. This is because the attention scores, which are computed based on the impressed attention, can offer valuable insights into the model's cross-modal decision-making process. CoLog projects the results of modalities into a latent space above the transformer blocks for intermediate fusion and executes the encoding process, concurrently. Therefore, forming an optimal latent space is crucial due to the inherent incompatibility of various modalities.

Furthermore, the CoLog architecture enables the identification of point and collective log abnormalities using a unified framework. According to our current knowledge, CoLog is the first approach that performs this task.

\section{Preliminaries}\label{sec3: preliminaries}

This section discusses the task's definition and the rationale for employing the transformer model to accomplish it. It also covers the assumptions and threat model used to evaluate potential threats.

\subsection{Task Definition}\label{subsec3: taskdefinition}

 This task aims to conduct a log analysis to detect anomalies during the system's execution. The scope of the analysis will include analyzing logs' information to detect abnormalities. A machine learning algorithm will be used to analyze these data. The performance of the anomaly detection method will be evaluated by measuring some metrics, such as accuracy, precision, recall, and F1-score, to ensure that the method does not generate too many false alerts.

\begin{figure}[h]
	\centering
	\includegraphics[width=1\textwidth]{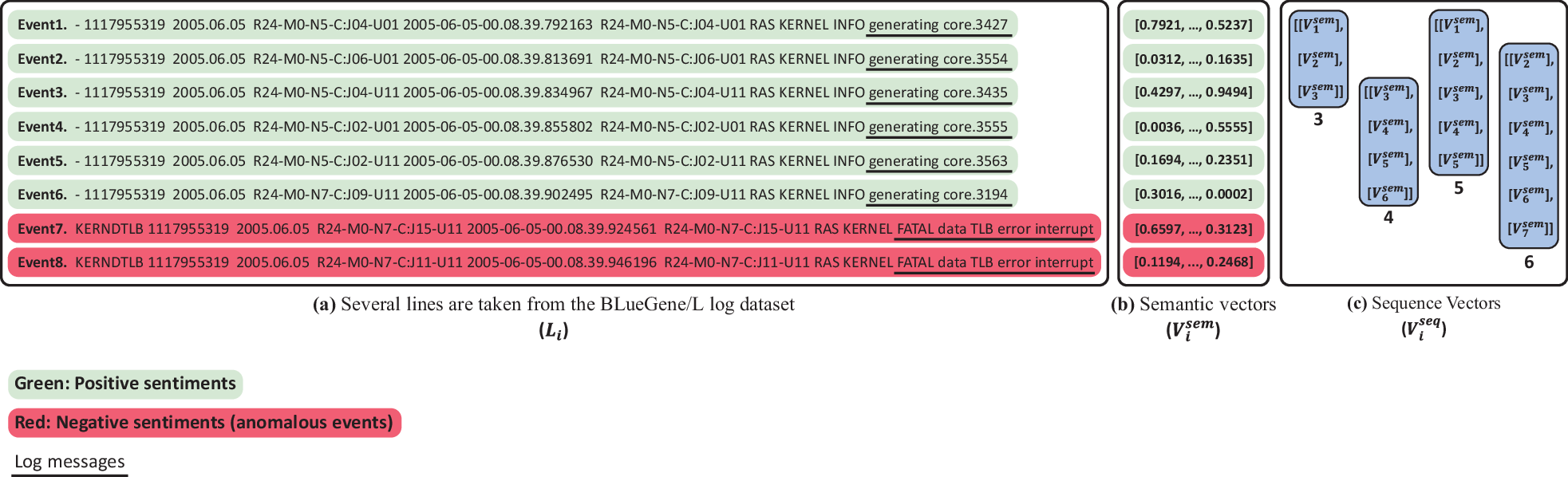}
	\caption{\textbf{(a)} Several lines extracted from the BLueGene/L log dataset.; \textbf{(b)} Semantic modality: The semantic modality is constructed using the extracted semantic vectors from log events' messages.; \textbf{(c)} Sequence modality: The construction of the sequence modality involves appending semantic vectors into sequence vectors based on window sizes of 3, 4, 5, and 6.}\label{fig3}
\end{figure}

When conducting multimodal sentiment analysis, researchers look at the dialogue from various perspectives (modalities) to determine the participants' sentimental states. The three most commonly utilized modalities are textual, audio, and visual. In the same vein, logs possess a nature that, akin to human conversations, conveys information from different aspects, such as statistical, temporal, semantic, and information based on the event's sequence. However, the semantic and sequence modalities in logs are the most crucial ones to consider. Besides, to detect anomalies from logs, we can implement sentimental pipelines to classify sentiments wherein recognizing negative sentiments alerts anomalous situations. In Figure~\ref{fig3}\textcolor{blue}{(a)}
, log messages' negative and positive sentimental states are presented in  red and green, respectively. Based on these observations, we propose to apply multimodal sentiment analysis to the log anomaly detection task. Figure~\ref{fig3}\textcolor{blue}{(b)} and Figure~\ref{fig3}\textcolor{blue}{(c)} respectively depict the semantic and sequence modalities. These figures show that semantic vectors of log events are derived based on log messages. These vectors generate sequence vectors based on the window size. In this case, the window size determines the length of the sequence vectors we create. These vectors would serve as input for a transformer-based multimodal framework that learns cross-modal dependencies and predicts point and collective log anomalies.

\subsection{Transformers: why?}\label{subsec3: transformers:why?}

Transformers exhibit a notable advantage in their processing speed when dealing with a sequence. They focus more on the crucial components, resulting in enhanced speed compared to alternative models. Also, transformers exhibit multiple benefits in the context of multimodal sentiment analysis. Transformers can enhance models' robustness in multimodal sentiment analysis, thereby constituting an additional benefit. Another advantage is their ability to incorporate impressed attention into their architecture to extract cross-modal interactions within non-aligned multimodal data. The significance of this matter lies in the complexity and multimodality of most real-world information. Considering transformers' benefits, we decided to base our multimodal model around them.

\subsection{Assumptions}\label{subsec3: assumptions}

Anomaly detection from the log strongly depends on the quality of the log. Multiple categories of assaults exist that may not be recorded in system logs. Suppose certain auditable occurrences, such as login attempts, are not duly recorded. In that case, it is plausible that CoLog may not locate such forms of abnormalities. Another type of attack that may not be recorded in logs is a log forging or log injection attack. The aforementioned is a man-in-the-middle assault in which an unauthenticated user interposes between the application and the server. Therefore, it is assumed that any attempt by an adversary to modify the system logging behavior is not feasible. In a manner analogous to that of pylogsentiment, we consider log entries like \textit{\textbf{[c010ce54$>$] mtrr wrmsr+0xf/0x2e}} in a kernel log as having a positive sentiment during training if they do not provide a human-readable log message. It is essential to clarify that our analogy between anomalies and "negative sentiments" does not imply a direct mapping to human emotions. Instead, it serves as an analytical abstraction: anomalies are deviations from expected behavior (negative polarity), while normal and benign events align with expected states (positive polarity). Unlike human sentiment, which emerges from subjective perception, log polarity is determined contextually by semantic embeddings and sequence patterns. For instance, a "restart" event can be benign in maintenance scenarios but anomalous when it occurs unexpectedly.

\subsection{Threat Model}\label{subsec3: threatmodel}

An anomaly is a data instance that diverges from the normal pattern. Various abnormalities exist, including point, collective, and contextual anomalies. A point anomaly refers to a singular data point that exhibits a significant deviation from the overall pattern of the dataset. A group of data instances can form an anomalous pattern known as a collective anomaly. A contextual anomaly is when a given data instance exhibits abnormal behavior only in a particular context while classifying normal in others. A point or collective anomaly can be classified as contextual when examined in a specific context. CoLog identifies point and collective anomalies through a unified framework based on CT. This involves evaluating each log entry for its sentiment based on the background or context in which it occurred and flagging any entry that exhibits negative sentiment as an anomaly.

\section{Method}\label{sec4: method}

This section explains architecture and data flow of CoLog, the specific procedures CoLog uses to gather and examine log data to look for anomalies. It includes the conditions under which data were collected, labeled and preprocessed, guidelines that specify how anomaly detection task should be carried out, and DL methodologies and techniques utilized for automatic log-based anomaly detection based on MSA.

\subsection{Overall Architecture}\label{subsec4: overallarchitecture}

\begin{figure}[h]
	\centering
	\includegraphics[width=1\textwidth]{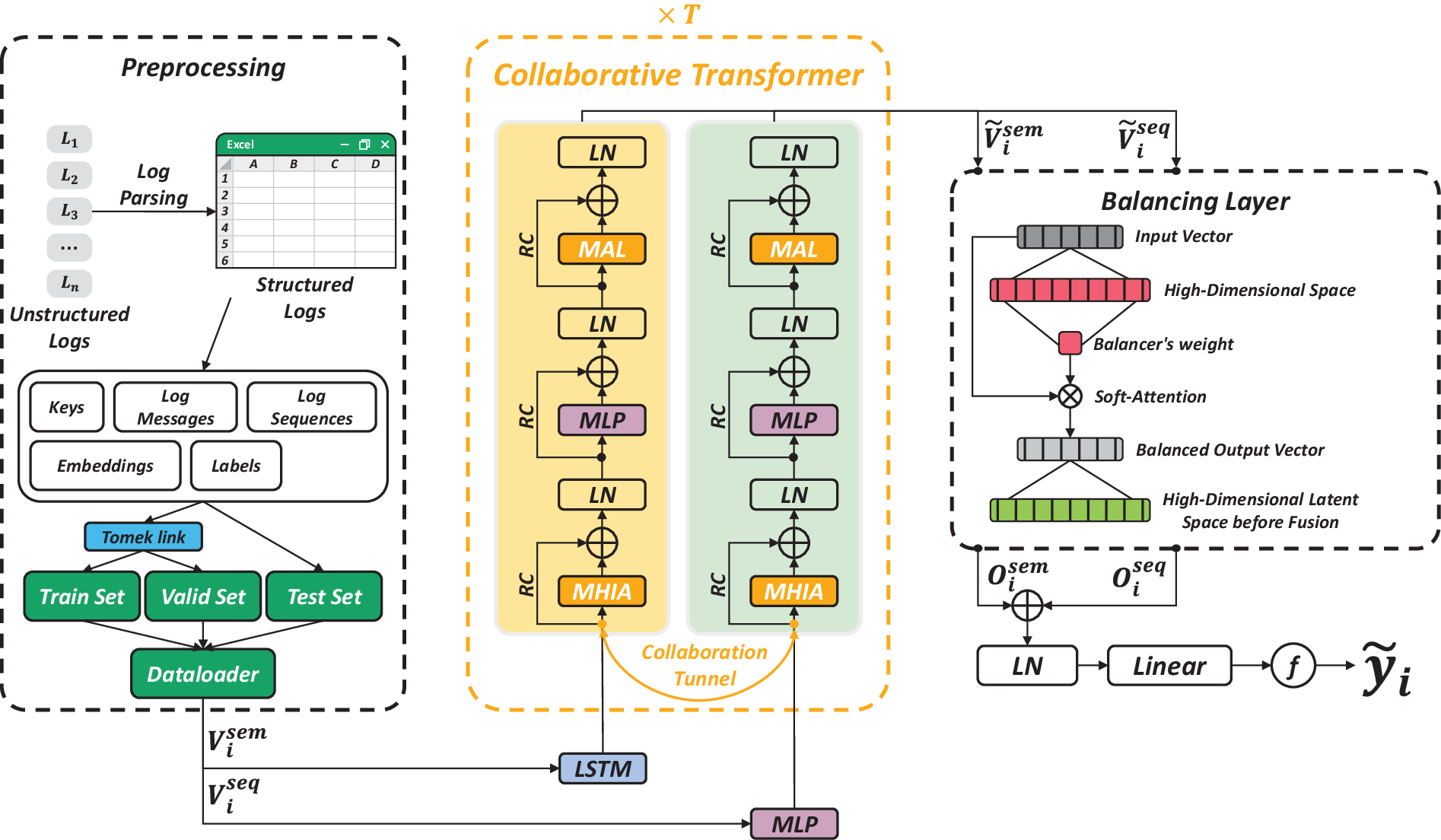}
	\caption{The overview of CoLog. Light green and gold colors demonstrate modality encoders. Each encoder in the collaborative transformer consists of MHIA, MLP, MAL, and LNs. MHIA and MAL are multi-head impressed attention and modality adaptation layer modules, respectively. The preprocess layer transforms unstructured logs into easily understandable data for the model. The purpose of the balancing layer is to regulate the influences of different modalities when calculating the final results.}\label{fig4}
\end{figure}

Figure~\ref{fig4} depicts CoLog's training framework and gives an overview of its architecture. Like most log anomaly detection techniques, CoLog begins by preprocessing raw logs. First, it extracts semantic and sequence modalities from the raw-valued logs. Before feeding the embedded vectors of modalities into the model, the Tomek link approach addresses the class imbalance issue.

Two transformer blocks called collaborative transformers with a modified GA mechanism called impressed attention are applied to these modalities. CoLog comprises a stack of $T$ identical CT blocks, but each one has a different set of training parameters. Each modality encoder comprises a multi-head impressed attention (MHIA), MLP, modality adaptation layer  (MAL), and Layer Normalizations (LNs). Modalities fed into the matching modality encoder in CT. The MLP layer, which employs a nonlinear activation function, receives the results of computing the attention scores from the attention layer. Each encoder has a residual connection and LN after the MHIA, MLP, and MAL layers. The results of the encoding blocks are fed into the balancing layer (BL) to determine the sentiment of a log message, where soft attention and fusion are performed. Remember that the negative sentiment is connected to any potential abnormal behaviors.

Due to the information sharing between different modalities in CT, the suggested architecture thus conveniently learns the shared representation between modalities.

\subsection{Preprocessing}\label{subsec4: preprocessing}

In the preprocessing phase, the log parsing procedure extracts a specific element, denoted as a log message, from a log file. Then, every log message is segmented into tokens and transformed into lowercase. The process involves constructing a lexicon based on the training dataset, resulting in a dictionary comprising $V_{size}$ distinct terms. After that, each word is embedded in a 300-dimensional vector. In cases where a term from the validation or testing dataset is not present in our extracted vocabulary, we substitute it with the $"UNK"$ token. In addition, irregularities exist in the length of individual log messages. Consequently, if the log message' length falls below the specified threshold, it is padded with zeroes. Conversely, in cases where the length exceeds the established threshold, we truncate to abbreviate it to the predefined message length.

\subsubsection{Log Parsing}\label{subsubsec4: logparsing}

Each log entry contains various separate parts. Log parsing is the act of transforming unstructured log data into a structured format to enable machine interpretation. Every log file is parsed to convert it from an unstructured form to a structured format with entities like timestamp, service name, level information, and log message. We must separate the log message from the log entry because it is the only entity with the sentiment. In addition, each log message is transformed into 384-dimensional vectors using SBERT and stored separately from the log messages. We employ the nerlogparser \citep{ref114} and Drain \citep{ref115} log parsers in our implementation. Indeed, each dataset is processed and parsed using its corresponding parser.

\subsubsection{Log Parser: Drain}\label{subsubsec4: logparserdrain}

One of the representative algorithms for log parsing is the Drain. Drain employs a fixed-depth parse tree, which encodes specifically created parsing rules, to expedite the parsing procedure. Drain preprocesses logs first using regular expressions and user-defined domain knowledge. Second, Drain begins with the preprocessed log message at the root node of the parse tree. The parse tree's first layer nodes represent log groups whose log messages are of various lengths. Drain uses a concept of distance called token similarity to determine how similar the log messages are to one another. The tokens in the first positions of the log message are used to choose the subsequent internal node. Drain then determines whether to add the log message to an existing log group by calculating the similarity between each log group's log message and log event. Drain is quite effective and accurate. It can also parse log data in real-time, which makes it well-suited for log anomaly detection. Additionally, Drain is scalable, making it suitable for large-scale environments.

\subsubsection{Log Parser: nerlogparser}\label{subsubsec4: logparsernerlogparser}

The nerlogparser utilizes named entity recognition (NER), a technique employed to extract named entities from text. Nerlogparser identifies named entities in log files as words or phrases that include common fields seen in a log entry, such as timestamp, hostname, or service name. Named entity extraction is the process of recognizing each element in a log entry. To perform NER, nerlogparser employs BiLSTM. The primary advantage of the nerlogparser is its ability to automatically parse log data using a pre-trained model. Thus, there is no necessity to establish any rules or regular expressions. The nerlogparser can parse a wide range of log files due to its training on multiple log types.

\subsection{Task Formulation}\label{subsec4: taskformulation}

We employ the notation $L_{i}^{m}, {m}\in{\{sem,\ seq\}}$ to represent the unimodal raw-valued log modalities extracted from the log messages $\{L_{i-w},\ L_{i-(w-1)},\ ...,\ L_{i-1},\ L_{i}\}$, where $w$ is the window size and the notation $\{sem,\ seq\}$ specifies the two different types of modalities: semantic and sequence. The parsed raw log messages, which include sentimental words, are defined in mathematical terms as $L = \{L_{1},\ L_{2},\ \ldots,\ L_{\left|L\right|}\}$, where the raw log message $L_{i}$ extracted from the log entry $i$, $\left|L\right|$ denotes the total number of log messages. The architecture of model blocks, model parameters, and label space are unified through task formulation, wherein two transformer blocks receive semantic and sequence modalities. These transformers learn representations of log data. Finally, our model attempts to predict the integer value $\tilde{y}_{i}\in\{0,\ 1\}$ needed to classify log entry $i$.

Task formulation involves two subsections. formulating input features describes the process of transforming raw log modalities into input feature vectors for the model. However, label formulation applies MSA for log anomaly detection tasks by transforming log message labels into a space representing sentimental states.

\subsubsection{Formulating Input Features}\label{subsubsec4: formulatinginputfeatures}

The comprehension of events' sentiments mainly relies on the semantic information in log messages. Based on this observation, constructing semantic modality involves preprocessing and segmenting each raw log message into tokens from $t_{1}$ to $t_{n}$ as raw words. Subsequently, each token is added to a list that retains preprocessed log messages represented as segmented text:

\begin{equation}
	L_{i}^{sem} = [t_{1},\ t_{2},\ \ldots,\ t_{n}].\label{eq1}
\end{equation}
where,
\begin{align}
	n = {number \hspace{0.1cm} of \hspace{0.1cm} tokens} \label{eq2}
\end{align}

After this, word embedding vectors are acquired by utilizing word2vec \citep{ref116} to create embedding vectors of tokens in ${L}_{i}^{sem}$.

\begin{figure}[h]
	\centering
	\includegraphics[width=0.5\textwidth]{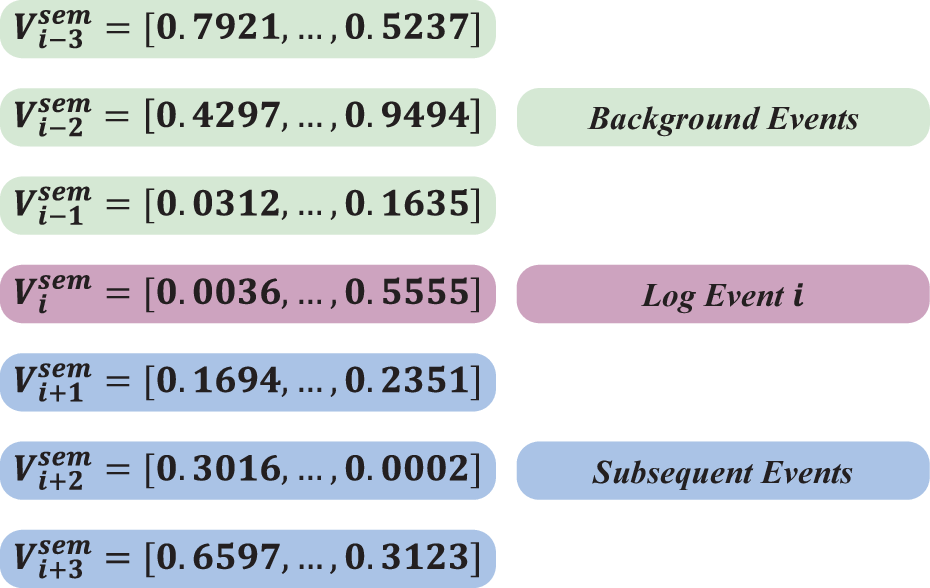}
	\caption{Illustration of differences between background and subsequent event vectors. Background and context sequence vectors are constructed based on background and subsequent event vectors. In mathematical terms, ${B}_{i}=[{V}_{{i}-{3}}^{sem},\ {V}_{{i}-{2}}^{sem},\ {V}_{{i}-{1}}^{sem}]$ and ${C}_{i}=[{V}_{{i}-{3}}^{sem},\ {V}_{{i}-{2}}^{sem},\ {V}_{{i}-{1}}^{sem},\ {V}_{{i}+{1}}^{sem},\ {V}_{{i}+{2}}^{sem},\ {V}_{{i}+{3}}^{sem}]$, where ${V}_{i}^{sem}$ is the semantic vector of the log message ${i}$ extracted by SBERT, ${B}_{i}$ is the background sequence vector of log message ${i}$, and ${C}_{i}$ is the context sequence vector of log message ${i}$.}\label{fig5}
\end{figure}

Furthermore, the raw sequence features of every log message are transformed into numerical sequential vectors. Log sequence vectors can be produced in two procedures: background and context. As illustrated in Figure~\ref{fig5}, generating background and subsequent event lists for every log entry is possible. A background sequence vector can be generated by concatenating background event vectors. The resulting vector is a $\left({W}\times{k}\right)$-dimensional, where $W$ represents the window size. Based on the above mentioned concepts, ${L}_{i}^{seq}$ can be defined as follows:

\begin{equation}
	L_{i}^{seq} = [[L_{j}]].\label{eq3}
\end{equation}
where,
\begin{align}
	{j}\in{\{i-W,\ i-(W-1),\ \ldots,\ i-1\}} \label{eq4}
\end{align}

Similarly, the context sequence vector, constructed by concatenating the background and subsequent event vectors, has a dimension size of ${2W}\times{k}$.

\begin{equation}
	L_{i}^{seq} = [[L_{j}]].\label{eq5}
\end{equation}
where,
\begin{align}
	{j}\in\{i-W,\ i-(W-1),\ \ldots,\ i-1,\ i+1,\ \ldots,\ i+(W-1),\ i+W\} \label{eq6}
\end{align}

The decision-making process regarding whether to use Equation~\ref{eq3} or Equation~\ref{eq5} is based on the type of sequence modality that goes through background or context extraction operations.

\subsubsection{Label Formulation}\label{subsubsec4: labelformulation}

The objective of CoLog is to utilize MSA to detect log anomalies and predict the sentiment reflected by log message ${i}$. In SA, the numerical value $\mathbf{1}$ is commonly used to denote positive sentiments, while $\mathbf{0}$ is typically employed to indicate negative sentiments. The task of identifying negative sentiments can be considered equivalent to the detection of log anomalies. Consequently, every instance within the dataset is assigned a label of either $\mathbf{1}$ or $\mathbf{0}$, corresponding to its classification as a normal or abnormal sample. In this manner the universal label ${y}_{i}=\{{y}_{i}^{p}\}$ comprises the polarity of log message ${i}$, where ${y}_{i}^{p}\in\{{0},\ {1}\}$.

Unlabeled datasets are labeled using words that reflect negative sentiment in the same dataset. The list of negative sentimental words for the unlabeled datasets is based on the conditions proposed in pylogsentiment \citep{ref38}.

\subsubsection{Leveling the Playing Field: Class Imbalance}\label{subsubsec4: levelingtheplayingfieldclassimbalance}

The issue of class imbalance is a prevalent concern in machine learning, characterized by an unequal distribution of classes within the training data. The potential outcome of this scenario is the development of models that exhibit bias toward the majority class, resulting in suboptimal performance on the minority class. Various techniques have been devised to address the issue of class imbalance, including data-level approaches, e.g., over-sampling methods \citep{ref117, ref118} and under-sampling techniques \citep{ref119, ref120}, and algorithm-level approaches, e.g., ensemble methods \citep{ref121, ref122}. The techniques mentioned above are designed to enhance the efficiency of machine learning models when dealing with imbalanced datasets. This is achieved by either fixing the data imbalance or adapting the learning algorithm to consider the distribution of classes.

For class balance, we employ the under-sampling technique called the Tomek link \citep{ref120}. The Tomek link is utilized for under-sampling to generate a novel distribution of the majority class. The log data frequently contains a repetitive majority class, which can be regarded as noise. Consequently, we opted for this approach, which is commonly employed to eliminate noisy and borderline majority class instances. A Tomek link is established between two data items belonging to different classes that are the nearest neighbors to each other. Assuming the entire dataset is denoted as ${L}=\{{L}_{1},\ {L}_{2},\ \ldots,\ {L}_{\left|{L}\right|}\}$, the balanced dataset ${L}_{balanced}$ can be defined mathematically as follows:

\begin{equation}
	L_{balanced} = tomeklink(L).\label{eq7}
\end{equation}
where ${tomeklink \hspace{0.05cm} (\hspace{0.05cm} \cdot \hspace{0.05cm})}$ is a function aimed at removing Tomek links $({L}_{i},\ {L}_{j})$ from dataset ${L}$. $({L}_{i},\ {L}_{j})$ is a Tomek link provided that there is no ${L}_{k}$ whose Euclidean distance from any member of $({L}_{i},\ {L}_{j})$ is not less than the Euclidean distance of ${L}_{i}$ and ${L}_{j}$. Eliminating Tomek links continues until every pair of nearest neighbor vectors belongs to the same class. It is crucial to note that ${L}_{i}$ and ${L}_{j}$ belong to distinct classes.

\subsection{Collaborative Transformer}\label{subsec4: collaborativetransformer}

The attention layer is CoLog's most crucial component. To determine the mapping relationship among modalities of log data, CT is created by modifying the attention layers of two identical transformer blocks that concurrently learn data representations in an end-to-end manner. As a result, when learning information from one modality, we use the information from the other modality as guidelines enabled by the MHIA mechanism.

The attention mechanism of an unimodal transformer encoder is defined as follows:

\begin{equation}
	{attention}({Q},\ {K},\ {C})={softmax}(\frac{{{QK}}^{T}}{\sqrt{d}}){C}.\label{eq8}
\end{equation}
where ${Q}$, ${K}$, and ${C}$ are the respective queries, keys, and contexts. We implement the self-attention mechanism with ${Q}={K}={C}$. Query, key, and context vectors enable the model to understand the relative importance of all words within the context of the full sequence. The operation ${{QK}}^{T}$ produces a squared attention matrix that contains the correlation between row of input matrix ${V}^{m},\ {m}\in\{{sem},\ {seq}\}$ if this input has a size of ${N}\times{k}$. The expression $\sqrt{d}$ is a factor used to adjust the scale where ${d}={k}$.

The concept of stacking several self-attentions attending data from various representation sub-spaces in different positions is known as MHA as follows:

\begin{equation}
	{MHA}({Q},\ {K},{C})={concat}({h}_{1},\ {h}_{2},\ \ldots,\ {h}_{h}){W}_{o}.\label{eq9}
\end{equation}
where,
\begin{align}
	{h}_{i}&={attention}({{QW}}_{i}^{Q},\ {{KW}}_{i}^{K},\ {{CW}}_{i}^{C}) \nonumber \\
	{i}&\in\{{1},\ {2},\ \ldots,\ {h}-{1},\ {h}\} \label{eq10}
\end{align}
where ${W}^{Q}$, ${W}^{K}$, ${W}^{C}\in{R}^{{d}\times{d}}$. We stack ${h}$ layers of self-attention to perform MHA to learn complex patterns between events. The projection into the balancing layer for classification follows encoding through the transformer blocks, with the output of each transformer block being used.

\begin{figure}[h]
	\centering
	\includegraphics[width=1\textwidth]{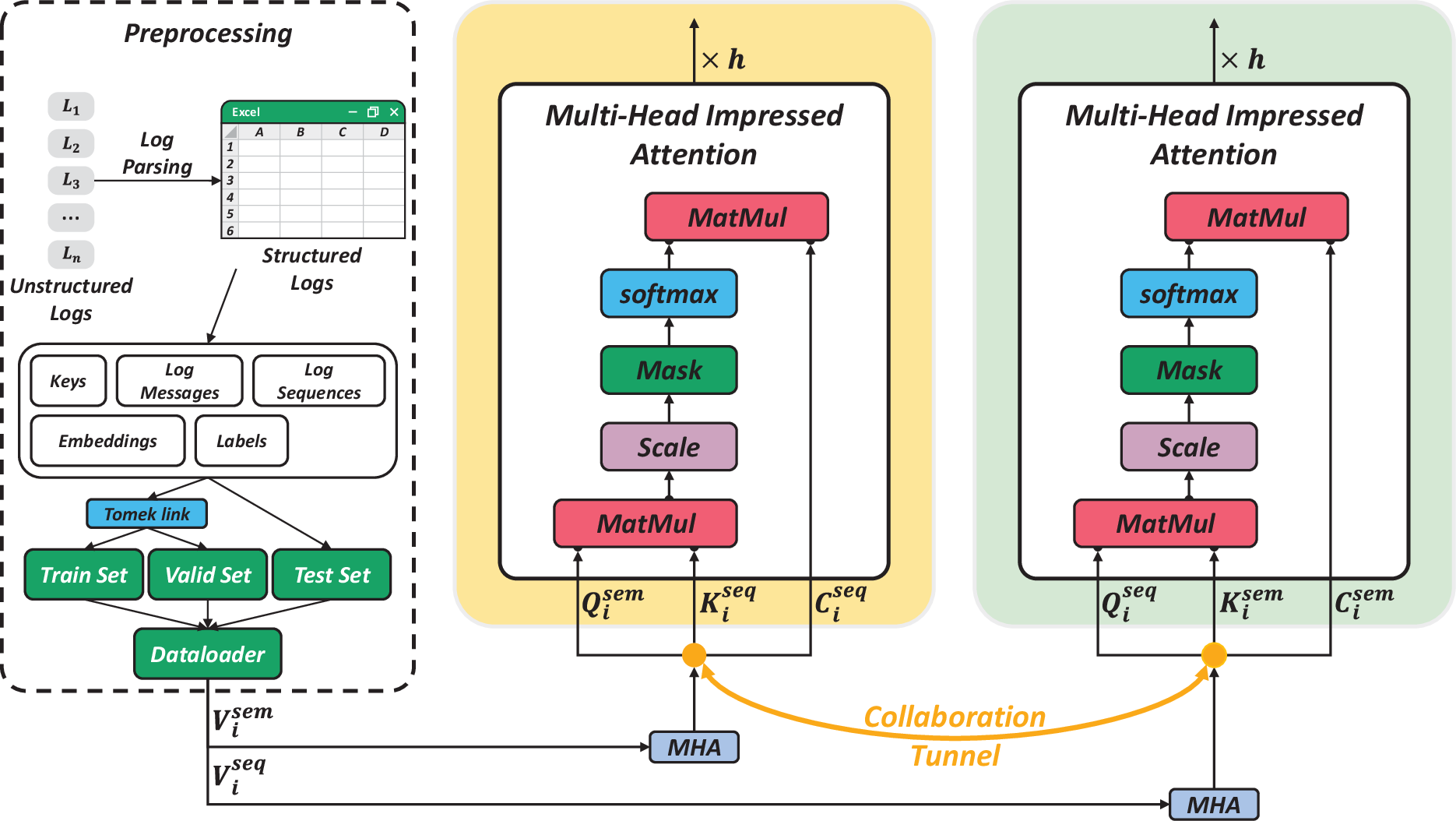}
	\caption{The architecture of the multi-head impressed attention layer. The MHIA process involves calculating the attention scores of the current modality through the MHA mechanism, using the $Q$ vector from the current modality and the $K$ and $C$ vectors from the secondary modality. According to MHIA architecture, various modalities are encoded concurrently.}\label{fig6}
\end{figure}

As illustrated in Figure~\ref{fig4}, integrating MHIA in CT involves substituting MHA with MHIA in each modality encoder, which is demonstrated with light green and gold colors. The architecture of MHIA is illustrated in Figure~\ref{fig6}. MHIA requires concurrent encoding across utilized modalities. In addition, the MHIA implementation includes computing the attention scores of the at-hand modality by utilizing the ${Q}$ vector from the at-hand modality and the ${K}$ and ${C}$ vectors from the secondary modality using the MHA mechanism. Each modality encoder transformer in the CT architecture contains a corresponding MHIA block.

\subsection{Concurrent-Encoding}\label{subsec4: concurrentencoding}

The concurrent-encoding process means simultaneous encoding of each modality. As mentioned above, this concept implies that the encoding block used for a particular modality is not unrolled before moving to another modality. The projection into latent space for classification follows encoding through the transformer blocks by the balancing layer.

\subsection{Modality Adaptation Layer}\label{subsec4: modalityadaptationlayer}

Since different modalities are collaboratively encoded in the CT architecture, it is essential to ensure that the extracted representations are free of unnecessary information. Because logs have several modalities, encoding each modality using knowledge from other modalities is crucial in obtaining an improved and more informative representation of that modality. However, due to the intrinsic variations between different modalities, encoding them collaboratively can result in inconsistencies and impurities in the representation of each modality. MAL is proposed as a solution to address these challenges. MAL comprises ${N}$ soft attention layers, which their corresponding outputs are stacked. Each modality ${V}^{m}$, ${m}\in\{{sem}$, ${seq}\}$ is initially transformed into a new high-dimensional representation space ${V}_{high}^{m}$ to start MAL. This transformation is calculated using Equation~\ref{eq11} as follows:

\begin{equation}
	{V}_{{high}}^{m}={V}^{m}{W}_{{high}}^{m}.\label{eq11}
\end{equation}
where ${V}^{m}\in{R}^{{N}\times{K}}$ is the input matrix on which CoLog performs MAL and ${W}_{{high}}^{m}\in{R}^{{k}\times\mathbf{2}{k}}$ is a transformation matrix that embeds ${V}^{m}$ into a higher dimension shared among all MALs.

In the second phase, weights for each node in the sequence must be generated from the acquired high-dimensional space to implement soft attention and extract a pure and global representation of every node in the sequence. To do this, we need the ${w}_{i}^{{node}}$, ${i}\in[{1},\ {N}]$'s. each ${w}_{i}^{{node}}$ must have a size of ${1}\times{2}{k}$, while the set of vectors $\{{w}_{i}^{{node}}\}$ is unique to each node in input sequence. The weights of each node are derived using Equation~\ref{eq12}, based on the following notions.

\begin{equation}
		{w}^{m}={softmax}({V}_{{high}}^{m}{({W}_{{low}}^{m})}^{T}).\label{eq12}
\end{equation}
where,
\begin{align}
	{W}_{{low}}^{m}={concat}({w}_{i}^{{node}}).\label{eq13}
\end{align}
where ${W}_{{low}}^{m}$ is a transformation matrix that extracts weights from high-dimension space for each node.

Equation~\ref{eq14} computes the soft attention of the input matrix based on the weights acquired from the previous phase as follows:

\begin{equation}
	{\widetilde{{V}}}_{i}^{m}={softattention}({V}^{m})=\sum_{{j}={0}}^{{N}}{{w}^{m}[{i}][{j}]}\odot{V^{m}[{j}]}.\label{eq14}
\end{equation}
where ${\widetilde{{V}}}_{i}^{m}$ represents the computation output obtained from an individual node within the sequence and $\odot$ demonstrates the Hadamard product.

The output of the MAL for an input matrix ${V}^{m}$ with a sequence length of ${N}$ can be defined as stacking all ${\widetilde{{V}}}_{i}^{m}$ vectors according to following relation:

\begin{equation}
	{\widetilde{{V}}}_{out}^{m}={layernorm}({\widetilde{{V}}}^{m}+{V}^{m}).\label{eq15}
\end{equation}
where,
\begin{align}
	{\widetilde{{V}}}^{m}={stacking}({\widetilde{{V}}}_{1}^{m},\ {\widetilde{{V}}}_{2}^{m},\ \ldots,\ {{{V}}}_{N}^{m}).\label{eq16}
\end{align}

\begin{figure}[h]
	\centering
	\includegraphics[width=1\textwidth]{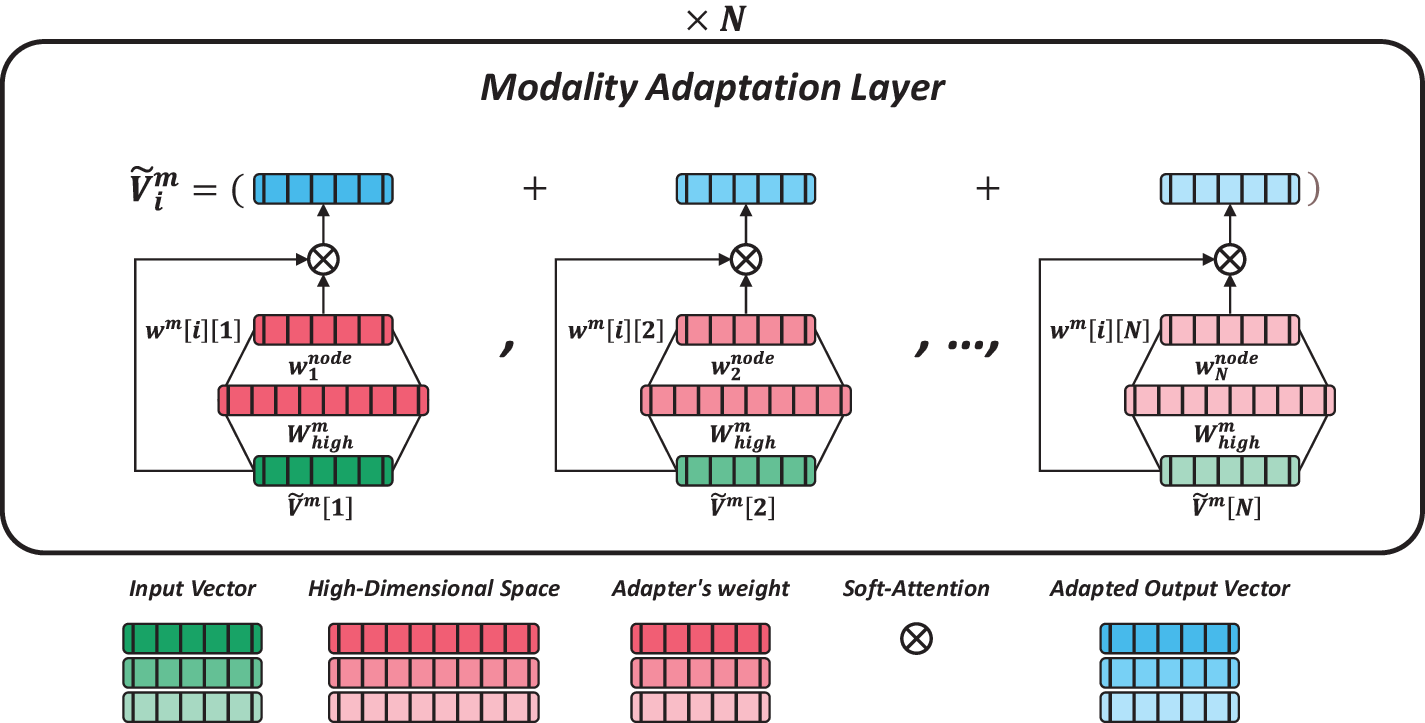}
	\caption{The architecture of modality adaptation layer. MAL achieves an overall representation for each modality by assigning weights to each node in the input sequence. It can also remove impurities of the modalities that are encoded in a collaborative manner.}\label{fig7}
\end{figure}

The transformation matrix ${W}_{{high}}^{m}$, shared among all nodes, maps each node in the input sequence to a high-dimensional space, as illustrated in Figure~\ref{fig7}. The softmax function and vector ${w}_{i}^{{node}}$ allocate the weight ${w}^{m}[{i}]$ for each node ${i}$, which is distinct for every node. The final output is obtained by computing the weighted sum of all nodes, which is then fed into the balancing and classification layers to make the final decision. We perform this to enable each modality to recognize the corresponding cross-node representations and remove impurities of each modality. By completing this task, every node vector can elicit important information from other nodes' respective representations, contributing to each node's global sentiment polarity.

\subsection{Balancing Layer}\label{subsec4: balancinglayer}

At the end of the modality transformer blocks, we add a balancing layer that projects the modality into a new representation space. Due to the inherent differences between various log modalities, CoLog summarizes feature vectors of each modality in the same semantic space. Therefore, as seen in Figure~\ref{fig4}, the learned outcomes from each modality encoder are projected into a latent space for meaningful fusion where CoLog interprets the feature vectors of each modality in a space with high dimensions. Furthermore, since different modalities may contribute different importance to determining the normality or abnormality of the input, we extract the balancer's weight for each modality from their respective high-dimensional representations. Based on the balancer's weights, we can perform soft attention to balance the contribution of modalities before fusion. The balancing layer is the same as MAL of size $\mathbf{1}$, with the distinction that the resulting vectors are projected into a latent space to ensure a meaningful fusion. Integrating a latent space in BL is due to the different nature of the modalities.

The output of the balancing layer for semantic and sequence modalities is demonstrated using the symbols ${O}_{i}^{{sem}}$ and ${O}_{i}^{{seq}}$, respectively.

\subsection{Classification}\label{subsec4: classification}

Finally, after being projected to latent space, balanced feature vectors of each modality are summed, normalized, and transformed into 2-dimensional vectors to apply the activation function. According to Figure~\ref{fig4}, the vectors corresponding to each modality undergo element-wise summation following the successful computation of all transformer blocks. Subsequently, these vectors are used to predict each input sample by MLP and layer normalization, as demonstrated by:

\begin{equation}
	{y}_{i}={W}_{{classification}}({layernorm}({O}_{i}^{{sem}}+{O}_{i}^{{seq}})).\label{eq17}
\end{equation}
where ${W}_{{classification}}$ denotes the learning matrix of MLP.

\subsection{Anomaly Detection}\label{subsec4: anomalydetection}

If various CoLog inputs are provided to the model within a single variable, the CoLog input can be referred to as ${I}=\{{{keys},\ {V}^{{sem}},\ {V}^{{seq}},\ {embeddings},\ {labels}}\}$. CoLog receives ${I}$ and tries to predict the set of labels ${y}$. The anomaly detection phase is similar to the training process. It is important to note that CoLog can detect anomalies in two distinct manners. This involves prioritizing the detection of point anomalies only or the detection of point and collective anomalies.

\subsubsection{Point Anomaly Detection}\label{subsubsec4: pointanomalydsetection}

Based on the methodology provided in Section~\ref{sec4: method}, CoLog takes input ${I}$ and attempts to make predictions about them during the point anomaly detection procedure. This procedure aims to receive an individual log entry by CoLog, analyze it in collaboration with its background or context, and determine its polarity.

\subsubsection{Point and Collective Anomaly Detection}\label{subsubsec4: pointandcollectiveanomalydsetection}

Point and collective anomaly detection through a unified framework entail receiving input by CoLog and attempting to ascertain the polarity of an individual log event and its background or context polarity. From this concept, it can be inferred that in this scenario, the task of CoLog is a 4-class classification. These classes encompass the abnormality of an event, the abnormality of the event and its background or context, the abnormality of just the background or context of the event, and the normality of both the event and its background or context. Hence, it is imperative to label the background or context of the events in this particular scenario.

A log event sequence ${L}=\{{L}_{1},\ {L}_{2},\ \ldots,\ {L}_{|{L}|}\}$ is abnormal if one of its log messages reflects a negative sentiment. Based on this, the background or context of each log event can also be labeled. Finally, based on the methodology provided in Section~\ref{sec4: method}, CoLog takes input ${I}$ and generates predictions during the point and collective anomaly detection procedure. This procedure aims to receive an individual log entry by CoLog, analyze it in collaboration with its background or context, and determine its polarity as well as its background or context polarity.

\section{Experiments}\label{sec5: experiments}

This section will begin by describing the OS log datasets. Subsequently, we will discuss the experimental setup of CoLog and the evaluation results of the proposed model.

\subsection{Log Datasets}\label{subsec5: logdatasets}

Experiments are conducted on benchmark OS log datasets, including the publicly available BlueGene/L, Hadoop, and Zookeeper datasets. Table~\ref{tab2} presents a comprehensive statistical information of all datasets utilized for evaluating CoLog's performance.

\begin{table}[h]
	\caption{A summary of publicly available OS log datasets.}\label{tab2}%
	\begin{tabular}{@{}lllll@{}}
		\toprule
		Dataset                 & \# Lines     & \# Positive Records & \# Negative Records & Data Size                   \\
		\midrule[0.01cm]
		Spark\footnotemark[1]   & $33,236,604$ & $31,513,147$        & $1,723,457$         & $2.75 \hspace{0.15cm} GB$   \\
		Honey5\footnotemark[2]  & $124,386$    & $67,798$            & $56,588$            & $12.6 \hspace{0.15cm} MB$   \\
		Windows\footnotemark[2] & $25,000$     & $18,599$            & $6,401$             & $3.48 \hspace{0.15cm} MB$   \\
		Casper                  & $11,086$     & $9,874$             & $1,212$             & $930 \hspace{0.15cm} KB$    \\
		Jhuisi                  & $11,737$     & $9,063$             & $2,674$             & $0.98 \hspace{0.15cm} MB$   \\
		Nssal                   & $107,093$    & $91,349$            & $15,732$            & $8.53 \hspace{0.15cm} MB$   \\
		Honey7                  & $8,712$      & $8,162$             & $550$               & $734 \hspace{0.15cm} KB$    \\
		Zookeeper               & $74,380$     & $25,873$            & $48,507$            & $9.95 \hspace{0.15cm} MB$   \\
		Hadoop                  & $394,308$    & $382,870$           & $11,438$            & $48.61 \hspace{0.15cm} MB$  \\
		BlueGene/L              & $4,747,963$  & $4,399,486$         & $348,477$           & $708.76 \hspace{0.15cm} MB$ \\														
		\botrule
	\end{tabular}
	\footnotetext[1]{The provided datasets are intended to evaluate the robustness of CoLog.}
	\footnotetext[2]{The provided datasets are intended to evaluate the generalization of CoLog.}
\end{table}

Spark \citep{ref123}, Honey5 \citep{ref124}, and Windows \citep{ref123} datasets are not used during CoLog training. These datasets are used to test the generalizability and robustness of CoLog. Spark is a management solution for handling large amounts of data. Normal and abnormal Spark system actions are both represented in the logs obtained from 32 hosts. Honey5 originates from the Forensic Challenge 5 in 2010, which the Honeynet Project conducted. The dataset provided is an instance of a Linux OS that has been compromised. As the last dataset, we use the Windows dataset, obtained by consolidating many logs from the CUHK laboratory computer operating on Windows 7. The Windows OS implemented a component-based servicing (CBS) mechanism to facilitate programs' secure and controlled installation processes.

The first dataset of the other seven datasets of this collection used in the model training process was taken from a disk image provided by Digital Corpora and given the name nps-2009-casper-rw \citep{ref125}. A bootable USB was used to create a dump of the ext3 file system. The OS logs from a machine running the Linux OS are included in this dataset. The digital forensic research workshop (DFRWS), an annual security conference, provides the second and third system log datasets we use to train CoLog. They presented an OS log forensics challenge in 2009. The DFRWS forensic challenge 2009 studied log files to find an intruder who illegally communicated classified material. This case involves jhuisi \citep{ref126} and nssal \citep{ref126} hosts. Two Linux-based Sony PlayStation 3 hosts. Additional OS logs were obtained via the honeynet forensic challenge 7 (Honey7) \citep{ref127}. Honey7 includes a disk image of a compromised Linux server. The cloned disk images contained the directory \textit{\textbf{/var/log/}} for all datasets. Then, authentication, kernel, and system logs are obtained. Zookeeper manages distributed systems. The CUHK laboratory collected Zookeeper logs \citep{ref123} from 32 hosts for 26 days. Big data tool Hadoop distributes jobs across machines. A Hadoop cluster with 46 cores on five machines created the Hadoop log dataset \citep{ref128}. In this dataset, machine downtime and network disconnections are anomalies. The 131,072 processor, 32,768 GB memory BlueGene/L supercomputer at lawrence livermore national laboratories (LLNL) provides an open dataset \citep{ref129}. The records comprise alert and non-alert, where alerts signify abnormal behaviors.

\subsection{Experimental Setup}\label{subsec5: experimentalsetup}

This section describes the hardware and software configurations of the system we used to run experiments. We then discuss CoLog's hyperparameters and the metrics utilized to evaluate CoLog.

\subsubsection{System Configs}\label{subsubsec5: systemconfigs}

We use a Colab pro machine with 12 cores on the central processing unit (CPU), 83.5 gigabytes of random-access memory (RAM), and an NVIDIA A100 40GB graphics processing unit (GPU) to execute the experiments for both the CoLog and the other methods.

Python 3.10.12 is used to develop CoLog, and PyTorch \citep{ref130} 2.0.1+cu118 is used as the backend. We use the imbalanced-learn \citep{ref131} library to implement the Tomek link and evaluate it against other class-balancing techniques. The Ray library \citep{ref132} in Python was utilized to determine the model's optimal parameters. The sentence-transformers library \citep{ref111} was also employed to extract sentence embedding vectors from the log messages.

\subsubsection{Hyperparameters}\label{subsubsec5: hyperparameters}

The Ray Python library chooses the most optimal hyperparameters for CoLog. Using the above mentioned library, 108 distinct configurations of CoLog were analyzed. In addition, an examination of window size, class imbalance, and the training ratio has been conducted independently to determine the optimal values for each hyperparameter. The number of CT layers is 2, and the number of heads is 4. The batch size is 32. The learning rate for training the CoLog is set to 5e-5, whereas the learning rate decay is set to 0.5, and the learning rate will decrease three times at this rate. The Adam optimizer method is employed. Adam is considered suitable for this case due to its minimal RAM requirements. The training process has a maximum epoch of 20 and an early stopping criterion set to 5. The training process will be terminated if no further improvement is shown after 5 epochs. The model warm-up process consists of 5 epochs. The dropout rate of 0.1 is used. The lengths of the semantic and sequence modalities are 60, whereas the embedding size of the word vectors is 300. The hidden size of 256 is used for the CoLog model. The projection vector size is also 2048. The sequence modality type is set to context, and a window size 1 is used. Dimensions of sequence modality are set to 384. We employ a holdout validation protocol. The datasets were divided into three pieces based on the following proportions: 60\% for training, 20\% for validating, and 20\% for testing. In the context of the BGL dataset, the ratio mentioned above is 10\%, 45\%, and 45\%, which can be attributed to constraints imposed by limited resources. Also, the Tomek link technique was employed to address the data imbalance. Finally, CoLog is trained on each log dataset and tested on the same dataset for generating results.

\subsubsection{Metrics}\label{subsubsec5: metrics}

The performance of CoLog is assessed using multiple metrics, including precision, recall, F1-score, and accuracy. These metrics are computed based on the values of true positives (TP), false positives (FP), true negatives (TN), and false negatives (FN). The scikit-learn library \citep{ref133} is used to implement metrics with the "macro average" option. Evaluation metrics are computed for each dataset on individual files, and then the average is derived across all files.

\bmhead{Precision}

Precision quantifies the proportion of correctly identified positive instances across all positive predictions generated by the CoLog. The precision formula is defined as follows:

\begin{equation}
	{Precision}=\frac{{TP}}{{TP}+{FP}}.\label{eq18}
\end{equation}
where ${TP}$ is correctly predicted positive instances and ${FP}$ is incorrectly predicted positive instances.

\bmhead{Recall or Sensitivity}

Recall quantifies the proportion of actual positive cases correctly detected by the CoLog. The mathematical expression representing the concept of recall is as follows:

\begin{equation}
	{Recall}=\frac{{TP}}{{TP}+{FN}}.\label{eq19}
\end{equation}
where ${FN}$ is incorrectly predicted negative instances.

\bmhead{F1-score}

F1-score pertains to evaluating the balance between precision and recall. The mathematical expression representing the F1-score is as follows:

\begin{equation}
	{F1}\textendash{score}={2}\times\frac{{Precision}\times{Recall}}{{Precision}+{Recall}}=\frac{{2}\times{TP}}{{2}\times{TP}+{FP}+{FN}}.\label{eq20}
\end{equation}

\bmhead{Accuracy}

Accuracy quantifies the proportion of correct predictions made by the CoLog. The following equation gives the mathematical expression representing accuracy:

\begin{equation}
	{Accuracy}=\frac{{TP}+{TN}}{{TP}+{FP}+{FN}+{TN}}.\label{eq21}
\end{equation}
where ${TN}$ is correctly predicted negative instances.

\subsection{Results}\label{subsec5: results}

\begin{figure}[h]
	\centering
	\includegraphics[width=1\textwidth]{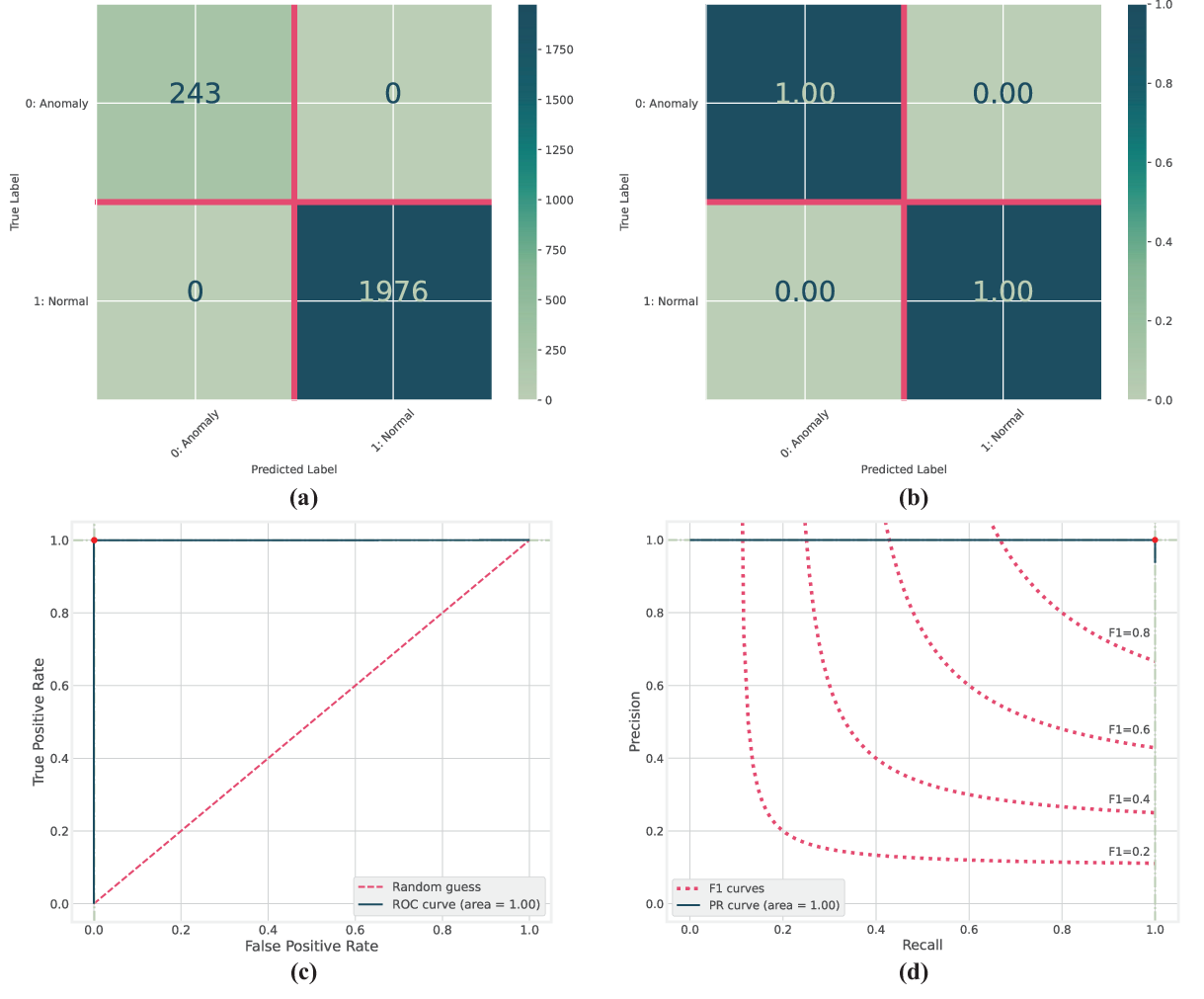}
	\caption{The collection of output visualizations generated by the CoLog when applied to the Casper log dataset. (a) confusion matrix, (b) normalized confusion matrix, (c) receiver operating characteristics curve, and (d) precision-recall curve}\label{fig8}
\end{figure}

\begin{figure}[h]
	\centering
	\includegraphics[width=1\textwidth]{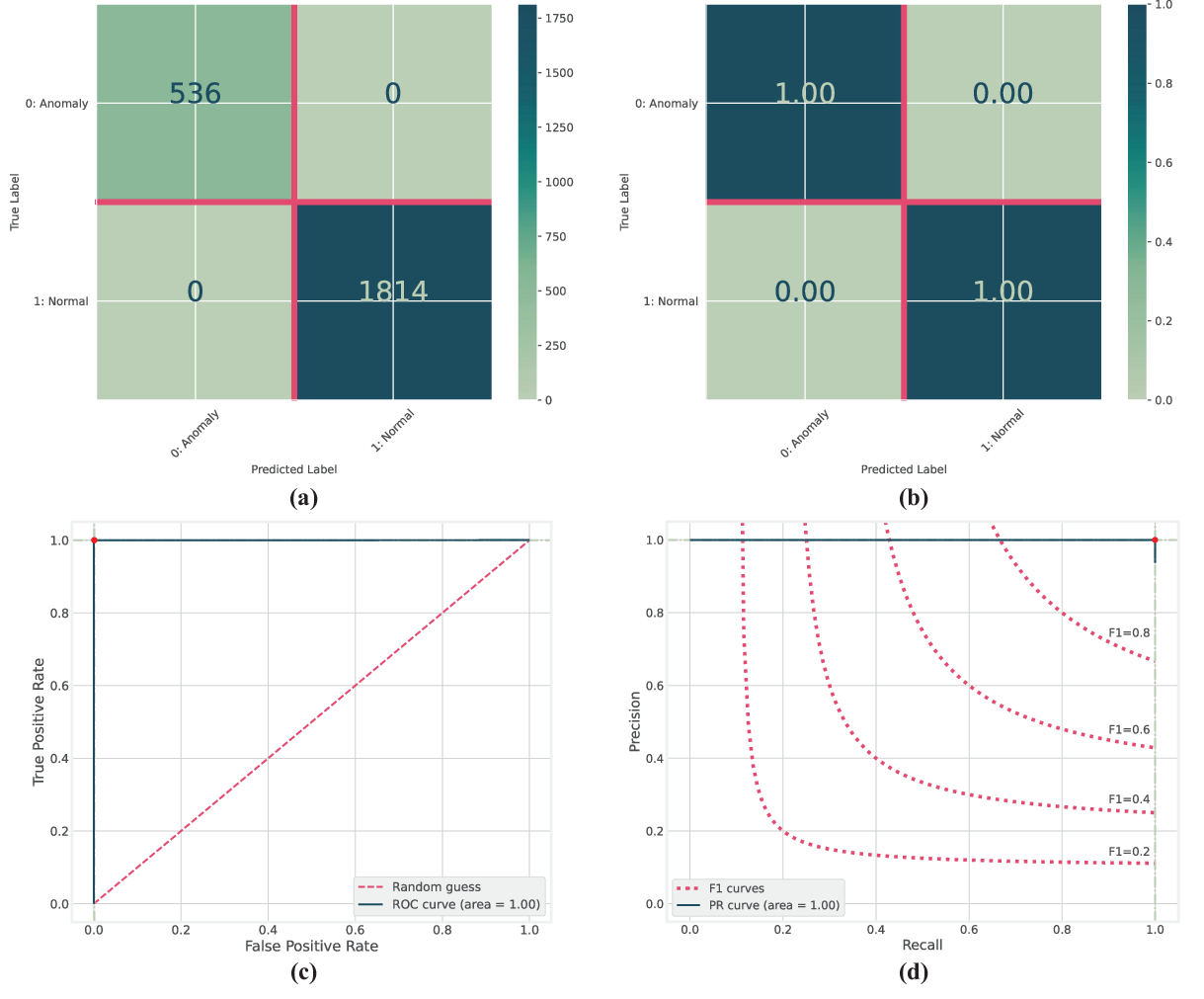}
	\caption{The collection of output visualizations generated by the CoLog when applied to the Jhuisi log dataset. (a) confusion matrix, (b) normalized confusion matrix, (c) receiver operating characteristics curve, and (d) precision-recall curve}\label{fig9}
\end{figure}

\begin{figure}[h]
	\centering
	\includegraphics[width=1\textwidth]{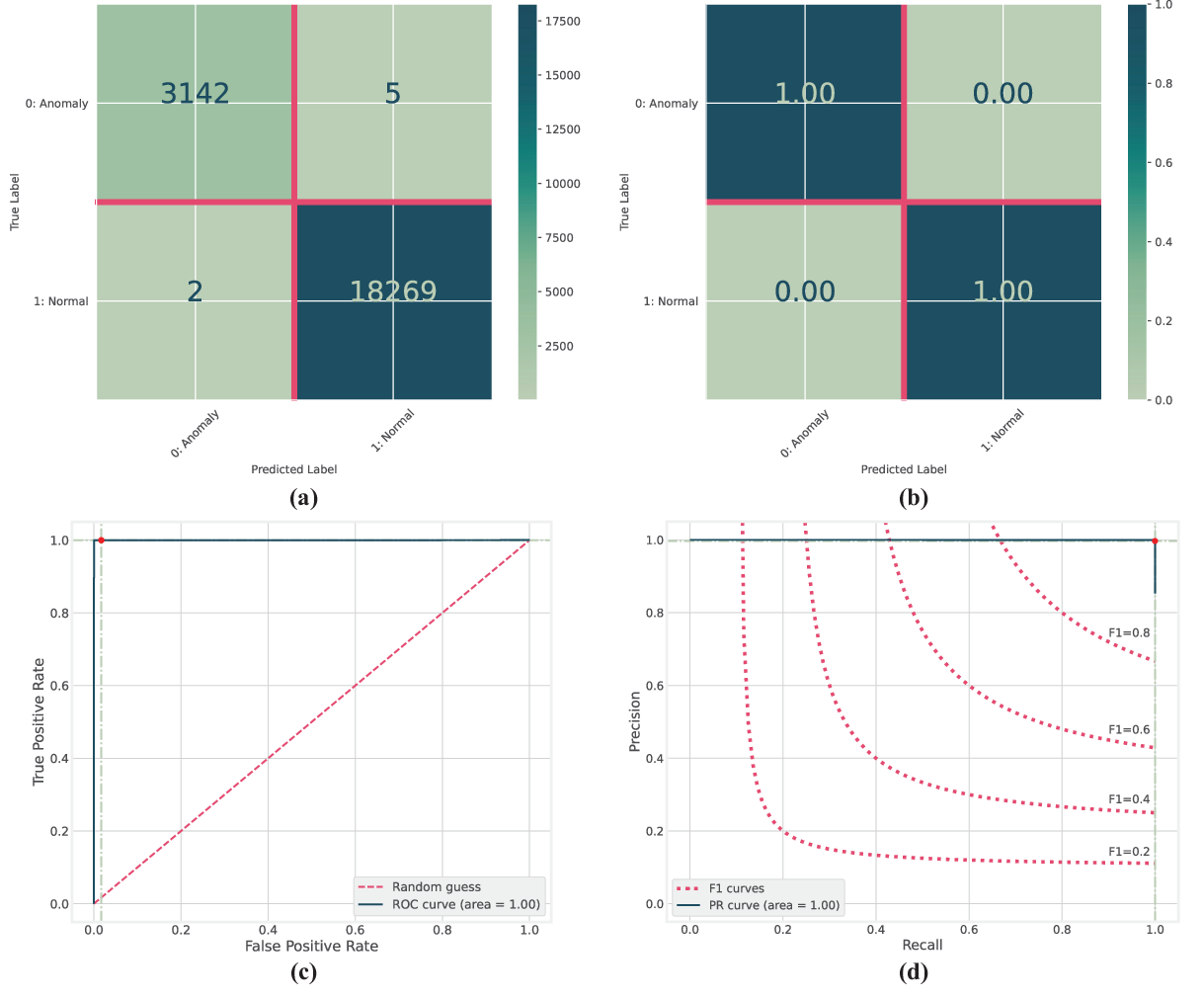}
	\caption{The collection of output visualizations generated by the CoLog when applied to the Nssal log dataset. (a) confusion matrix, (b) normalized confusion matrix, (c) receiver operating characteristics curve, and (d) precision-recall curve}\label{fig10}
\end{figure}

\begin{figure}[h]
	\centering
	\includegraphics[width=1\textwidth]{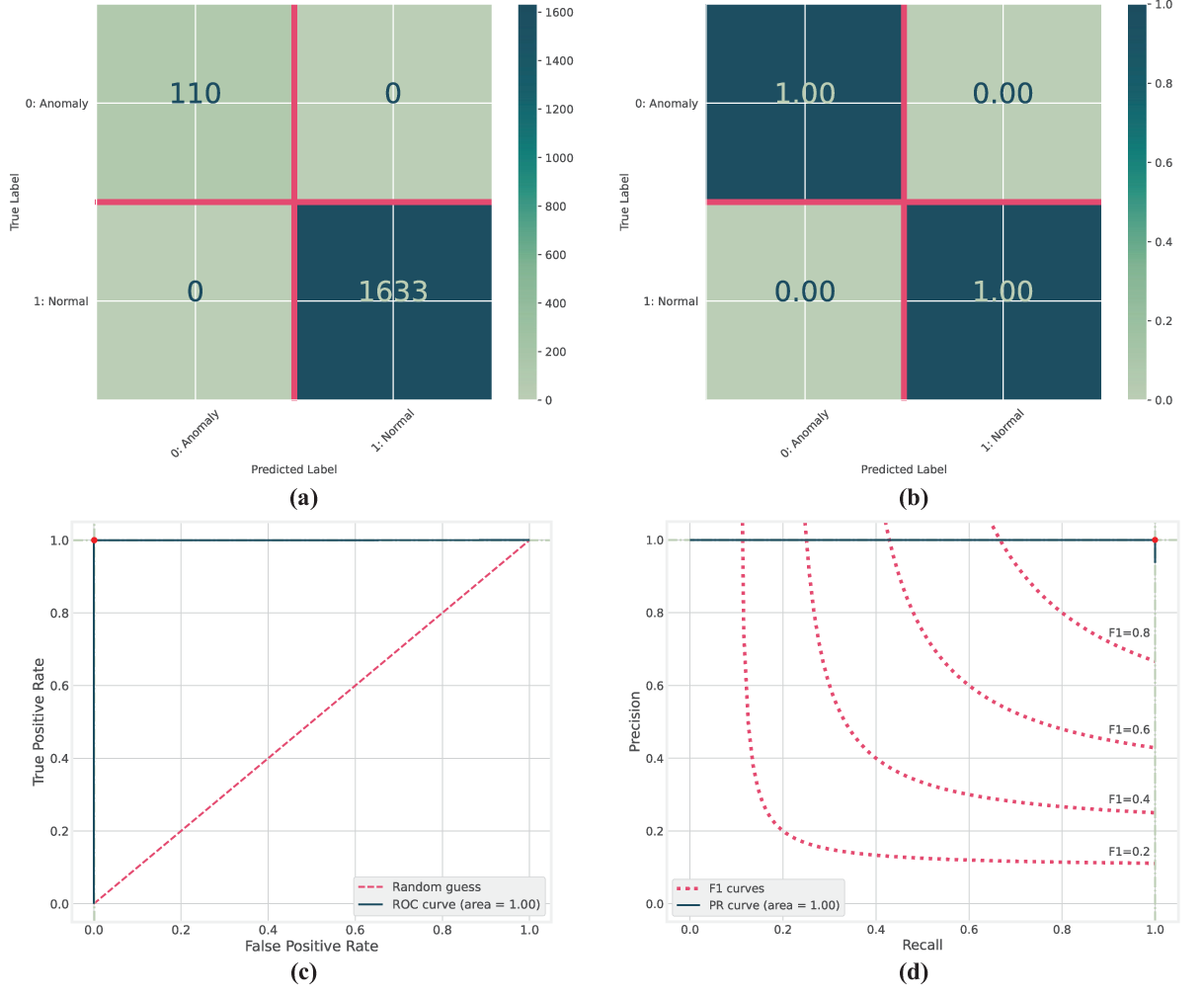}
	\caption{The collection of output visualizations generated by the CoLog when applied to the Honey7 log dataset. (a) confusion matrix, (b) normalized confusion matrix, (c) receiver operating characteristics curve, and (d) precision-recall curve}\label{fig11}
\end{figure}

\begin{figure}[h]
	\centering
	\includegraphics[width=1\textwidth]{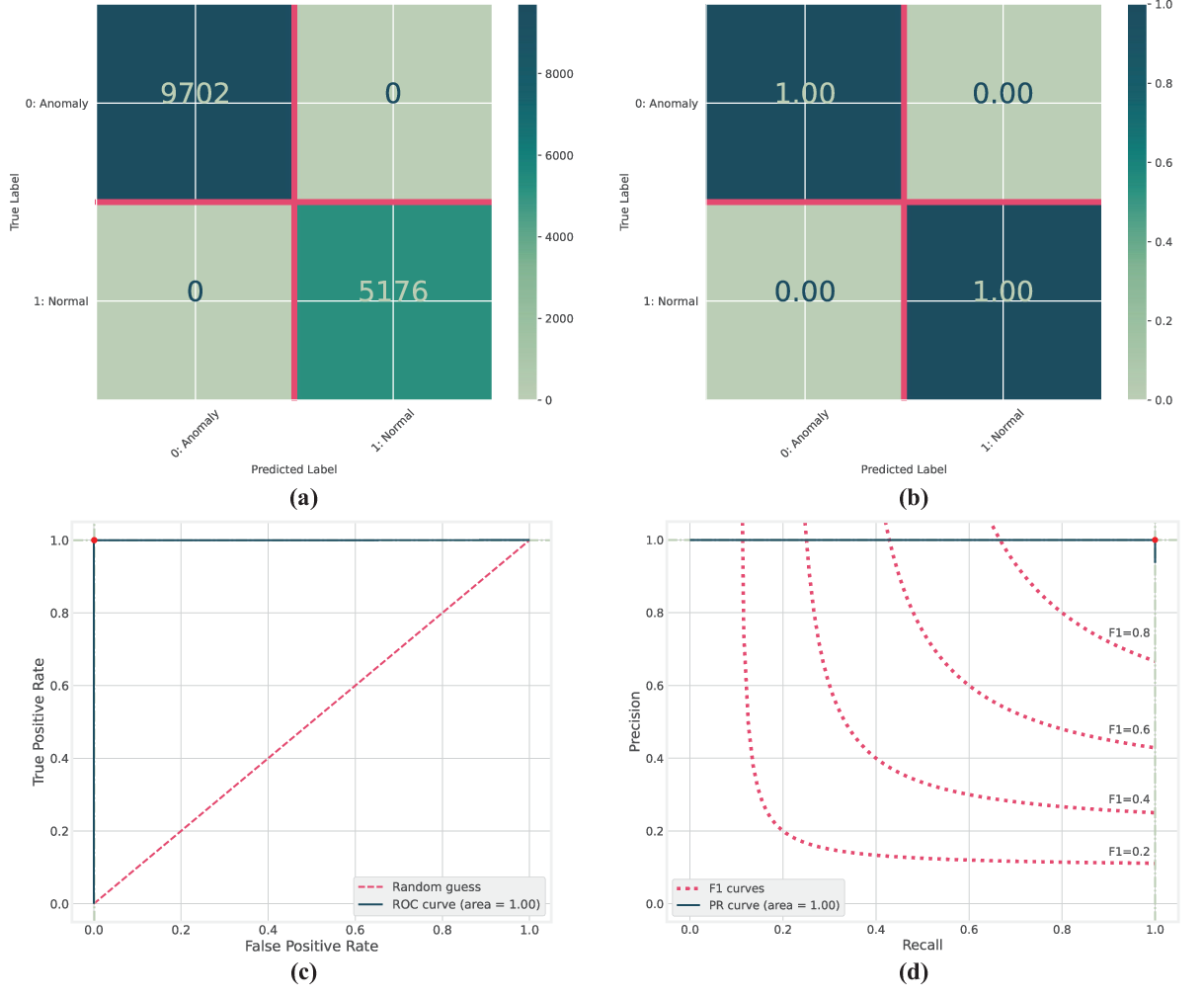}
	\caption{The output collection of visualizations generated by the CoLog when applied to the Zookeeper dataset. (a) confusion matrix, (b) normalized confusion matrix, (c) receiver operating characteristics curve, and (d) precision-recall curve}\label{fig12}
\end{figure}

\begin{figure}[h]
	\centering
	\includegraphics[width=1\textwidth]{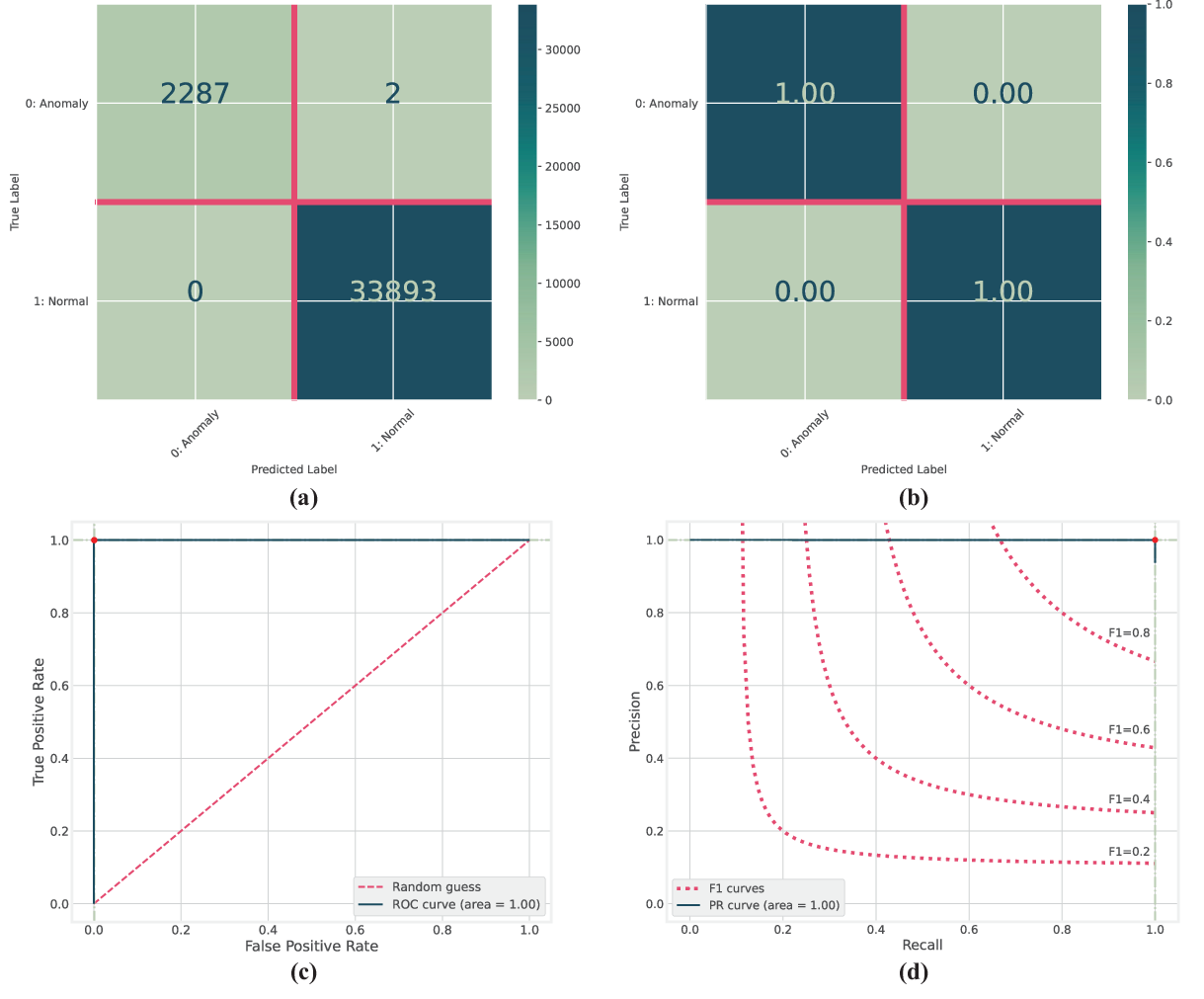}
	\caption{The collection of output visualizations generated by the CoLog when applied to the Hadoop log dataset. (a) confusion matrix, (b) normalized confusion matrix, (c) receiver operating characteristics curve, and (d) precision-recall curve}\label{fig13}
\end{figure}

\begin{figure}[h]
	\centering
	\includegraphics[width=1\textwidth]{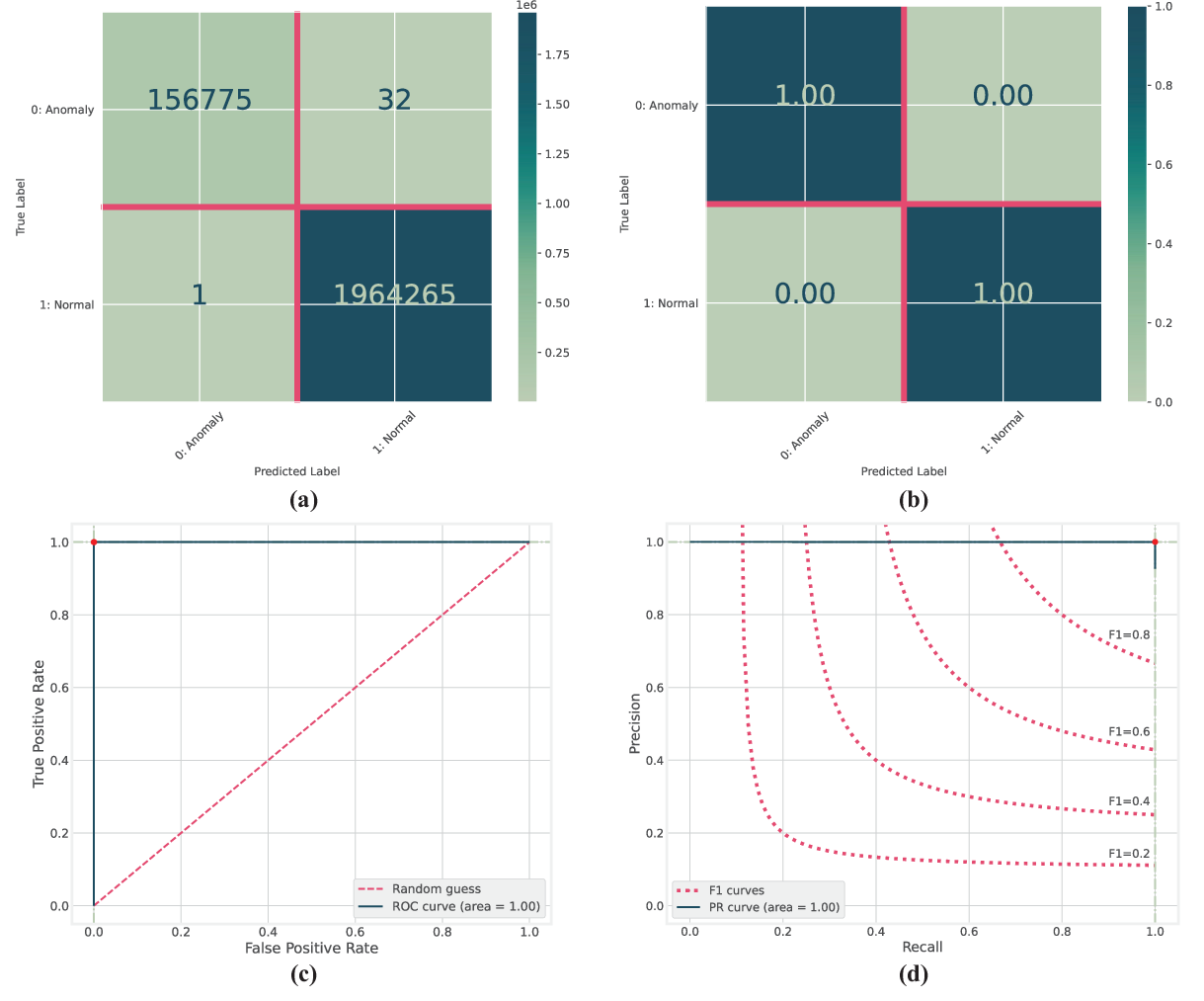}
	\caption{The collection of output visualizations generated by the CoLog when applied to the BlueGene/L log dataset. (a) confusion matrix, (b) normalized confusion matrix, (c) receiver operating characteristics curve, and (d) precision-recall curve}\label{fig14}
\end{figure}

In this section, we discuss experiments conducted on the CoLog, whereby several diagrams, including the confusion matrix, normalized confusion matrix, receiver operating characteristic (ROC) curve, and precision-recall (PR) curve, were utilized. The comprehensive CoLog outcomes on publicly available log datasets are presented in Table~\ref{comprehensivecoLogoutcomes}. Additionally, the graphical depictions of these results are illustrated in Figure~\ref{fig8} to Figure~\ref{fig14}. Also, subsections are organized as follows. In the subsequent analysis, we compare CoLog and other log anomaly detection approaches by applying them to the above mentioned datasets. Next, we will discuss the outcomes of the hyperparameter tuning process, the impact of the training ratio, the influence of various approaches for addressing the class imbalance, the effect of window size, and the presentation of the vectors derived from the CoLog. Next, let us examine the generalization or robustness of the CoLog in detail. Subsequently, we discuss the significant design components of the CoLog as the ablation study. Next, we undertake groundbreaking research. Subsequently, we will compare our contributions with other log anomaly detection techniques.

The CoLog has exhibited exceptional versatility when applied to diverse datasets, consistently attaining a high level of performance. The method's efficacy is apparent through its high accuracy in classifying all tested datasets, where CoLog achieves near-perfect or perfect classification scores. CoLog demonstrates flawless classification performance on the Casper, Jhuisi, Honey7, and Zookeeper datasets, exhibiting a complete absence of false positives or negatives. Similarly, CoLog exhibited outstanding performance on the Nssal and Hadoop datasets, with few occurrences of false positives and negatives, where all anomalies in the Hadoop dataset are detected. Even when applied to the large BGL dataset, CoLog maintains exceptional performance, accurately identifying 1,964,265 true positives and 156,775 true negatives while exhibiting few mistakes with only 1 false negative. The area under the curve (AUC) values for the ROC and PR curves are consistently near 100 or 100 across all datasets, providing additional evidence of the CoLog's strong performance.

\begin{table}[h]
	\caption{Comprehensive CoLog outcomes on publicly available log datasets.}\label{comprehensivecoLogoutcomes}%
	\begin{tabular}{@{}l|l|l|l|l|l|l|l@{}}
		\toprule
		Dataset                & Average & Class      & \# Samples & Precision & Recall   & F1-Score & Accuracy \\
		\botrule
		Casper                 &         & 0: Anomaly & 243        & $100$     & $100$    & $100$    & $100$    \\
		\cmidrule{3-7}%
		                       &         & 1: Normal  & 1976       & $100$     & $100$    & $100$    &          \\
		\cmidrule{2-7}%
		                       & Macro   &            & 2219       & $100$     & $100$    & $100$    &          \\
		\botrule
		Jhuisi                 &         & 0: Anomaly & 536        & $100$     & $100$    & $100$    & $100$    \\
		\cmidrule{3-7}%
		                       &         & 1: Normal  & 1814       & $100$     & $100$    & $100$    &          \\
		\cmidrule{2-7}%
		                       & Macro   &            & 2350       & $100$     & $100$    & $100$    &          \\
		\botrule
		Nssal                  &         & 0: Anomaly & 3147       & $99.936$  & $99.841$ & $99.889$ & $99.967$ \\
		\cmidrule{3-7}%
		                       &         & 1: Normal  & 18271      & $99.972$  & $99.989$ & $99.981$ &          \\
		\cmidrule{2-7}%
		                       & Macro   &            & 21418      & $99.955$  & $99.915$ & $99.935$ &          \\
		\botrule
		Honey7                 &         & 0: Anomaly & 110        & $100$     & $100$    & $100$    & $100$    \\
		\cmidrule{3-7}%
		                       &         & 1: Normal  & 1633       & $100$     & $100$    & $100$    &          \\
		\cmidrule{2-7}%
		                       & Macro   &            & 1743       & $100$     & $100$    & $100$    &          \\
		\botrule
		Zookeeper              &         & 0: Anomaly & 9702       & $100$     & $100$    & $100$    & $100$    \\
		\cmidrule{3-7}%
		                       &         & 1: Normal  & 5176       & $100$     & $100$    & $100$    &          \\
		\cmidrule{2-7}%
		                       & Macro   &            & 14878      & $100$     & $100$    & $100$    &          \\
		\botrule
		Hadoop                 &         & 0: Anomaly & 2289       & $100$     & $99.913$ & $99.956$ & $99.994$ \\
		\cmidrule{3-7}%
		                       &         & 1: Normal  & 33893      & $99.994$  & $100$    & $99.997$ &          \\
		\cmidrule{2-7}%
		                       & Macro   &            & 36182      & $99.997$  & $99.956$ & $99.977$ &          \\
		\botrule
		BlueGene/L             &         & 0: Anomaly & 156807     & $99.999$  & $99.980$ & $99.989$ & $99.998$ \\
		\cmidrule{3-7}%
		                       &         & 1: Normal  & 1964266    & $99.998$  & $100$    & $99.999$ &          \\
		\cmidrule{2-7}%
		                       & Macro   &            & 2121073    & $99.999$  & $99.990$ & $99.994$ &          \\
		\botrule
	\end{tabular}
\end{table}

The exceptional outcomes of CoLog can be attributed to several significant contributions. Initially, in CoLog, semantic and sequence log modalities collaborate. Utilizing sequential features besides semantic features augments the available knowledge and boosts the model's capacity to identify anomalies. Furthermore, distinct neural networks are used for each modality, which enables CoLog to control the interactions between modalities, yielding more reliable outcomes. Finally, a latent space is used to fuse the modalities. The significance of this matter lies in the fact that various modalities exhibit dissimilar characteristics, necessitating a sophisticated fusion mechanism for their meaningful integration. These contributions demonstrate the adaptability of CoLog and establish its reliability as a tool for detecting log anomalies in various scenarios.

\subsubsection{CoLog Versus the Field: A Comparative Study}\label{subsubsec5: coLogversusthefieldacomparativestudy}

We compared CoLog and some log anomaly detection algorithms available in the \href{https://github.com/logpai/loglizer}{loglizer} \citep{ref110} and \href{https://github.com/logpai/deep-loglizer}{deeploglizer} \citep{ref29} toolboxes. The loglizer is utilized for machine learning algorithms, whereas the deeploglizer is employed for deep learning-based methods. These toolboxes consist of five supervised detection approaches, including the logistic regression (LR) \citep{ref134}, support vector machines (SVM) \citep{ref108}, and Decision Tree (DT) [109] from the loglizer toolbox, as well as the Attentional BiLSTM [62] and CNN [45] from the deeploglizer toolbox. Furthermore, four unsupervised techniques, including the isolation forest (IF) \citep{ref135} and principal component analysis (PCA) \citep{ref18} algorithms from loglizer, the LSTM \citep{ref36} and Transformer \citep{ref79} implemented in the deeploglizer. In addition, we evaluate the performance of CoLog in comparison to pylogsentiment. The pylogsentiment is a tool for sentiment analysis in OS logs. It uses GRU to identify abnormalities and classify log entries according to their sentiment. This aids in detecting potential issues and evaluating a system's overall state.

Prior to feeding the log datasets into the loglizer or deeploglizer models, we employes the Drain and nerlogparser tools to separate all the log records and extract the log message. The hyperparameters of the models are set to the same configuration as recommended in the loglizer and deeploglizer toolboxes. Furthermore, sliding partitioning is implemented to perform log partitioning. Then, algorithms in loglizer and deeploglizer are employed to process these sequences of log events. Table~\ref{tab3} to Table~\ref{tab9} present the performance evaluation of CoLog and other anomaly detection methods.

\begin{table}[h]
	\caption{CoLog in comparison to other log anomaly detection methods on the Casper dataset. Bold numbers indicate the outstanding results.}\label{tab3}%
	\begin{tabular}{@{}lllll@{}}
		\toprule
		Anomaly Detection Technique                & Precision      & Recall         & F1-Score       & Accuracy       \\
		\botrule
		Supervised Methods                                                                                             \\
		\midrule[0.01cm]
		Logistic Regression \citep{ref134}         & $66.959$       & $60.863$       & $59.700$       & $90.993$       \\
		Support Vector Machines \citep{ref108}     & $58.614$       & $60.757$       & $59.482$       & $90.967$       \\
		Decision Tree \citep{ref109}               & $83.466$       & $77.037$       & $79.488$       & $94.998$       \\
		Attentional BiLSTM \citep{ref62}           & $99.766$       & $99.872$       & $99.819$       & $99.834$       \\
		Convolutional Neural Network \citep{ref45} & $99.766$       & $99.872$       & $99.819$       & $99.834$       \\
		pylogsentiment \citep{ref38}               & $99.487$       & $99.413$       & $99.449$       & $99.459$       \\
		\botrule
		Unsupervised Methods                                                                                           \\		
		\midrule[0.01cm]
		Isolation Forest \citep{ref135}            & $52.407$       & $50.650$       & $49.926$       & $88.149$       \\
		Principal Component Analysis \citep{ref18} & $51.205$       & $50.282$       & $49.362$       & $87.480$       \\
		LSTM \citep{ref36}                         & $97.973$       & $98.843$       & $98.380$       & $98.505$       \\
		Transformer \citep{ref79}                  & $98.409$       & $99.100$       & $98.738$       & $98.837$       \\
		\botrule
		CoLog\footnotemark[1]                      & $\mathbf{100}$ & $\mathbf{100}$ & $\mathbf{100}$ & $\mathbf{100}$ \\														
		\botrule
	\end{tabular}
	\footnotetext[1]{CoLog is a supervised method.}
\end{table}

\begin{table}[h]
	\caption{CoLog in comparison to other log anomaly detection methods on the Jhuisi dataset. Bold numbers indicate the outstanding results.}\label{tab4}%
	\begin{tabular}{@{}lllll@{}}
		\toprule
		Anomaly Detection Technique                & Precision      & Recall         & F1-Score       & Accuracy       \\
		\botrule
		Supervised Methods                                                                                             \\
		\midrule[0.01cm]
		Logistic Regression \citep{ref134}         & $68.373$       & $66.127$       & $64.886$       & $79.182$       \\
		Support Vector Machines \citep{ref108}     & $63.643$       & $66.506$       & $64.051$       & $80.914$       \\
		Decision Tree \citep{ref109}               & $91.534$       & $89.769$       & $90.550$       & $93.313$       \\
		Attentional BiLSTM \citep{ref62}           & $95.123$       & $97.270$       & $96.169$       & $99.341$       \\
		Convolutional Neural Network \citep{ref45} & $97.139$       & $94.885$       & $95.982$       & $99.341$       \\
		pylogsentiment \citep{ref38}               & $98.867$       & $98.761$       & $98.813$       & $98.850$       \\
		\botrule
		Unsupervised Methods                                                                                           \\		
		\midrule[0.01cm]
		Isolation Forest \citep{ref135}            & $57.519$       & $52.774$       & $51.700$       & $74.707$       \\
		Principal Component Analysis \citep{ref18} & $44.828$       & $49.299$       & $45.014$       & $75.448$       \\
		LSTM \citep{ref36}                         & $96.879$       & $92.385$       & $94.508$       & $99.121$       \\
		Transformer \citep{ref79}                  & $96.879$       & $92.385$       & $94.508$       & $99.121$       \\
		\botrule
		CoLog\footnotemark[1]                      & $\mathbf{100}$ & $\mathbf{100}$ & $\mathbf{100}$ & $\mathbf{100}$ \\												
		\botrule
	\end{tabular}
	\footnotetext[1]{CoLog is a supervised method.}
\end{table}

\begin{table}[h]
	\caption{CoLog in comparison to other log anomaly detection methods on the Nssal dataset. Bold numbers indicate the outstanding results.}\label{tab5}%
	\begin{tabular}{@{}lllll@{}}
		\toprule
		Anomaly Detection Technique                & Precision         & Recall            & F1-Score          & Accuracy          \\
		\botrule
		Supervised Methods                                                                                                         \\
		\midrule[0.01cm]
		Logistic Regression \citep{ref134}         & $85.133$          & $74.728$          & $76.476$          & $97.604$          \\
		Support Vector Machines \citep{ref108}     & $80.206$          & $74.935$          & $76.474$          & $97.655$          \\
		Decision Tree \citep{ref109}               & $94.791$          & $87.700$          & $89.470$          & $98.063$          \\
		Attentional BiLSTM \citep{ref62}           & $96.750$          & $98.805$          & $97.754$          & $99.813$          \\
		Convolutional Neural Network \citep{ref45} & $96.703$          & $98.243$          & $97.460$          & $99.789$          \\
		pylogsentiment \citep{ref38}               & $97.170$          & $96.050$          & $96.602$          & $99.020$          \\
		\botrule
		Unsupervised Methods                                                                                                       \\		
		\midrule[0.01cm]
		Isolation Forest \citep{ref135}            & $65.504$          & $57.352$          & $56.101$          & $80.967$          \\
		Principal Component Analysis \citep{ref18} & $52.642$          & $53.827$          & $49.505$          & $80.614$          \\
		LSTM \citep{ref36}                         & $96.148$          & $97.669$          & $96.896$          & $99.742$          \\
		Transformer \citep{ref79}                  & $96.304$          & $99.354$          & $97.778$          & $99.813$          \\
		\botrule
		CoLog\footnotemark[1]                      & $\mathbf{99.955}$ & $\mathbf{99.915}$ & $\mathbf{99.935}$ & $\mathbf{99.967}$ \\														
		\botrule
	\end{tabular}
	\footnotetext[1]{CoLog is a supervised method.}
\end{table}

\begin{table}[h]
	\caption{CoLog in comparison to other log anomaly detection methods on the Honey7 dataset. Bold numbers indicate the outstanding results.}\label{tab6}%
	\begin{tabular}{@{}lllll@{}}
		\toprule
		Anomaly Detection Technique                & Precision      & Recall         & F1-Score       & Accuracy       \\
		\botrule
		Supervised Methods                                                                                             \\
		\midrule[0.01cm]
		Logistic Regression \citep{ref134}         & $68.143$       & $70.000$       & $69.042$       & $96.286$       \\
		Support Vector Machines \citep{ref108}     & $68.143$       & $70.000$       & $69.042$       & $96.286$       \\
		Decision Tree \citep{ref109}               & $93.260$       & $83.307$       & $86.359$       & $97.471$       \\
		Attentional BiLSTM \citep{ref62}           & $\mathbf{100}$ & $\mathbf{100}$ & $\mathbf{100}$ & $\mathbf{100}$          \\
		Convolutional Neural Network \citep{ref45} & $\mathbf{100}$ & $\mathbf{100}$ & $\mathbf{100}$ & $\mathbf{100}$          \\
		pylogsentiment \citep{ref38}               & $99.970$       & $99.107$       & $99.535$       & $99.943$       \\
		\botrule
		Unsupervised Methods                                                                                           \\		
		\midrule[0.01cm]
		Isolation Forest \citep{ref135}            & $47.509$       & $50.948$       & $48.064$       & $85.966$       \\
		Principal Component Analysis \citep{ref18} & $60.871$       & $58.898$       & $55.943$       & $88.279$       \\
		LSTM \citep{ref36}                         & $96.212$       & $98.810$       & $97.429$       & $98.155$       \\
		Transformer \citep{ref79}                  & $96.923$       & $99.048$       & $97.932$       & $98.524$       \\
		\botrule
		CoLog\footnotemark[1]                      & $\mathbf{100}$ & $\mathbf{100}$ & $\mathbf{100}$ & $\mathbf{100}$ \\												
		\botrule
	\end{tabular}
	\footnotetext[1]{CoLog is a supervised method.}
\end{table}

\begin{table}[h]
	\caption{CoLog in comparison to other log anomaly detection methods on the Zookeeper dataset. Bold numbers indicate the outstanding results.}\label{tab7}%
	\begin{tabular}{@{}lllll@{}}
		\toprule
		Anomaly Detection Technique                & Precision      & Recall         & F1-Score       & Accuracy       \\
		\botrule
		Supervised Methods                                                                                             \\
		\midrule[0.01cm]
		Logistic Regression \citep{ref134}         & $98.369$       & $98.464$       & $98.416$       & $98.562$       \\
		Support Vector Machines \citep{ref108}     & $97.987$       & $97.769$       & $97.877$       & $98.078$       \\
		Decision Tree \citep{ref109}               & $98.599$       & $98.880$       & $98.737$       & $98.851$       \\
		Attentional BiLSTM \citep{ref62}           & $95.783$       & $92.928$       & $94.300$       & $98.387$       \\
		Convolutional Neural Network \citep{ref45} & $95.121$       & $92.662$       & $93.850$       & $98.252$       \\
		pylogsentiment \citep{ref38}               & $99.722$       & $99.898$       & $99.810$       & $99.973$       \\
		\botrule
		Unsupervised Methods                                                                                           \\		
		\midrule[0.01cm]
		Isolation Forest \citep{ref135}            & $32.607$       & $50.000$       & $39.473$       & $65.215$       \\
		Principal Component Analysis \citep{ref18} & $79.011$       & $51.209$       & $42.121$       & $66.021$       \\
		LSTM \citep{ref36}                         & $95.825$       & $91.887$       & $93.750$       & $98.252$       \\
		Transformer \citep{ref79}                  & $99.890$       & $99.416$       & $99.652$       & $99.361$       \\
		\botrule
		CoLog\footnotemark[1]                      & $\mathbf{100}$ & $\mathbf{100}$ & $\mathbf{100}$ & $\mathbf{100}$ \\												
		\botrule
	\end{tabular}
	\footnotetext[1]{CoLog is a supervised method.}
\end{table}

\begin{table}[h]
	\caption{CoLog in comparison to other log anomaly detection methods on the Hadoop dataset. Bold numbers indicate the outstanding results.}\label{tab8}%
	\begin{tabular}{@{}lllll@{}}
		\toprule
		Anomaly Detection Technique                & Precision         & Recall            & F1-Score          & Accuracy          \\
		\botrule
		Supervised Methods                                                                                                         \\
		\midrule[0.01cm]
		Logistic Regression \citep{ref134}         & $48.523$          & $50.000$          & $49.250$          & $97.046$          \\
		Support Vector Machines \citep{ref108}     & $48.523$          & $50.000$          & $49.250$          & $97.046$          \\
		Decision Tree \citep{ref109}               & $48.523$          & $50.000$          & $49.250$          & $97.046$          \\
		Attentional BiLSTM \citep{ref62}           & $97.640$          & $97.955$          & $97.792$          & $97.902$          \\
		Convolutional Neural Network \citep{ref45} & $99.719$          & $99.847$          & $99.783$          & $99.955$          \\
		pylogsentiment \citep{ref38}               & $99.886$          & $99.732$          & $99.809$          & $99.905$          \\
		\botrule
		Unsupervised Methods                                                                                                       \\		
		\midrule[0.01cm]
		Isolation Forest \citep{ref135}            & $47.702$          & $50.000$          & $48.824$          & $54.034$          \\
		Principal Component Analysis \citep{ref18} & $49.995$          & $49.996$          & $49.996$          & $58.214$          \\
		LSTM \citep{ref36}                         & $99.850$          & $97.397$          & $98.589$          & $99.715$          \\
		Transformer \citep{ref79}                  & $97.280$          & $99.833$          & $98.518$          & $99.685$          \\
		\botrule
		CoLog\footnotemark[1]                      & $\mathbf{99.997}$ & $\mathbf{99.956}$ & $\mathbf{99.977}$ & $\mathbf{99.994}$ \\														
		\botrule
	\end{tabular}
	\footnotetext[1]{CoLog is a supervised method.}
\end{table}

\begin{table}[h]
	\caption{CoLog in comparison to other log anomaly detection methods on the BlueGene/L dataset. Bold numbers indicate the outstanding results.}\label{tab9}%
	\begin{tabular}{@{}lllll@{}}
		\toprule
		Anomaly Detection Technique                & Precision         & Recall            & F1-Score          & Accuracy          \\
		\botrule
		Supervised Methods                                                                                                         \\
		\midrule[0.01cm]
		Logistic Regression \citep{ref134}         & $54.028$          & $51.852$          & $52.092$          & $90.368$          \\
		Support Vector Machines \citep{ref108}     & $46.314$          & $50.000$          & $48.087$          & $92.628$          \\
		Decision Tree \citep{ref109}               & $60.576$          & $50.998$          & $50.303$          & $92.348$          \\
		Attentional BiLSTM \citep{ref62}           & $97.640$          & $97.955$          & $97.792$          & $97.902$          \\
		Convolutional Neural Network \citep{ref45} & $97.640$          & $97.955$          & $97.792$          & $97.902$          \\
		pylogsentiment \citep{ref38}               & $99.892$          & $99.963$          & $99.928$          & $99.980$          \\
		\botrule
		Unsupervised Methods                                                                                                       \\		
		\midrule[0.01cm]
		Isolation Forest \citep{ref135}            & $53.081$          & $50.047$          & $51.519$          & $47.389$          \\
		Principal Component Analysis \citep{ref18} & $51.168$          & $54.260$          & $38.970$          & $48.487$          \\
		LSTM \citep{ref36}                         & $97.414$          & $98.296$          & $97.806$          & $97.902$          \\
		Transformer \citep{ref79}                  & $97.640$          & $97.955$          & $97.792$          & $97.902$          \\
		\botrule
		CoLog\footnotemark[1]                      & $\mathbf{99.999}$ & $\mathbf{99.990}$ & $\mathbf{99.994}$ & $\mathbf{99.998}$ \\														
		\botrule
	\end{tabular}
	\footnotetext[1]{CoLog is a supervised method.}
\end{table}

The F1-scores of all techniques are presented at the mean level in Table~\ref{tab10}. Of all the methods, pylogsentiment demonstrates superior performance due to its balancing between the two sentiment classes. This results in a more effective deep learning model for reliably identifying anomalous activities within the minority class. The pylogsentiment achieves an average F1-score of 99.135. The performance of other deep learning-based models, including Transformer, CNN, attentional BiLSTM, and LSTM, on all datasets are similar. For instance, they yield similar mean F1-scores of 97.845, 97.812, 97.661, and 96.765, respectively. These methods exhibit enhanced performance due to their utilization of deep learning for anomaly detection. Deep learning is adept at anomaly detection in logs due to its capability to manage unstructured data and autonomously extract features. It effectively handles large data volumes and, when employing log semantics, adapts to evolving abnormalities over time. Transformer produces better results because it uses word-level representations to represent log entries. In Table~\ref{tab10}, the IF attained a mean F1-score of 49.372, whereas PCA algorithm recorded a mean F1-score of 47.273. Generally, machine learning methods often struggle with log anomaly detection due to the complexity and unstructured nature of log data, distribution shifts introducing unseen anomalies, and the need for sophisticated feature engineering. Additionally, these models can be sensitive to hyperparameter settings, requiring extensive tuning.

\begin{table}[h]
	\caption{Ranking CoLog and other log anomaly detection methods. Bold numbers indicate the outstanding results.}\label{tab10}%
	\begin{tabular}{@{}llc@{}}
		\toprule
		Anomaly Detection                          & & Mean              \\
		Technique                                  & & F1-score          \\
		\midrule[0.01cm]
		Principal Component Analysis \citep{ref18} & & $47.273$          \\
		Isolation Forest \citep{ref135}            & & $49.372$          \\
		Support Vector Machines \citep{ref108}     & & $66.323$          \\
		Logistic Regression \citep{ref134}         & & $67.123$          \\
		Decision Tree \citep{ref109}               & & $77.737$          \\
		LSTM \citep{ref36}                         & & $96.765$          \\
		Attentional BiLSTM \citep{ref62}           & & $97.661$          \\
		Convolutional Neural Network \citep{ref45} & & $97.812$          \\
		Transformer \citep{ref79}                  & & $97.845$          \\
		pylogsentiment \citep{ref38}               & & $99.135$          \\
		\botrule
		CoLog                                      & & $\mathbf{99.987}$ \\														
		\botrule
	\end{tabular}
\end{table}

Table~\ref{tab3} to Table~\ref{tab9} indicate that CoLog yields superior performance compared to other techniques across all datasets. Moreover, Table~\ref{tab10} clearly illustrates that CoLog significantly improves anomaly identification. Compared to the second-ranked pylogsentiment, it demonstrates a greater mean F1-score across all datasets. Given that the pylogsentiment has a mean F1-score of 99.135, a further improvement of 0.865 is required to attain perfect anomaly detection on the employed benchmark datasets. CoLog decreases this value to 0.013. This signifies an estimated 98.50\% improvement in the available range to enhance the results.

In many cases, unsupervised approaches frequently exhibit inferior performance compared to supervised techniques in log anomaly detection. This is attributable to the absence of labeled data, which limits understanding of complex patterns and reduces accuracy. They also tend to have more false positives. Supervised approaches enhance their accuracy through training on labeled data over time. In this work, we frame log anomaly detection as a multimodal sentiment analysis task, transforming the problem from traditional unsupervised anomaly detection to a supervised classification setting. Conventional anomaly detection methods often treat anomalies as rare or outlier events without explicit labels, which limits the ability of models to learn discriminative features effectively due to the highly imbalanced and variable nature of log data. By interpreting anomalies as negative sentiments within log messages, our approach leverages sentiment analysis techniques to annotate log entries with sentiment labels, thereby enabling supervised learning. This shift allows the model to utilize labeled data for training, improving detection accuracy by capturing subtle semantic patterns and contextual nuances that indicate anomalous system behavior. Consequently, our multimodal sentiment analysis framework benefits from both the rich semantic representation of logs and the robust classification capabilities of supervised learning models, leading to enhanced detection performance compared to purely unsupervised approaches.

Additionally, CoLog's exceptional efficacy in anomaly detection is due to its novel methodology for learning log data via both sequence and semantic modalities. In contrast to conventional approaches that may focus on a single facet of log data, CoLog incorporates multiple modalities to produce a more comprehensive representation of log data. This dual-modality learning enables CoLog to gain a more thorough understanding of log data that might otherwise be overlooked. Another significant component enhancing CoLog's efficacy is its collaborative approach to acquiring log representations. Unlike previous approaches that may analyze logs in isolation or with minimal interaction across various data modalities, CoLog guarantees the simultaneous learning of sequence and semantic information. This collaborative method yields more comprehensive and refined representations of log data, thereby enhancing the model's ability to distinguish between normal and abnormal activity. Jointly examining logs facilitates enhanced adaptability to various log contexts and scenarios. This versatility is essential in practical applications where logs may differ markedly in form and content. In addition, in many traditional methods, the integration of various data modalities can lead to noise and inconsistencies, ultimately degrading performance. One of the standout features of CoLog is its modality adaptation layer, which plays a crucial role in refining its performance by removing impurities that may arise from combining different modalities. This layer acts as a filter, ensuring that only relevant and complementary features are fused. By eliminating conflicting or redundant information, the MAL prevents the degradation of performance that could occur if modalities were combined without such careful consideration. CoLog's ability to filter out these impurities ensures that the resulting log representations are cleaner and more reliable, enhancing the overall accuracy of the analysis. As a result, CoLog is better equipped to handle the variability and complexity inherent in log data, making it a more effective tool for administrators in the field. CoLog also utilizes a latent space for meaningfully fusing different modalities of log data. Instead of simply concatenating features from each modality or processing them separately, CoLog explores a shared latent space where these features can be integrated effectively. By mapping log data into a common latent space, CoLog can capture complex relationships and dependencies that are essential for accurate anomaly detection.

Table~\ref{inferencetimetable} evaluates the efficiency of different models by recording the time spent on training and testing processes on the Hadoop dataset. The LSTM, Attentional BiLSTM, CNN, and Transformer methodologies necessitate considerably less time for training and inference compared to pylogsentiment and CoLog. This results from utilizing log keys rather than the semantic features of logs, which require far fewer computer resources. CoLog requires less training time than pylogsentiment, although it is inferior during the inference phase.

\begin{table}[h]
	\caption{Efficiency of deep learning models on Hadoop Dataset.}\label{inferencetimetable}%
	\begin{tabular*}{\textwidth}{@{\extracolsep\fill}lcccccc}
		\toprule                                   & \multicolumn{2}{@{}c@{}}{Training Time} & \multicolumn{2}{@{}c@{}}{Inference Time} \\
		                                           & \multicolumn{2}{@{}c@{}}{(Seconds)}     & \multicolumn{2}{@{}c@{}}{(Seconds)}      \\
		\cmidrule{2-5}%
		Anomaly Detection Technique                & Overall              & Per-sample       & Overall              & Per-sample        \\
		\midrule[0.01cm]
		LSTM \citep{ref36}                         & $7.72$               & $0.00112$         & $0.14$               & $0.00002$        \\
		\midrule[0.01cm]
		Attentional BiLSTM \citep{ref62}           & $25.71$              & $0.00372$         & $0.21$               & $0.00003$        \\
		\midrule[0.01cm]
		Convolutional Neural Network \citep{ref45} & $14.85$              & $0.00215$         & $0.07$               & $0.00001$        \\
		\midrule[0.01cm]
		Transformer \citep{ref79}                  & $9.05$               & $0.00131$         & $0.09$               & $0.00001$        \\
		\midrule[0.01cm]
		pylogsentiment \citep{ref38}               & $2615.29$            & $0.01138$         & $94.78$              & $0.00262$        \\
		\botrule
		CoLog                                      & $2479.02$            & $0.01079$         & $124.00$             & $0.00343$        \\
		\botrule
	\end{tabular*}
\end{table}

\subsubsection{Enhancing Success: Study on Hyperparameters Tuning}\label{subsubsec5: enhancingsuccessstudyonhyperparameterstuning}

Hyperparameters tuning is essential in deep learning as it directly influences model performance and generalization \citep{ref136}. This process can substantially improve model accuracy and efficiency \citep{ref137}. Appropriate tuning addresses challenges such as overfitting or underfitting, which is essential in noisy and mutation-prone log data. Furthermore, it facilitates the examination of various model configurations, resulting in more resilient and dependable results. Considering the importance of tuning, specifically in log anomaly detection, we employ the Ray package in Python to tune CoLog. In the tuning procedure, we utilize three datasets, Casper, Jhuisi, and Honey7, and apply 324 distinct configurations of the CoLog on them. Based on the tuning results of the Casper, Jhuisi, and Honey7 datasets, the hyperparameters selection criterion for each dataset is to achieve optimal accuracy in the shortest time. The overall optimal configuration is the resultant of all datasets' optimum configurations. Due to page constraints, this paper only covers the 27 most efficient runs out of 108 runs for each dataset (Complete tuning results are available on CoLog's GitHub repository.). The chosen hyperparameters are detailed in Section~\ref{subsubsec5: hyperparameters}.

\begin{figure}[h]
	\centering
	\includegraphics[width=0.8\textwidth]{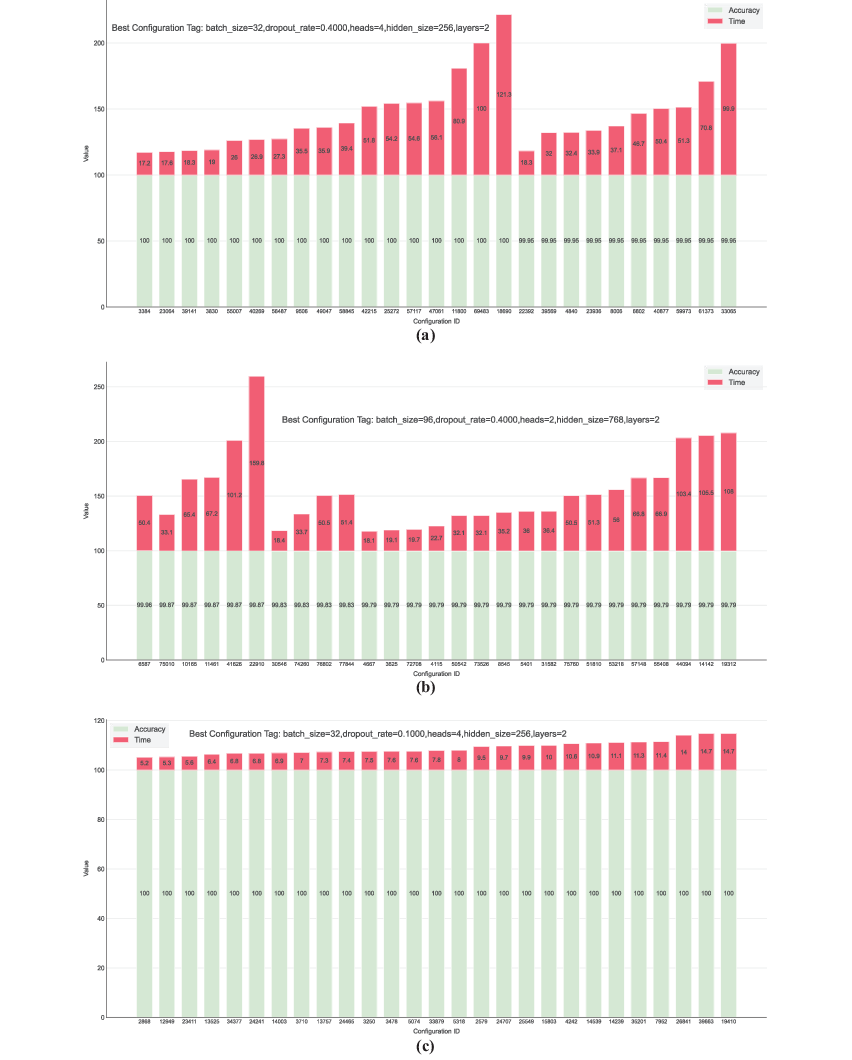}
	\caption{Collection of graphical illustrations of hyperparameters tuning practices performed by the CoLog on the (a) Casper, (b) Jhuisi, and (c) Honey7 datasets. The criterion for selecting the optimal configuration is to achieve the highest accuracy in the shortest time. Based on this criterion, the results are sorted from left to right. The optimal configuration is also depicted in the respective diagram for each dataset.}\label{fig15}
\end{figure}

Figure~\ref{fig15} illustrates the results of the top 27 executions of CoLog on the Casper, Jhuisi, and Honey7 datasets, arranged from left to right. The light green bars signify the model's accuracy in that configuration of CoLog, while the rose bars represent the total time required to attain that accuracy. Accordingly, configurations 3384, 6587, and 2868 have achieved optimal outcomes in the shortest time for Casper, Jhuisi, and Honey7 datasets, respectively. Figure~\ref{fig15} also illustrates the tag for the optimum configurations of each dataset.

\subsubsection{Consequences of Ratio: Examination of Train Ratio Impact}\label{subsubsec5: consequencesofratioexaminationoftrainratioimpact}

\begin{figure}[h]
	\centering
	\includegraphics[width=1\textwidth]{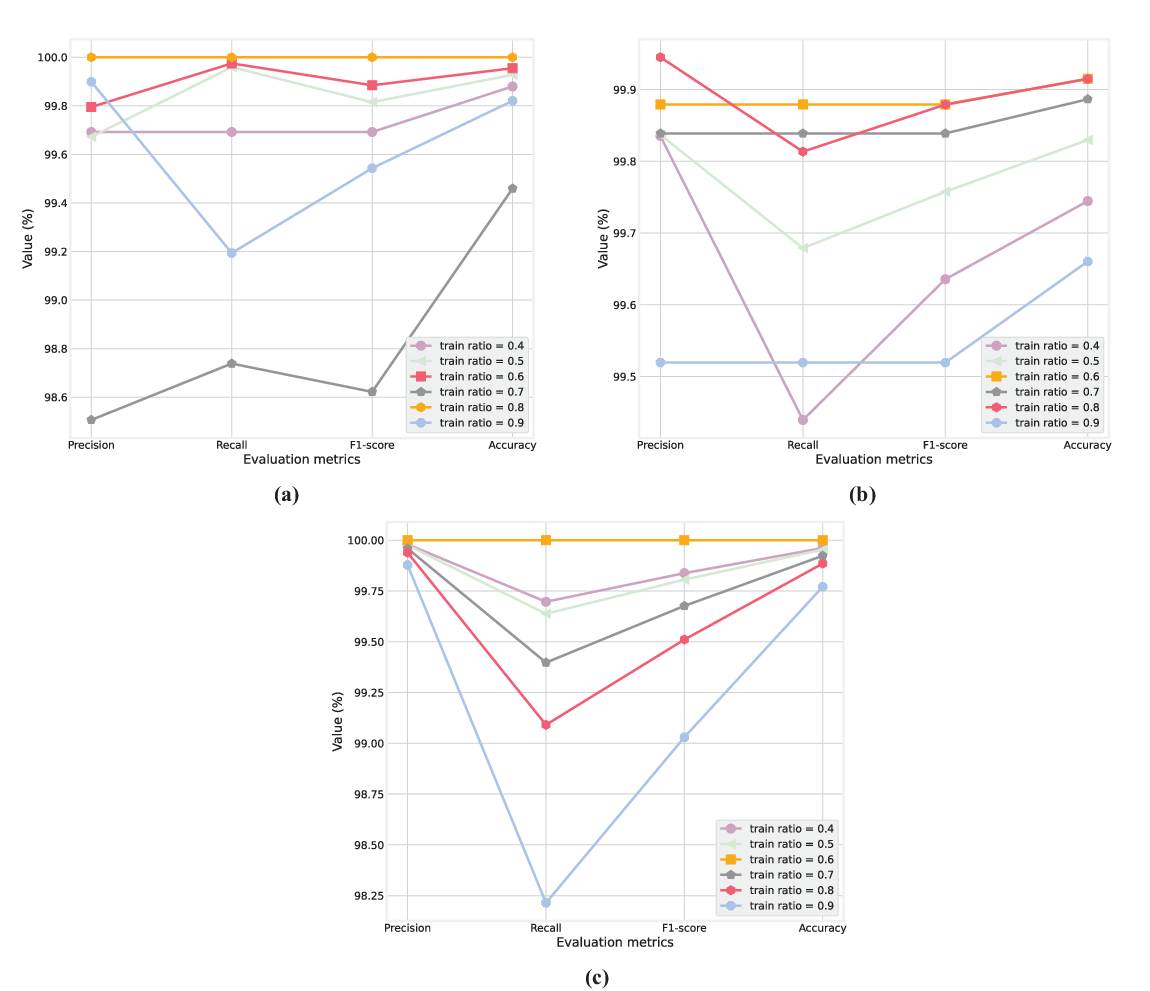}
	\caption{Visual illustrations of various train-test configurations performed by the CoLog on the (a) Casper, (b) Jhuisi, and (c) Honey7 datasets.}\label{fig16}
\end{figure}

Inadequate and imbalanced training data, such as log data, can result in overfitting, wherein the model is talented in training data but fails on novel data \citep{ref138}. Furthermore, models trained with a suitable train-test ratio generally exhibit superior performance \citep{ref139}. Consequently, determining an optimal train-test data ratio is essential. The percentage of training data influences the model's robustness to noise and data mutation, which are abundantly seen in the log data. An optimally selected ratio enables the model to learn underlying patterns in the data while maintaining robustness against outliers and noise. This is particularly vital in practical applications, for example, log-based anomaly detection, where data may be noisy and unreliable. We assess the influence of various train ratios by applying CoLog on the Casper, Jhuisi, and Honey7 datasets, which have been preprocessed with train ratios of 0.4, 0.5, 0.6, 0.7, 0.8, and 0.9. Figure~\ref{fig16} illustrates the outcomes of analyzing various training data rates. The optimum result occurs when CoLog is applied to the Jhuisi and Honey7 datasets with a training ratio of 0.6. The optimal ratio for the Casper dataset is 0.8. Based on the above mentioned ratios, the F1-score is 100 for the Casper dataset, 99.879 for the Jhuisi dataset, and 100 for the Honey7 dataset. Finding a proper training ratio is crucial in log anomaly detection. The ideal proportion of training data substantially affects the convergence time of deep learning models, which is vital in this context. A balanced ratio guarantees the model obtains diverse and representative samples during training, accelerating the learning process. An unbalanced ratio may result in slow convergence or cause it to become bound in local minima.

\subsubsection{Balancing Endeavor: Comparative Study on Class Imbalance Solving Methods}\label{subsubsec5: balancingendeavorcomparativestudyonclassimbalancesolvingmethods}

Data imbalance plays a crucial role in deep learning, profoundly affecting model performance and generalization. Imbalanced training data causes models to exhibit bias towards the majority class, resulting in suboptimal performance on the minority class \citep{ref140, ref141}. This poses significant challenges in automated log anomaly detection, where the minority class frequently signifies critical instances \citep{ref38}. Mitigating data imbalance enables deep learning models to generate precise predictions across all imbalanced classes, augmenting their practical applicability in real-world scenarios. Numerous approaches have been devised to alleviate the impact of data imbalance, focusing on either equalizing class distribution or modifying the learning process to prioritize the minority class \citep{ref117, ref118, ref119, ref120, ref121, ref122}.

\begin{figure}[h]
	\centering
	\includegraphics[width=1\textwidth]{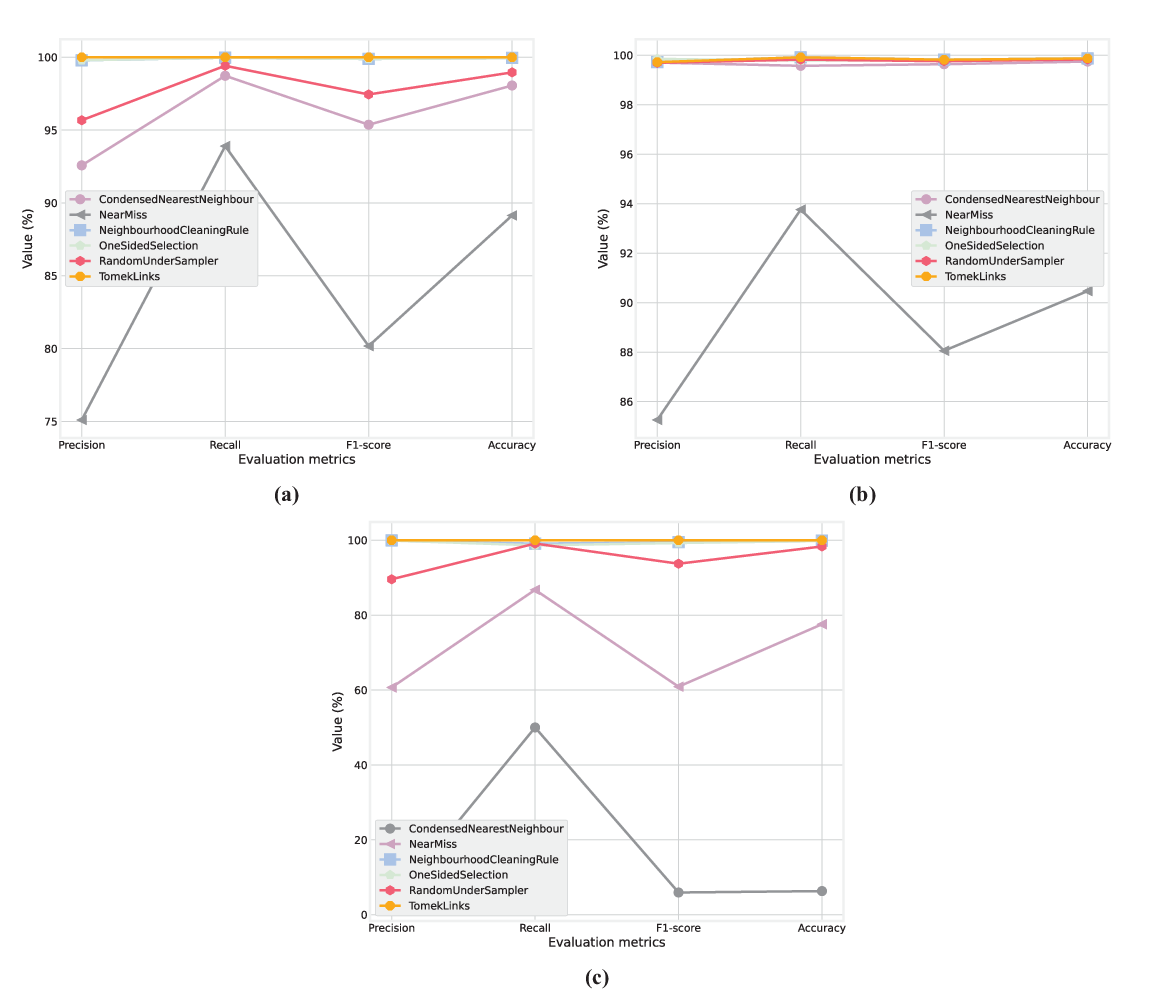}
	\caption{Comparison of Tomek link with other class imbalance solving methods performed by the CoLog on the (a) Casper, (b) Jhuisi, and (c) Honey7 datasets.}\label{fig17}
\end{figure}

CoLog employs under-sampling techniques due to the similarity among numerous log entries in the majority class. We implement class imbalance solving across all datasets with the Tomek link approach. Figure~\ref{fig17} illustrates the efficacy of the Tomek link relative to various class balancing techniques, including the condensed nearest neighbor rule \citep{ref142}, NearMiss \citep{ref119}, neighborhood cleaning rule \citep{ref143}, one-sided selection \citep{ref144}, and random under-sampling across the Casper, Jhuisi, and Honey7 datasets.

Figure~\ref{fig17} illustrates that the Tomek link exhibited the highest mean values across all datasets for precision, recall, F1-score, and accuracy, achieving 99.908\%, 99.972\%, 99.940\%, and 99.957\%, respectively. In the Honey7 dataset, the outcomes of the Tomek link method surpass slightly those of alternative techniques. However, the condensed nearest neighbor rule and NearMiss algorithms exhibit poor performance on this dataset. Except for NearMiss, all approaches yield successful results on the Jhuisi dataset. The Tomek link and neighborhood cleaning rule techniques also produce the same results in the Jhuisi dataset. Note that the Tomek link approach attains flawless results of 100\% in the Casper dataset.

\subsubsection{Marvels of Window Sizes: Study on the Effect of Window Size}\label{subsubsec5: marvelsofwindowsizesstudyontheeffectofwindowsize}

We utilize window size 1 with a context sequence type across all datasets. Table~\ref{tab11} to Table~\ref{tab13} illustrate the performance of the context-based window size 1 in contrast to various window sizes and sequence types on the Casper, Jhuisi, and Honey7 datasets.

\begin{table}[h]
	\caption{Comparative analysis of CoLog performance across various window sizes and sequence types within the Casper dataset. Bold numbers indicate the outstanding results.}\label{tab11}
	\begin{tabular*}{\textwidth}{@{\extracolsep\fill}llllll|lllll}
		\toprule
		                                  & \multicolumn{10}{c}{Sequence Type}                                                                                     \\
		\cmidrule[0.01cm]{2-11}
		                                  & \multicolumn{5}{c}{Background}     & \multicolumn{5}{c}{Context}                                                       \\
		\cmidrule[0.01cm]{2-5} \cmidrule[0.01cm]{5-11}
		\multicolumn{1}{c}{Window Size}   & 1       & 2       & 3       & 6              & 9       & 1              & 2              & 3       & 6       & 9       \\
		\midrule[0.005cm]
		Precision                         & $99.80$ & $99.80$ & $99.80$ & $\mathbf{100}$ & $99.80$ & $\mathbf{100}$ & $\mathbf{100}$ & $99.80$ & $99.80$ & $99.80$ \\
		Recall                            & $99.97$ & $99.97$ & $99.97$ & $\mathbf{100}$ & $99.97$ & $\mathbf{100}$ & $\mathbf{100}$ & $99.97$ & $99.97$ & $99.97$ \\
		F1-Score                          & $99.88$ & $99.88$ & $99.88$ & $\mathbf{100}$ & $99.88$ & $\mathbf{100}$ & $\mathbf{100}$ & $99.88$ & $99.88$ & $99.88$ \\
		Accuracy                          & $99.95$ & $99.95$ & $99.95$ & $\mathbf{100}$ & $99.95$ & $\mathbf{100}$ & $\mathbf{100}$ & $99.95$ & $99.95$ & $99.95$ \\
		\botrule
	\end{tabular*}
\end{table}

\begin{table}[h]
	\caption{Comparative analysis of CoLog performance across various window sizes and sequence types within the Jhuisi dataset. Bold numbers indicate the outstanding results.}\label{tab12}
	\begin{tabular*}{\textwidth}{@{\extracolsep\fill}llllll|lllll}
		\toprule
		                                  & \multicolumn{10}{c}{Sequence Type}                                                                       \\
		\cmidrule[0.01cm]{2-11}
		                                  & \multicolumn{5}{c}{Background}     & \multicolumn{5}{c}{Context}                                         \\
		\cmidrule[0.01cm]{2-5} \cmidrule[0.01cm]{5-11}
		\multicolumn{1}{c}{Window Size}   & 1       & 2       & 3       & 6       & 9       & 1              & 2       & 3       & 6       & 9       \\
		\midrule[0.005cm]
		Precision                         & $99.97$ & $99.76$ & $99.94$ & $99.88$ & $99.78$ & $\mathbf{100}$ & $99.92$ & $99.52$ & $99.85$ & $99.75$ \\
		Recall                            & $99.91$ & $99.76$ & $99.81$ & $99.88$ & $99.25$ & $\mathbf{100}$ & $99.72$ & $99.52$ & $99.79$ & $99.16$ \\
		F1-Score                          & $99.94$ & $99.76$ & $99.88$ & $99.88$ & $99.51$ & $\mathbf{100}$ & $99.82$ & $99.52$ & $99.82$ & $99.45$ \\
		Accuracy                          & $99.96$ & $99.83$ & $99.91$ & $99.91$ & $99.66$ & $\mathbf{100}$ & $99.87$ & $99.66$ & $99.87$ & $99.62$ \\
		\botrule
	\end{tabular*}
\end{table}

\begin{table}[h]
	\caption{. Comparative analysis of CoLog performance across various window sizes and sequence types within the Honey7 dataset. Bold numbers indicate the outstanding results.}\label{tab13}
	\begin{tabular*}{\textwidth}{@{\extracolsep\fill}llllll|lllll}
		\toprule
		                                  & \multicolumn{10}{c}{Sequence Type}                                                                                                          \\
		\cmidrule[0.01cm]{2-11}
		                                  & \multicolumn{5}{c}{Background}     & \multicolumn{5}{c}{Context}                                                                            \\
		\cmidrule[0.01cm]{2-5} \cmidrule[0.01cm]{5-11}
		\multicolumn{1}{c}{Window Size}   & 1       & 2              & 3              & 6              & 9              & 1              & 2       & 3              & 6       & 9       \\
		\midrule[0.005cm]
		Precision                         & $99.97$ & $\mathbf{100}$ & $\mathbf{100}$ & $\mathbf{100}$ & $\mathbf{100}$ & $\mathbf{100}$ & $99.97$ & $\mathbf{100}$ & $99.97$ & $99.97$ \\
		Recall                            & $99.55$ & $\mathbf{100}$ & $\mathbf{100}$ & $\mathbf{100}$ & $\mathbf{100}$ & $\mathbf{100}$ & $99.55$ & $\mathbf{100}$ & $99.55$ & $99.55$ \\
		F1-Score                          & $99.76$ & $\mathbf{100}$ & $\mathbf{100}$ & $\mathbf{100}$ & $\mathbf{100}$ & $\mathbf{100}$ & $99.76$ & $\mathbf{100}$ & $99.76$ & $99.76$ \\
		Accuracy                          & $99.94$ & $\mathbf{100}$ & $\mathbf{100}$ & $\mathbf{100}$ & $\mathbf{100}$ & $\mathbf{100}$ & $99.94$ & $\mathbf{100}$ & $99.94$ & $99.94$ \\
		\botrule
	\end{tabular*}
\end{table}

As shown in Tables 11 to 13, among all the datasets evaluated with varying window sizes, the context-based window size 1 yields the most optimal results at the lowest cost. In the Casper (i.e. Table~\ref{tab11}) dataset, all window sizes yield acceptable yet same results, except the background-based window size of 6 and the context-based window sizes of 1, 2, and 12. The above mentioned exceptions attain a 100\% result in the evaluation metrics. In the Jhuisi dataset (i.e. Table~\ref{tab12}), all results are independently significant, with optimal outcomes achieved with a context-based window size of 1. For Honey7 dataset (i.e. Table~\ref{tab13}), background-based window sizes of 2, 3, 6, and 9 yield 100\% among all metrics, while context-based window sizes of 1 and 3 also obtain 100\%. The rest configurations attain the same results of 99.97\%, 99.55\%, 99.76\%, and 99.94\% for precision, recall, F1-score, and accuracy, respectively.

\subsubsection{Log Landscapes: Visualization of CoLog's Output Vectors}\label{subsubsec5: loglandscapesvisualizationofcoLogsoutputvectors}

\begin{figure}[h]
	\centering
	\includegraphics[width=1\textwidth]{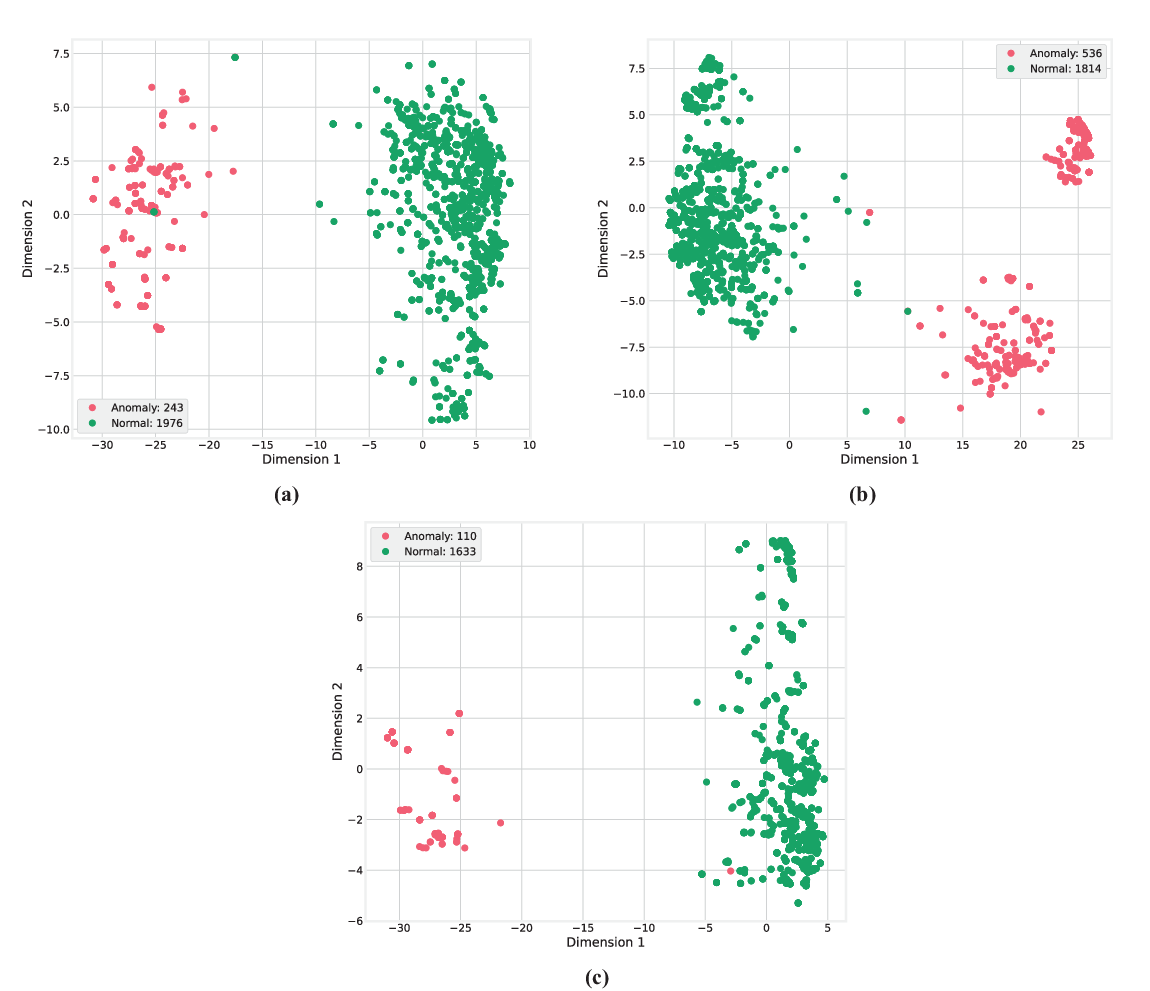}
	\caption{Visual illustrations of the CoLog's output vectors on the (a) Casper, (b) Jhuisi, and (c) Honey7 datasets utilizing PCA.}\label{fig18}
\end{figure}

In this experiment, we collect the learned vector representations of log messages from the pre-trained CoLog. Figure~\ref{fig18} illustrates the log vector representations on the Casper, Jhuisi, and Honey7 datasets, demonstrating their lower-dimensional representation of the test splits through the PCA dimensionality reduction technique. We prove that the log vector representations are classified accurately. The normal samples are concentrated closely together. Most anomalies are distributed further than normal samples and clustered together, where there is less error. Consequently, optimal performance could be achieved by applying an argmax prediction function.

\subsubsection{Resilience: Study on Generalization and Robustness}\label{subsubsec5: resiliencestudyongeneralizationandrobustness}

We look at the robustness and ability to be generalized of CoLog in identifying unknown abnormalities in previously untrained datasets. The datasets comprise Spark, Honey5, and Windows logs. Nevertheless, CoLog continues to demonstrate excellent results in identifying unknown abnormalities in addition to known anomalies, as demonstrated in Table~\ref{tab14}. CoLog attained F1-scores of 98.83\% and 99.12\% on the Honey5 and Windows datasets, respectively. CoLog has experienced training on several datasets and is capable of addressing unidentified abnormalities. There are three primary interpretations. Initially, we employ word2vec word embeddings to represent each log message's words. word2vec effectively captures the meanings of similar words, such as "warning" and "stopping". Secondly, we utilize sentence-transformers to derive analogous sentence embeddings for similar log messages. Third, we employ a supervised MSA model in a collaborative manner that attains superior performance on both known and novel datasets.

\begin{table}[h]
	\caption{The evaluation of CoLog on previously unobserved datasets.}\label{tab14}%
	\begin{tabular}{@{}l|l|l|l|l|l|l|l@{}}
		\toprule
		Dataset                & Average & Class      & \# Samples & Precision & Recall   & F1-Score & Accuracy \\
		\botrule
		Honey5                 &         & 0: Anomaly & 56588      & $98.375$  & $99.731$ & $99.049$ & $99.129$ \\
		\cmidrule{3-7}%
		                       &         & 1: Normal  & 67798      & $99.773$  & $98.625$ & $99.196$ &          \\
		\cmidrule{2-7}%
		                       & Macro   &            & 124386     & $99.074$  & $99.178$ & $99.122$ &          \\
		\botrule
		Windows                &         & 0: Anomaly & 6404       & $99.855$  & $96.705$ & $98.255$ & $99.119$ \\
        \cmidrule{3-7}%
                               &         & 1: Normal  & 18567      & $98.876$  & $99.952$ & $99.411$ &          \\
        \cmidrule{2-7}%
                               & Macro   &            & 24971      & $99.365$  & $98.328$ & $98.833$ &          \\
        \botrule
	\end{tabular}
\end{table}

\begin{figure}[h]
	\centering
	\includegraphics[width=0.5\textwidth]{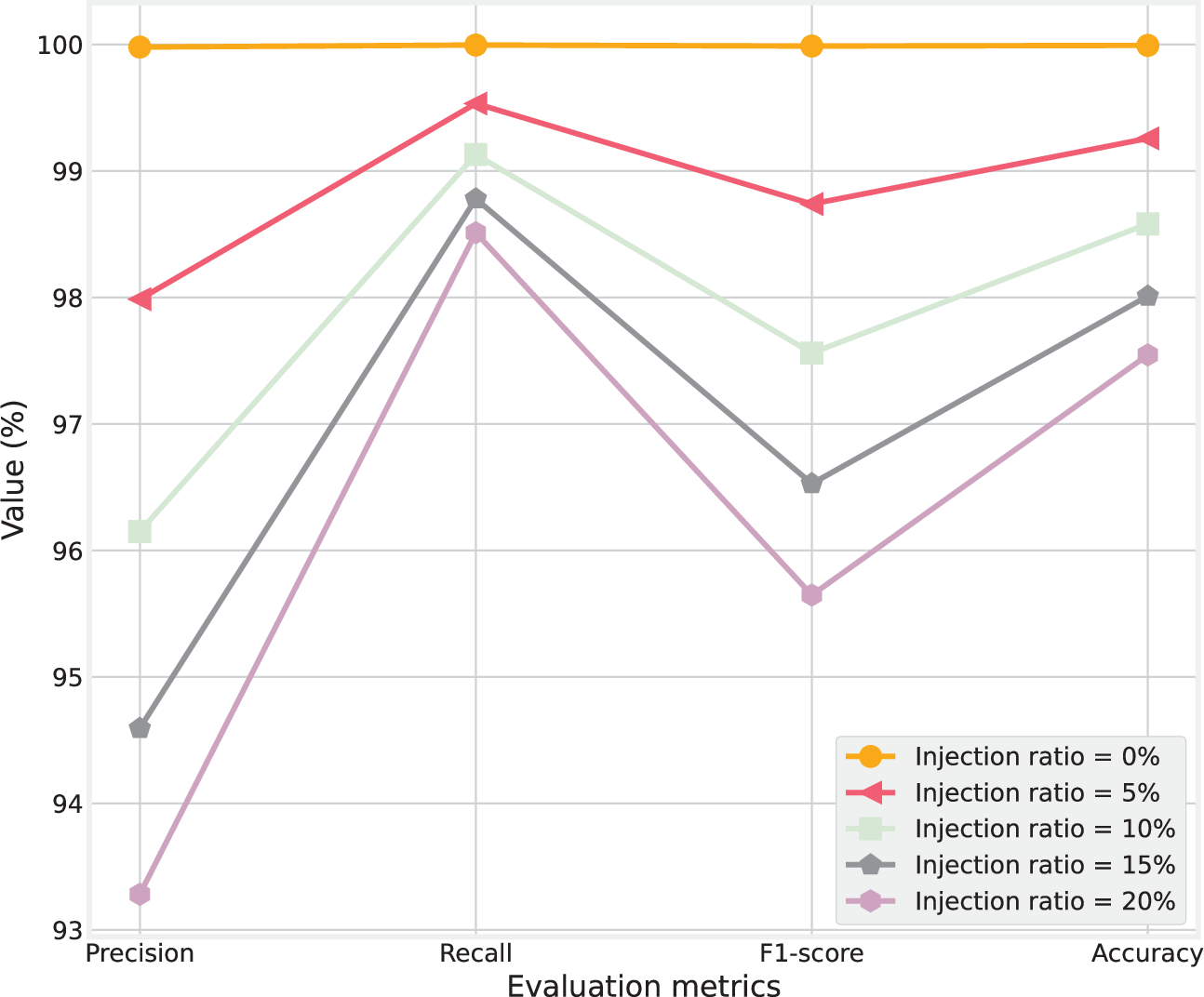}
	\caption{The visual illustration of CoLog's outcomes for various injection ratios of unstable log events on the Spark dataset.}\label{fig19}
\end{figure}

Furthermore, we perform robustness experiments on the Spark Dataset. We train and evaluate CoLog using the original Spark dataset, specifically the training set comprising 10\% and the validation set including 45\% of the total data. Subsequently, we test the trained model on the remaining 45\% of the dataset. This synthetic dataset incorporates new log events injected at ratios of 0\%, 5\%, 10\%, 15\%, and 20\% from alternative datasets. The results of the experiment are demonstrated in Figure~\ref{fig19}. CoLog shows commendable recall metrics. Despite the rising injection ratio of unstable log events, CoLog consistently achieves a high recall rate even at elevated injection levels. For instance, CoLog can maintain a recall of 98.51\% at an injection ratio of 20\%. It validates that our methodology is adequately adaptable to unstable log events. CoLog interprets all log events as semantic vectors, enabling it to detect unstable log events with similar semantic meanings. As a result, CoLog can efficiently analyze new log events and could keep on operating on the dataset containing unseen or unstable log events.

\subsubsection{Unraveling the Impact: Ablation Study}\label{subsubsec5: unravelingtheimpactablationstudy}

An ablation study was performed on the Casper, Jhuisi, and Honey7 datasets, with outcomes summarized in Table~\ref{tab15}.

\begin{table}[h]
	\caption{Ablation study of CoLog. Seq and Sem denote sequence and semantic, respectively. Also, CT, MHIA, MAL, and BL denote collaborative transformer, multi-head impressed attention, modality adaptation layer, and balancing layer, respectively.}\label{tab15}%
	\begin{tabular*}{\textwidth}{@{\extracolsep\fill}lcc|ccc|cccc}
		\toprule          & \multicolumn{2}{@{}c@{}|}{Modality} & \multicolumn{3}{@{}c@{}|}{Module} & \multicolumn{4}{@{}c@{}}{Metric}      \\
		\cmidrule{2-10}%
		Dataset           & Seq & Sem                & MHIA     & MAL      & BL       & Precision & Recall & F1-score & Accuracy  \\
		\midrule[0.01cm]
		Casper            & \cellcolor[HTML]{f4729c} $\times$ & \cellcolor[HTML]{a7dfb9} $\checkmark$                & \cellcolor[HTML]{f4729c} $\times$ & \cellcolor[HTML]{f4729c} $\times$ & \cellcolor[HTML]{f4729c} $\times$ & $98.80$ & $99.85$ & $99.31$ & $99.73$ \\
		\cmidrule{2-10}%
		                  & \cellcolor[HTML]{a7dfb9} $\checkmark$ & \cellcolor[HTML]{f4729c} $\times$                & \cellcolor[HTML]{f4729c} $\times$ & \cellcolor[HTML]{f4729c} $\times$ & \cellcolor[HTML]{f4729c} $\times$ & $98.41$ & $99.10$ & $98.74$ & $98.84$ \\
		\cmidrule{2-10}%
		                  & \cellcolor[HTML]{a7dfb9} $\checkmark$ & \cellcolor[HTML]{a7dfb9} $\checkmark$                & \cellcolor[HTML]{a7dfb9} $\checkmark$ & \cellcolor[HTML]{f4729c} $\times$ & \cellcolor[HTML]{a7dfb9} $\checkmark$ & $99.39$ & $99.92$ & $99.65$ & $99.86$ \\
		\cmidrule{2-10}%
		                  & \cellcolor[HTML]{a7dfb9} $\checkmark$ & \cellcolor[HTML]{a7dfb9} $\checkmark$                & \cellcolor[HTML]{a7dfb9} $\checkmark$ & \cellcolor[HTML]{f4729c}$\times$ & \cellcolor[HTML]{f4729c}$\times$ & $99.33$ & $99.51$ & $99.42$ & $99.77$ \\
		\cmidrule{2-10}%
		                  & \cellcolor[HTML]{a7dfb9} $\checkmark$ & \cellcolor[HTML]{a7dfb9} $\checkmark$                & \cellcolor[HTML]{a7dfb9} $\checkmark$ & \cellcolor[HTML]{a7dfb9} $\checkmark$ & \cellcolor[HTML]{f4729c} $\times$ & $99.33$ & $99.51$ & $99.42$ & $99.77$ \\
		\cmidrule{2-10}%
		                  & \cellcolor[HTML]{a7dfb9} $\checkmark$ & \cellcolor[HTML]{a7dfb9} $\checkmark$                & \cellcolor[HTML]{f4729c} $\times$ & \cellcolor[HTML]{a7dfb9} $\checkmark$ & \cellcolor[HTML]{f4729c} $\times$ & $99.97$ & $99.79$ & $99.88$ & $99.95$ \\
		\cmidrule{2-10}%
		                  & \cellcolor[HTML]{a7dfb9} $\checkmark$ & \cellcolor[HTML]{a7dfb9} $\checkmark$                & \cellcolor[HTML]{f4729c} $\times$ & \cellcolor[HTML]{a7dfb9} $\checkmark$ & \cellcolor[HTML]{a7dfb9} $\checkmark$ & $99.36$ & $99.72$ & $99.54$ & $99.82$ \\
		\cmidrule{2-10}%
		                  & \cellcolor[HTML]{a7dfb9} $\checkmark$ & \cellcolor[HTML]{a7dfb9} $\checkmark$                & \cellcolor[HTML]{f4729c} $\times$ & \cellcolor[HTML]{f4729c} $\times$ & \cellcolor[HTML]{a7dfb9} $\checkmark$ & $99.56$ & $99.74$ & $99.65$ & $99.86$ \\
		\cmidrule{2-10}%
		                  & \cellcolor[HTML]{a7dfb9} $\checkmark$ & \cellcolor[HTML]{a7dfb9} $\checkmark$                & \cellcolor[HTML]{a7dfb9} $\checkmark$ & \cellcolor[HTML]{a7dfb9} $\checkmark$ & \cellcolor[HTML]{a7dfb9} $\checkmark$ & $100$ & $100$ & $100$ & $100$ \\
		\midrule[0.01cm]
		Jhuisi            & \cellcolor[HTML]{f4729c} $\times$ & \cellcolor[HTML]{a7dfb9} $\checkmark$                & \cellcolor[HTML]{f4729c} $\times$ & \cellcolor[HTML]{f4729c} $\times$ & \cellcolor[HTML]{f4729c} $\times$ & $99.88$ & $99.88$ & $99.88$ & $99.92$ \\
		\cmidrule{2-10}%
						  & \cellcolor[HTML]{a7dfb9} $\checkmark$ & \cellcolor[HTML]{f4729c} $\times$                & \cellcolor[HTML]{f4729c} $\times$ & \cellcolor[HTML]{f4729c} $\times$ & \cellcolor[HTML]{f4729c} $\times$ & $96.88$ & $92.39$ & $94.51$ & $99.12$ \\
		\cmidrule{2-10}%
						  & \cellcolor[HTML]{a7dfb9} $\checkmark$ & \cellcolor[HTML]{a7dfb9} $\checkmark$                & \cellcolor[HTML]{a7dfb9} $\checkmark$ & \cellcolor[HTML]{f4729c} $\times$ & \cellcolor[HTML]{a7dfb9} $\checkmark$ & $99.61$ & $99.54$ & $99.58$ & $99.70$ \\
		\cmidrule{2-10}%
						  & \cellcolor[HTML]{a7dfb9} $\checkmark$ & \cellcolor[HTML]{a7dfb9} $\checkmark$                & \cellcolor[HTML]{a7dfb9} $\checkmark$ & \cellcolor[HTML]{f4729c}$\times$ & \cellcolor[HTML]{f4729c}$\times$ & $97.98$ & $98.29$ & $98.13$ & $98.68$ \\
		\cmidrule{2-10}%
						  & \cellcolor[HTML]{a7dfb9} $\checkmark$ & \cellcolor[HTML]{a7dfb9} $\checkmark$                & \cellcolor[HTML]{a7dfb9} $\checkmark$ & \cellcolor[HTML]{a7dfb9} $\checkmark$ & \cellcolor[HTML]{f4729c} $\times$ & $99.22$ & $99.09$ & $99.15$ & $99.40$ \\
		\cmidrule{2-10}%
						  & \cellcolor[HTML]{a7dfb9} $\checkmark$ & \cellcolor[HTML]{a7dfb9} $\checkmark$                & \cellcolor[HTML]{f4729c} $\times$ & \cellcolor[HTML]{a7dfb9} $\checkmark$ & \cellcolor[HTML]{f4729c} $\times$ & $99.42$ & $99.49$ & $99.46$ & $99.62$ \\
		\cmidrule{2-10}%
						  & \cellcolor[HTML]{a7dfb9} $\checkmark$ & \cellcolor[HTML]{a7dfb9} $\checkmark$                & \cellcolor[HTML]{f4729c} $\times$ & \cellcolor[HTML]{a7dfb9} $\checkmark$ & \cellcolor[HTML]{a7dfb9} $\checkmark$ & $99.77$ & $99.51$ & $99.64$ & $99.74$ \\
		\cmidrule{2-10}%
						  & \cellcolor[HTML]{a7dfb9} $\checkmark$ & \cellcolor[HTML]{a7dfb9} $\checkmark$                & \cellcolor[HTML]{f4729c} $\times$ & \cellcolor[HTML]{f4729c} $\times$ & \cellcolor[HTML]{a7dfb9} $\checkmark$ & $99.63$ & $99.04$ & $99.33$ & $99.53$ \\
		\cmidrule{2-10}%
						  & \cellcolor[HTML]{a7dfb9} $\checkmark$ & \cellcolor[HTML]{a7dfb9} $\checkmark$                & \cellcolor[HTML]{a7dfb9} $\checkmark$ & \cellcolor[HTML]{a7dfb9} $\checkmark$ & \cellcolor[HTML]{a7dfb9} $\checkmark$ & $100$ & $100$ & $100$ & $100$ \\
		\midrule[0.01cm]
		Honey7            & \cellcolor[HTML]{f4729c} $\times$ & \cellcolor[HTML]{a7dfb9} $\checkmark$                & \cellcolor[HTML]{f4729c} $\times$ & \cellcolor[HTML]{f4729c} $\times$ & \cellcolor[HTML]{f4729c} $\times$ & $99.97$ & $99.55$ & $99.76$ & $99.94$ \\
		\cmidrule{2-10}%
						  & \cellcolor[HTML]{a7dfb9} $\checkmark$ & \cellcolor[HTML]{f4729c} $\times$                & \cellcolor[HTML]{f4729c} $\times$ & \cellcolor[HTML]{f4729c} $\times$ & \cellcolor[HTML]{f4729c} $\times$ & $96.92$ & $99.05$ & $97.93$ & $98.52$ \\
		\cmidrule{2-10}%
						  & \cellcolor[HTML]{a7dfb9} $\checkmark$ & \cellcolor[HTML]{a7dfb9} $\checkmark$                & \cellcolor[HTML]{a7dfb9} $\checkmark$ & \cellcolor[HTML]{f4729c} $\times$ & \cellcolor[HTML]{a7dfb9} $\checkmark$ & $99.97$ & $99.55$ & $99.76$ & $99.94$ \\
		\cmidrule{2-10}%
						  & \cellcolor[HTML]{a7dfb9} $\checkmark$ & \cellcolor[HTML]{a7dfb9} $\checkmark$                & \cellcolor[HTML]{a7dfb9} $\checkmark$ & \cellcolor[HTML]{f4729c}$\times$ & \cellcolor[HTML]{f4729c}$\times$ & $99.94$ & $99.09$ & $99.51$ & $99.89$ \\
		\cmidrule{2-10}%
						  & \cellcolor[HTML]{a7dfb9} $\checkmark$ & \cellcolor[HTML]{a7dfb9} $\checkmark$                & \cellcolor[HTML]{a7dfb9} $\checkmark$ & \cellcolor[HTML]{a7dfb9} $\checkmark$ & \cellcolor[HTML]{f4729c} $\times$ & $99.97$ & $99.55$ & $99.76$ & $99.94$ \\
		\cmidrule{2-10}%
						  & \cellcolor[HTML]{a7dfb9} $\checkmark$ & \cellcolor[HTML]{a7dfb9} $\checkmark$                & \cellcolor[HTML]{f4729c} $\times$ & \cellcolor[HTML]{a7dfb9} $\checkmark$ & \cellcolor[HTML]{f4729c} $\times$ & $100$ & $100$ & $100$ & $100$ \\
		\cmidrule{2-10}%
						  & \cellcolor[HTML]{a7dfb9} $\checkmark$ & \cellcolor[HTML]{a7dfb9} $\checkmark$                & \cellcolor[HTML]{f4729c} $\times$ & \cellcolor[HTML]{a7dfb9} $\checkmark$ & \cellcolor[HTML]{a7dfb9} $\checkmark$ & $99.97$ & $99.55$ & $99.76$ & $99.94$ \\
		\cmidrule{2-10}%
						  & \cellcolor[HTML]{a7dfb9} $\checkmark$ & \cellcolor[HTML]{a7dfb9} $\checkmark$                & \cellcolor[HTML]{f4729c} $\times$ & \cellcolor[HTML]{f4729c} $\times$ & \cellcolor[HTML]{a7dfb9} $\checkmark$ & $99.51$ & $99.51$ & $99.51$ & $99.89$ \\
		\cmidrule{2-10}%
						  & \cellcolor[HTML]{a7dfb9} $\checkmark$ & \cellcolor[HTML]{a7dfb9} $\checkmark$                & \cellcolor[HTML]{a7dfb9} $\checkmark$ & \cellcolor[HTML]{a7dfb9} $\checkmark$ & \cellcolor[HTML]{a7dfb9} $\checkmark$ & $100$ & $100$ & $100$ & $100$ \\
		\botrule
	\end{tabular*}
\end{table}

To conduct an ablation study, we eliminate sequence modality from multimodal information to assess the impact of background or context of events on model performance. The elimination of sequence modality results in reduced performance, which signifies that sequential signals are essential for addressing log-based anomaly detection and illustrate the complementarity between log events and the background or context of events. Additionally, we eliminate module MHIA from CoLog and update it with MHA. Furthermore, we eliminate MAL and the balancing layer, which was intended to enable the fusion of various modalities, resulting in a decrease in all performance metrics. These results demonstrate the efficacy of MHIA, MAL, and the balancing layer in representation learning for detecting anomalies in logs. Finally, evaluate the distinct interconnections among diverse modules. The results demonstrate that CoLog's enhancements are not merely additive; they originate from the synergistic combination of its different modules.

\subsubsection{Uncharted Territory: Groundbreaking Study}\label{subsubsec5: unchartedterritorygroundbreakingstudy}

A unified framework for detecting point and collective anomalies in logs optimizing the process and guaranteeing consistency in the analysis. This comprehensive method reduces blind spots that may arise from examining each anomaly type in isolation and improves the accuracy and reliability of the detection system. Utilizing a unified framework enables to deployment of resources more efficiently and addresses anomalies, hence enhancing the overall resilience. The unified framework enables the application of different modalities for improved and informative point and collective anomaly detection. As the discussion in Section~\ref{subsec4: classification}, CoLog assigns labels to log messages and the background or context of log events. Subsequently, upon receiving input ${I}$, it aims to learn and predict the label associated with each input. This method will involve four classes. Where $\mathbf{0}$ indicates both modalities are negative, $\mathbf{1}$ signifies that only the log event is anomalous, 2 denotes that only the background or context of the log event is anomalous, and 3 represents a normal instance of input. The experimental results of groundbreaking study on the Casper, Jhuisi, Nssal, Honey7, Zookeeper, Hadoop, and BlueGene/L datasets are shown in Table~\ref{tab16}. It can be seen that CoLog performs excellent performance on this task. CoLog achieves excellent stable outcomes across all metrics on all datasets. For example, CoLog achieved a recall of 99.94\% on the zookeeper dataset. The efficacy of CoLog in detecting point and collective anomalies in log files is primarily related to its ability to learn representations from various modalities of log files via distinct transformer blocks along with its ability to capture interactions across these modalities. This approach enables the CoLog to utilize the informative characteristics of log modalities, including semantic and sequential features of logs. Through a collaboration tunnel, MHIA, MAL, and the balancing layer, CoLog can more efficiently detect both point and collective anomalies. This collaboration guarantees that no essential insights are ignored. Note that the development of CoLog for more modalities is straightforward.

\begin{table}[h]
	\caption{The experimental results of a groundbreaking study on all training datasets.}\label{tab16}%
	\begin{tabular}{@{}lllll@{}}
		\toprule
		Dataset    & Precision & Recall  & F1-score & Accuracy \\
		\midrule[0.01cm]
		Casper     & $99.43$   & $99.40$ & $99.41$  & $99.64$  \\
		Jhuisi     & $99.82$   & $99.48$ & $99.65$  & $99.70$  \\
		Nssal      & $99.32$   & $99.70$ & $99.51$  & $99.91$  \\
		Honey7     & $99.77$   & $98.90$ & $99.33$  & $99.77$  \\
		Zookeeper  & $99.98$   & $99.94$ & $99.96$  & $99.99$  \\
		Hadoop     & $99.17$   & $99.84$ & $99.50$  & $99.98$  \\
		BlueGene/L & $99.94$   & $99.89$ & $99.91$  & $100$  \\
		\botrule
	\end{tabular}
\end{table}

\subsubsection{Contribution Chronicles: Comparative Study of Contributions}\label{subsubsec5: contributionchroniclescomparativestudyofcontributions}

The comprehensive evaluation of contributions in existing literature compared to our study's advancements underscores significant distinctions and synergies. FastLogAD \citep{ref76} presents efficient utilization of normal data coupled with an innovative anomaly generation method. In contrast, our approach, CoLog, augments this by encoding log records across multiple modalities. The pylogsentiment \citep{ref38} focuses on SA for log anomaly detection and addressing class imbalance. CoLog surpasses this by applying MSA to log anomaly detection that outperforms state-of-the-art methods. RAGLog \citep{ref91}, a retrieval-augmented generation model, primarily emphasizes storing normal log entries in a vector database. Our method, however, extends beyond mere storage to learn about signatures of abnormalities. UMFLog \citep{ref83} employs a dual-model architecture integrating BERT for semantic feature extraction and VAE for statistical feature analysis, along with handling long log data sequences. CoLog similarly learns long sequences of log records effectively but transcends UMFLog by incorporating a collaborative approach to capture interactions between modalities and utilizing MAL to address heterogeneity. LogMS \citep{ref58} implements a two-step model. A multi-source information fusion-based LSTM for anomaly detection followed by a GRU network for probability label estimation. Conversely, CoLog's unified framework for detecting abnormalities offers a streamlined and more comprehensive solution. MDFULog \citep{ref44} addresses noise in log data and adopts an informer-based anomaly detection approach, whereas CoLog's modules, i.e., collaborative transformer and MAL, inherently mitigate noise. Finally, multi-feature fusion (MFF) \citep{ref43} leverages late fusion for MFF-based anomaly detection by evaluating HTTP textual content, status code, and frequency features. Yet, CoLog's intermediate fusion mechanism significantly enhances performance metrics. Collectively, the contributions of CoLog not only embody advancements found in existing methods but also amalgamate their strengths within a cohesive, unified framework. Thus, CoLog's approach is not only multifaceted, addressing a wide spectrum of challenges presented by log anomaly detection, but also showcases superior performance metrics compared to individual contributions of FastLogAD, pylogsentiment, RAGLog, UMFLog, LogMS, MDFULog, and MFF. The comparative analysis in Table~\ref{tab17} distinctly illustrates how CoLog stands at the forefront of innovation. CoLog's ability to effectively encode log records utilizing various modalities, addressing heterogeneity through MAL, and demonstrating exceptional performance in detecting both point and collective abnormalities within a unified architecture validates CoLog as an innovative approach in the field of log anomaly detection. 

\begin{sidewaystable}
	\caption{The comprehensive overview of contributions in existing literature compared to our study.}\label{tab17}
	\begin{tabular*}{\textheight}{@{\extracolsep\fill}l|c|c|l|c|l}
		\toprule%
		Paper                        & Year & Methodology                    & Datasets Used                                          & Mean F1-score & Key Contributions               \\
		\midrule[0.01cm]
		FastLogAD                    & 2024 & Sequence-based                 & 1) HDFS\footnotemark[1] \citep{ref18}                  & $94.18$       & 1) Efficient utilization        \\
		\citep{ref76}                &      & unimodal                       & 2) BlueGene/L                                          &               & of normal data.                 \\
                                     &      & unsupervised learning          & 3) Thunderbird                                         &               & 2) Innovative anomaly           \\
                                     &      &                                & \citep{ref123}                                         &               & generation method.              \\
		\midrule[0.01cm]
		pylogsentiment               & 2020 & Semantic-based                 & 1) Spark   \hspace{0.75cm} 6) Nssal                    & $99.14$       & 1) Implements sentiment         \\
		\citep{ref38}                &      & unimodal                       & 2) Honey5  \hspace{0.53cm} 7) Honey7                   &               & analysis for log anomaly        \\
                                     &      & supervised learning            & 3) Windows \hspace{0.31cm} 8) Zookeeper                &               & detection.                      \\
                                     &      &                                & 4) Casper  \hspace{0.6cm}  9) Hadoop                   &               & 2) Addressing class             \\
                                     &      &                                & 5) Jhuisi  \hspace{0.6cm}  10) BlueGene/L              &               & imbalance.                      \\
		\midrule[0.01cm]
		RAGLog                       & 2024 & LLM-based                      & 1) BlueGene/L                                          & $89.00$       & 1) A retrieval-augmented        \\
		\citep{ref91}                &      & unimodal                       & 2) Thunderbird                                         &               & generation model that           \\
                                     &      & unsupervised learning          &                                                        &               & employs a vector database       \\
                                     &      &                                &                                                        &               & to store normal log entries.    \\
		\midrule[0.01cm]
		UMFLog                       & 2023 & Independent                    & 1) HDFS                                                & $99.56$       & 1) Employs a dual-model         \\
		\citep{ref83}                &      & network-based                  & 2) BlueGene/L                                          &               & architecture with BERT for      \\
                                     &      & multimodal                     & 3) Thunderbird                                         &               & semantic feature extraction     \\
                                     &      & unsupervised learning          &                                                        &               & and VAE for statistical         \\
                                     &      &                                &                                                        &               & feature analysis.               \\
                                     &      &                                &                                                        &               & 2) Handles long sequences       \\
                                     &      &                                &                                                        &               & of log data.                    \\
		\midrule[0.01cm]
		LogMS                        & 2024 & Early fusion-based             & 1) HDFS                                                & $99.10$       & 1) Employs a two-step model.    \\
		\citep{ref58}                &      & multimodal unsupervised        & 2) BlueGene/L                                          &               & The first step uses a           \\
                                     &      & \& semi-supervised learning    &                                                        &               & multi-source information        \\
                                     &      &                                &                                                        &               & fusion-based LSTM to detect     \\
                                     &      &                                &                                                        &               & anomalies by utilizing          \\
                                     &      &                                &                                                        &               & semantic, sequential, and       \\
                                     &      &                                &                                                        &               & quantitative data. Following    \\
                                     &      &                                &                                                        &               & that, a probability label       \\
                                     &      &                                &                                                        &               & estimation-based GRU            \\
                                     &      &                                &                                                        &               & network is used.          \\
		\midrule[0.01cm]
		MDFULog                      & 2023 & Intermediate fusion-based      & 1) HDFS                                                & $97.00$       & 1) Addresses noise in log data. \\
		\citep{ref44}                &      & multimodal                     & 2) OpenStack                                           &               & 2) Informer-based anomaly       \\
                                     &      & supervised learning            & \citep{ref123}                                         &               & detection.                      \\
		\midrule[0.01cm]
		MFF                          & 2023 & Late fusion-based              & 1) LLSD2                                               & $93.10$       & 1) Detects web scanning         \\
		\citep{ref43}                &      & multimodal                     & \citep{ref43}                                          &               & behavior by considering HTTP    \\
                                     &      & supervised learning            &                                                        &               & textual content, status code,   \\
                                     &      &                                &                                                        &               & and frequency features.         \\
                                     &      &                                &                                                        &               & 2) Employ late fusion           \\
                                     &      &                                &                                                        &               & MFF-based network to            \\
                                     &      &                                &                                                        &               & detect anomalies.               \\
		\midrule[0.01cm]
		CoLog                        & 2025 & Intermediate fusion-based      & 1) Spark   \hspace{0.75cm} 6) Nssal                    & $99.99$       & 1) Encodes log records          \\
		                             &      & multimodal                     & 2) Honey5  \hspace{0.53cm} 7) Honey7                   &               & collaboratively according to    \\
                                     &      & supervised learning            & 3) Windows \hspace{0.31cm} 8) Zookeeper                &               & various log modalities.         \\
                                     &      &                                & 4) Casper  \hspace{0.6cm}  9) Hadoop                   &               & 2) Employs MAL to address       \\
                                     &      &                                & 5) Jhuisi  \hspace{0.6cm}  10) BlueGene/L              &               & heterogeneity among modalities. \\
                                     &      &                                &                                                        &               & 3) Outperforms state-of-the-art \\
                                     &      &                                &                                                        &               & methods.                        \\
                                     &      &                                &                                                        &               & 4) Detects point and collective \\
                                     &      &                                &                                                        &               & abnormalities within a unified  \\
                                     &      &                                &                                                        &               & framework.                      \\
		\botrule
	\end{tabular*}
	\footnotetext[1]{Hadoop Distributed File System}
\end{sidewaystable}

\section{Conclusion}\label{sec6: conclusion}

In this paper, we introduce CoLog, a novel framework for collaborative encoding and anomaly detection that utilizes various log modalities. CoLog employs a supervised learning methodology, with its outcomes in log anomaly detection highlighting its scholarly significance in establishing an upper performance limit for this domain. With adequate labeled data, CoLog develops a standard of "theoretical accuracy" for evaluating unsupervised or semi-supervised approaches. Consequently, CoLog serves as a reference baseline for forthcoming research trajectories, offering comparative value and direction for the development of more efficient models. Additionally, the proposed approach addresses the inherent heterogeneity of log data by implementing MAL, ensuring robust performance across multiple benchmarks. CoLog's ability to detect both point and collective anomalies within a unified framework distinguishes it from conventional methods that focus solely on one type of anomaly. This comprehensive detection capability underpins the versatility and adaptability of CoLog, making it a significant advancement in the field of log anomaly detection. The potential applications of CoLog are vast, spanning from enhancing security measures through more accurate anomaly detection in cybersecurity logs to improving operational efficiency by identifying system irregularities in real-time. Additionally, its robust performance under varying noise conditions demonstrates its reliability and effectiveness in diverse environments.

\subsection{Limitations}\label{subsec6: limitations}

\bmhead{Real-time application}

The present study assesses CoLog in a batch-processing mode. Real-time anomaly detection frequently entails supplementary constraints, like latency and limitations on computational resources. Examining and enhancing CoLog's performance in real-time settings is essential for effective implementation.

\bmhead{Log entries that are not understandable by humans}

Certain log entries, especially those from operating systems, may be difficult for humans to decipher, presenting a difficulty for first-time interpretation and labeling of such data.

\bmhead{Adapting to evolving log structures}

The logs develop over time with system updates, necessitating continual adaptation and retraining of the model to preserve validity.

\bmhead{The Capacity for Noise Resilience}

While CoLog exhibits robustness to varying noise levels, certain extreme scenarios or certain noise types may still substantially impair the performance.

\subsection{Future Works}\label{subsec6: futureworks}

Future work could explore the integration of CoLog with real-time monitoring systems to enable proactive anomaly management and automated responses to detected anomalies on industrial systems. Moreover, extending CoLog's application to other domains, such as financial fraud detection or health monitoring, could further validate its efficacy and adaptability. Since CoLog assumes a moderately imbalanced dataset and stable log templates, which generally hold in benchmark datasets but may vary in dynamic production systems. In extreme imbalance scenarios (e.g., anomaly ratio <1\%) or in environments with rapidly evolving templates, performance may degrade due to representation drift. To mitigate this, CoLog's balancing layer can be combined with adaptive resampling, online calibration, or few-shot fine-tuning. We also note that continual learning extensions could further improve adaptability. Future work will explore these directions to ensure robustness under diverse operational conditions.
Additionally, alternative modalities of log data, including quantitative information, may also be employed in the anomaly detection process. The development of CoLog for additional modalities is straightforward. Also, investigating adaptive mechanisms for better managing various types of noise could be a beneficial avenue. The promising performance and broad applicability of CoLog suggest that it holds considerable potential for driving advancements in multiple industries where accurate and timely anomaly detection is critical. By addressing current limitations in log analysis, CoLog paves the way for more sophisticated and comprehensive solutions to complex data challenges.

\backmatter

\bmhead{Acknowledgements}

We express our profound appreciation to \href{mailto:mahdi.azh1998@gmail.com}{Mahdi Asgharzadeh} for his significant contributions to the development of the Alarmif website.

\section*{Declarations}

\bmhead{Funding}

Not applicable.

\bmhead{Conflict of interest/Competing interests}

The authors declare that they have no conflicts of interest or competing interests that could have influenced the content, analysis, or conclusions of this work.

\bmhead{Consent for publication}

All authors have provided their consent for the publication of this work, affirming their agreement with its content and its dissemination in the designated journal.

\bmhead{Data availability}

The datasets generated and/or analysed during the current study are available in the loghub repository at \href{https://github.com/logpai/loghub}{https://github.com/logpai/loghub} and in the pylogsentiment repository at \href{https://github.com/studiawan/pylogsentiment}{https://github.com/studiawan/pylogsentiment}.

\bmhead{Materials availability}

The datasets, computing resources, algorithms, and libraries employed in this AI development can be obtained upon request, contingent upon relevant access policies and license agreements.

\bmhead{Code availability}

The paper's implementation code is openly accessible on GitHub at \href{https://github.com/NasirzadehMoh/CoLog}{https://github.com/NasirzadehMoh/CoLog}, guaranteeing transparency and reproducibility for ongoing development and collaboration.

\bmhead{Author contribution}

Methodology, theoretical analysis, algorithm, implementation, experiments, and writing - \textbf{Mohammad Nasirzadeh}; Supervision, architecture design, and revision - \textbf{Jafar Tahmoresnezhad}; Theoretical analysis and architecture design - \textbf{Parviz Rashidi-Khazaee}.

\bigskip

\bibliography{sn-bibliography}

\end{document}